\documentclass{article}

\usepackage[utf8]{inputenc}
\usepackage[T1]{fontenc}
\usepackage{PRIMEarxiv}
\usepackage{authblk}
\usepackage{amsmath, amssymb, amsfonts, bm}
\usepackage{amsthm}
\usepackage{enumitem} 
\usepackage{graphicx}
\usepackage{subcaption}
\usepackage{booktabs}
\usepackage{multirow}
\usepackage{longtable}
\usepackage{amsmath, amssymb, amsthm}
\usepackage{tablefootnote}
\newtheorem{theorem}{Theorem}
\usepackage{tabularx}
\usepackage{colortbl}
\usepackage{adjustbox}
\usepackage{siunitx}
\usepackage{xcolor}
\usepackage{algorithm}
\usepackage{hyperref}
\usepackage{algorithmicx}
\usepackage{algpseudocode}
\usepackage{mathrsfs}
\usepackage{xcolor}
\usepackage{textcomp}
\usepackage{comment}
\usepackage{url}
\usepackage{natbib}
\usepackage{tcolorbox}
\usepackage{fancyhdr}
\usepackage{microtype}
\usepackage[title]{appendix}
\usepackage{setspace}
\usepackage[left]{lineno}
\usepackage{geometry}
\usepackage[dvipsnames]{xcolor}         
\usepackage{hyperref}
\usepackage{threeparttable}
\usepackage{booktabs}
\usepackage{pifont}                 

\usepackage{xurl}             
\usepackage{hyperref}         
\hypersetup{
  breaklinks=true,            
  colorlinks=true,            
  urlcolor=blue               
}

\hypersetup{
  colorlinks=true,                      
  citecolor=Blue,                       
  linkcolor=Blue,                       
  urlcolor=Blue                         
}
\microtypecontext{spacing=nonfrench}

\title{Virtual Sensing to Enable Real-Time Monitoring of Inaccessible Locations \& Unmeasurable Parameters}

\author[1]{Kazuma Kobayashi}
\author[1]{Farid Ahmed}
\author[2,3]{Jaewan Park}
\author[4]{Subhankar Sarkar}
\author[2,3]{Seid Koric}
\author[1,4,5]{Souvik Chakraborty}
\author[1,3]{Syed Bahauddin Alam\textsuperscript{*}}

\affil[1]{Nuclear, Plasma \& Radiological Engineering Department, Grainger College of Engineering,  University of Illinois Urbana-Champaign, Urbana, IL, USA}
\affil[2]{Mechanical Science and Engineering Department, Grainger College of Engineering, University of Illinois Urbana-Champaign, Urbana, IL, USA}
\affil[3]{National Center for Supercomputing Applications, Urbana, IL, USA}
\affil[4]{Department of Applied Mechanics, Indian Institute of Technology Delhi, New Delhi, India}
\affil[5]{Yardi School of Artificial Intelligence, Indian Institute of Technology Delhi}
\affil[*]{Corresponding author: \href{mailto:alams@illinois.edu}{alams@illinois.edu}}

\newtheorem{assumption}{Assumption}

\begin{document}

\clearpage
\setcounter{page}{1}
\maketitle
\pagestyle{fancy}



\begin{abstract}
\noindent \begin{tcolorbox}[colback=blue!10!white, colframe=blue!50!black,
    left=0mm, right=0mm]

Real-time monitoring of safety-critical interior states remains an open problem across energy, environmental, and industrial systems where direct instrumentation is infeasible. Existing approaches based on governing equations, discrete state vectors, or fixed sensor locations fail to provide mesh-independent, field-level reconstructions at arbitrary interior coordinates under real-time constraints. Here we introduce neural operator-based virtual sensing as a general framework for recovering inaccessible interior fields from sparse boundary measurements, instantiated with MIMONet, a multi-input, multi-output neural operator that fuses heterogeneous inputs and decodes coupled physical fields through a shared latent representation. Across three engineering-grade evaluations of escalating complexity, from confined recirculating flows to pressurized water reactor subchannels and compact power-system heat exchangers, MIMONet achieves $\leq 5\%$ relative reconstruction error and sub-millisecond inference on an NVIDIA H200, with calibrated uncertainty and noise resilience demonstrated on the heat-exchanger case. We validate the framework on three independent real-world datasets spanning electrochemical energy, atmospheric science, and physical oceanography: current-density mapping in a hydrogen fuel cell, wind-speed profiling on instrumented meteorological towers, and North Atlantic ocean-state from a global ocean reanalysis. MIMONet improves over classical virtual-sensing baselines by $72$--$84\%$: $84\%$ on fuel-cell internal-field recovery from $10\%$ sensor coverage and $83\%$ on hub-height wind-speed recovery from lower-level tower sensors. Most notably, it recovers sea-surface height, a field with zero direct sensors, from its learned coupling to observed temperature and salinity at $4.1\times$ lower error than the best classical predictor, a cross-modal field recovery beyond the reach of classical interpolation. These results establish operator-based virtual sensing as a practical route to real-time field observability in systems where the states that matter most cannot be directly measured.
\end{tcolorbox}
\end{abstract}


\section{Introduction}
Across energy, environmental, and industrial systems, safety-critical physical states are often located where sensors cannot reach: behind radiation shielding, within electrochemical layers, across atmospheric columns, or beneath ocean surfaces. Recovering these interior fields from sparse accessible measurements is therefore central to reactor monitoring, fuel-cell diagnostics, wind-energy assessment, and geophysical observation. The challenge is most acute in next-generation nuclear reactors, whose compact, increasingly autonomous operation demands real-time estimates of internal pressure, temperature, and flow precisely where radiation, thermal limits, and geometric confinement preclude durable instrumentation~\cite{meta2025deal, us2025deal_reuters, hossain2024sensor, lee2026agentic}. The same inaccessibility, in different physical form, recurs across the electrochemical, atmospheric, and oceanographic systems examined here. Although high-fidelity simulation provides indispensable reference fields for design and safety analysis, resolving the coupled governing equations can require hundreds of seconds per solution and is therefore unsuitable for continuous monitoring. This motivates virtual sensing: reconstructing unmeasurable interior states from limited boundary observations. Rather than producing isolated point estimates, an effective virtual sensor must recover coupled fields and derived quantities, such as wall shear and pressure gradients, at arbitrary inaccessible locations while providing millisecond-scale inference and calibrated uncertainty~\cite{niresi2024physics,liu2019understanding, garg2025distributionfree}. Such capability can restore the physical observability required by digital twins~\cite{zhao2024virtual,hossain2025virtual}.

Virtual sensing encompasses model-based and data-driven approaches~\cite{niresi2024physics,zhao2024virtual,zhao2024virtual_b}. Model-based methods include Luenberger and unknown-input observers~\cite{luenberger1971}, Kalman-filter families~\cite{julier2004unscented,assimilation2009ensemble,niresi2024physics}, moving-horizon estimation~\cite{alessandri2020moving,tuveri2023regularized}, Gaussian-process latent-force models~\cite{bilbao2022gplf}, and spatial estimators such as kriging~\cite{hu2025online}. These methods are effective when the governing dynamics or covariance structure is known. Importantly, classical PDE observers can reconstruct spatially distributed states from boundary measurements; their limitation is therefore not an inability to recover fields in principle. Rather, they recursively propagate a state under an explicit governing model and observation operator, typically using a prescribed grid or spatial basis whose coefficients are estimated. Consequently, evaluation at new spatial coordinates depends on that representation, and the governing model remains part of online inference. Data-driven virtual sensors instead learn input--output relationships directly from measurements, using models ranging from support vector machines and recurrent networks to graph architectures that capture sensor interactions~\cite{niresi2024physics,vs_rnn_21,zhao2024virtual,dimitrov2022virtual,masti2021machine,vs_gnn_24_25,zhao2024virtual_b}. Their established implementations typically return scalar or low-dimensional outputs at a predetermined set of locations fixed during training. The problem considered here occupies a different deployment regime: learning a map from sparse, heterogeneous measurements on a sensing region to a spatially continuous, coupled physical field over a potentially disjoint reconstruction region. Once trained, the model must evaluate this field at arbitrary query coordinates through a single forward pass, without recursively evolving an explicit PDE model or retraining for each operating condition. The distinction is therefore not between methods that can and cannot estimate hidden states, but between online model-based state propagation and amortized, coordinate-conditioned field reconstruction. Achieving the latter requires learning mappings between input and output functions. The question addressed here is therefore whether operator-based virtual sensing can reconstruct spatially continuous, coupled interior fields at arbitrary inaccessible coordinates from sparse, heterogeneous measurements while providing real-time inference and calibrated uncertainty.

Scientific machine learning offers two distinct approaches to field reconstruction. Physics-informed neural networks (PINNs) represent the solution of an individual PDE instance by embedding the governing equations, boundary conditions, and available observations in the training objective~\cite{raissi2019physics,goswami2020transfer,chakraborty2021transfer,navaneeth2023stochastic}. Their mesh-free formulation has enabled forward and inverse analysis across complex physical systems, but standard PINNs require instance-specific optimization when boundary conditions or system parameters change. Neural operators address a fundamentally different problem: rather than approximating a single solution function, they learn an operator $\mathcal{G}$ between input and output function spaces across a family of admissible conditions~\cite{lu2021learning}. This amortizes the training cost and enables rapid inference for previously unseen operating states. Physics-informed neural operators (PINOs) combine these paradigms by imposing governing-equation residuals during operator training, thereby reducing dependence on labelled solution data while retaining operator-level generalization~\cite{li2021physicsPINO,tripura2024physics,sarkar2026physicsgeometry}. Coordinate-conditioned operators such as DeepONet are particularly suited to virtual sensing because they evaluate reconstructed fields at arbitrary query points without retraining or regridding. Other operator families provide complementary representations of the underlying functional map. FNO captures global interactions through Fourier-domain kernel integration~\cite{li2020fourier}; WNO uses spatially and spectrally localized wavelet representations to resolve multiscale solution features~\cite{tripura2023wavelet}; and GeoFNO extends spectral learning to irregular geometries through a learned correspondence with a structured computational domain~\cite{li2023fourier,recfno2024}. Graph-based operators directly accommodate irregular domains and unstructured discretizations: GNO performs kernel integration over domain graphs~\cite{li2020neural}, while Sp$^{2}$GNO combines spatial message passing with spectral graph representations to capture both localized interactions and long-range dependencies~\cite{sarkar2025spatiospectral}. Attention-based and nonlinear-decoder architectures, including OFormer, NOMAD, and CoDA-NO, further improve the representation of complex operator maps~\cite{kissas2022operator,seidman2022nomad,rahman2024codano_b}. These developments establish a broad foundation for learning PDE solution operators, but their published formulations were designed primarily for PDE surrogacy and do not jointly address the combination required for field-level virtual sensing: sparse and heterogeneous scalar- and function-valued observations, geometrically separated sensing and reconstruction domains, and simultaneous recovery of multiple coupled physical fields. This regime requires an operator that fuses heterogeneous observations into a shared physical representation and decodes interdependent fields at arbitrary inaccessible coordinates.

To address this gap, we introduce MIMONet, a multi-input, multi-output branch--trunk neural operator designed for field-level virtual sensing. MIMONet extends the branch--trunk formulation in three ways. First, separate branch networks encode heterogeneous observations, including scalar operating conditions, spatially distributed functions, and time-dependent sensor histories, into a common latent space. Second, these representations are fused multiplicatively, introducing explicit cross-modal interactions before spatial decoding. Third, a shared trunk network maps arbitrary query coordinates to channel-specific basis representations that decode multiple physical fields from the same fused latent state. Sharing the latent representation across outputs enables the model to capture dependencies among coupled quantities such as pressure, velocity, temperature, and turbulent kinetic energy, rather than treating each field as an independent regression target. The resulting operator maps sparse measurements to spatially continuous, multi-field reconstructions at arbitrary coordinates, including locations geometrically separated from the sensing region, without requiring regridding or online solution of the governing equations. Combined with stochastic inference and conformal calibration, the framework also provides uncertainty intervals with prescribed empirical coverage. MIMONet thus unifies heterogeneous sensor fusion, cross-domain field reconstruction, coupled-output decoding, and uncertainty-aware real-time inference within a single operator architecture.

The problem class can be stated precisely. This work learns an operator $\mathcal{G}:\mathcal{U}\rightarrow L^2(\mathcal{Y};\mathbb{R}^m)$ that maps sparse observations acquired on a sensing region $\mathcal{X}$ to continuous, coupled fields over a potentially disjoint reconstruction region $\mathcal{Y}$. Unlike classical state observers, it does not recursively propagate a state using an explicit governing model during inference; unlike regression-based virtual sensors, its output coordinates are not fixed during training. Throughout this Article, real-time denotes amortized forward-pass evaluation within a monitoring loop, rather than temporal forecasting or recursive state estimation.

We evaluate MIMONet through two complementary validation stages. First, three engineering-grade thermal--fluid configurations of increasing complexity, lid-driven cavity flow, a pressurized-water-reactor subchannel, and a compact power-system heat exchanger, test whether operator-based virtual sensing remains accurate as geometric confinement, multiphysics coupling, and sensor inaccessibility increase. Across these cases, MIMONet is compared with GeoFNO, NOMAD, and the learned-kriging model KCN, and achieves relative reconstruction errors of at most $5\%$ and sub-millisecond inference throughout, with calibrated uncertainty and robustness under substantial sensor corruption characterised in detail on the heat-exchanger case. Second, we assess practical performance on three independent real-world datasets: internal current-density reconstruction in a proton-exchange-membrane fuel cell~\cite{suarez2023cdm}, hub-height wind-speed reconstruction from lower-level measurements in the IEA-Wind tall-tower dataset~\cite{ramon2020iea}, and North Atlantic ocean-state reconstruction from sparse, disjoint observations in GLORYS12~\cite{lellouche2021glorys}. These problems progress from sparse internal-field recovery to extrapolation beyond the instrumented range and, finally, to cross-modal field recovery: reconstructing sea-surface height, a field that no sensor observes, from its learned coupling to the observed temperature and salinity fields. MIMONet reduces error by $84\%$ for the fuel-cell field and $83\%$ for hub-height wind speed relative to the best practical classical baselines, while reducing sea-surface-height error by a factor of $4.1$. Together, the simulation and real-world studies establish operator-based virtual sensing as a general framework for real-time field observability whenever the physical states of interest are inaccessible, uninstrumented, or not directly measurable, with the nuclear thermal--fluid case as its most safety-critical instance.

\section{Results}

We first define the operator-based virtual-sensing framework and then evaluate
it in three simulation-based configurations of increasing complexity: a
lid-driven cavity, a pressurized-water-reactor subchannel, and a compact
power-system heat exchanger. Subsequent analyses examine calibrated uncertainty,
comparative accuracy and robustness under degraded sensing, local extrapolation
beyond the training range, and real-time inference efficiency. 

\subsection{Operator-Based Virtual-Sensing Framework}
\label{sec:operator_framework}

\begin{figure}[!t]
    \centering
    \includegraphics[width=1.00\textwidth]{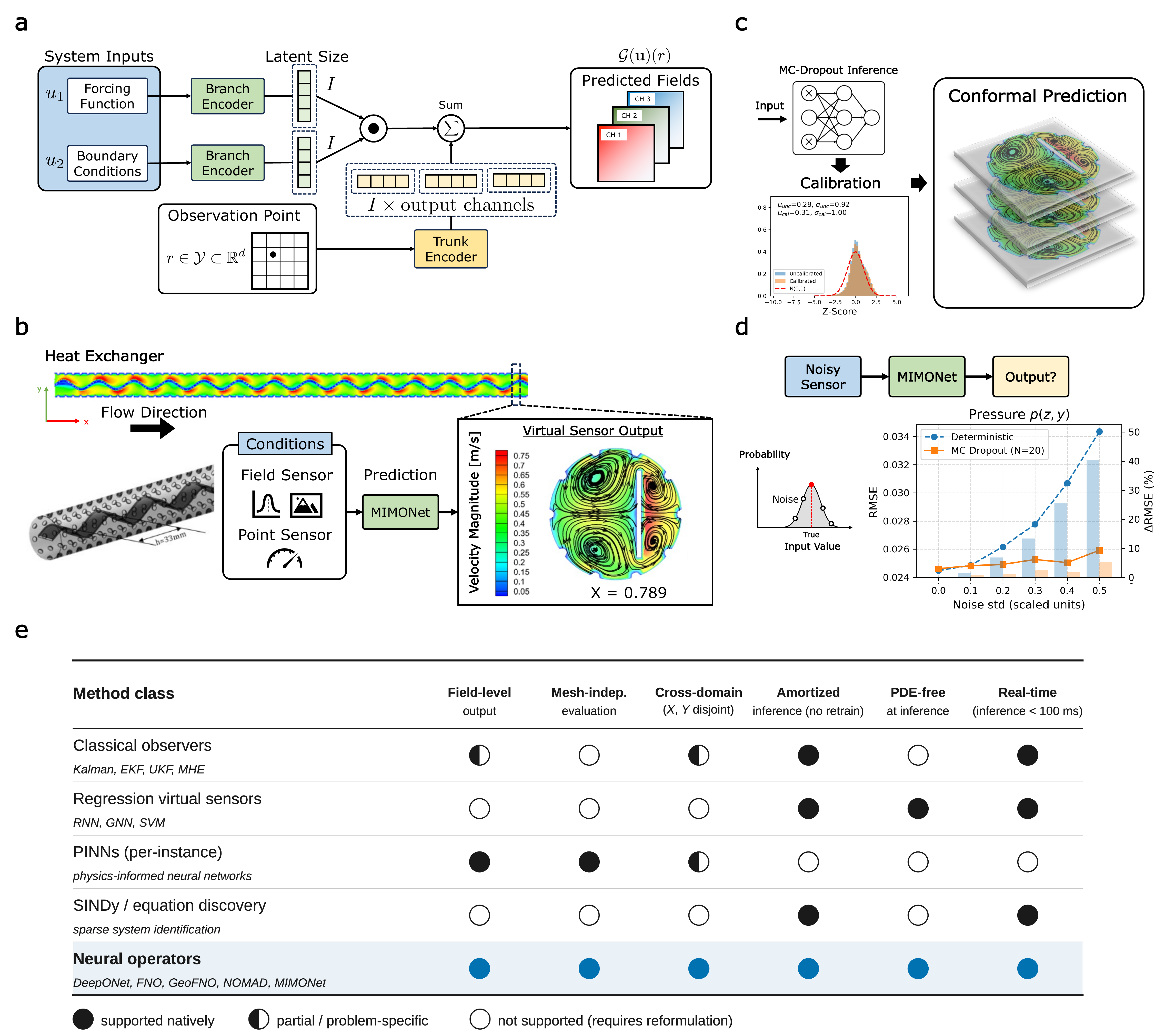}
    \caption{
    \textbf{MIMONet for full-field virtual sensing from sparse, heterogeneous measurements.} \textbf{a}, Modality-specific branch encoders map system inputs $\boldsymbol{u}=(u_1,\dots,u_n)$ into a shared latent representation through multiplicative fusion; a trunk network decodes it at arbitrary query coordinates $\mathbf{r}\in\mathcal{Y}$. \textbf{b}, The trained operator reconstructs coupled pressure, temperature and velocity fields in inaccessible regions from accessible sensors and operating conditions. \textbf{c}, MC-dropout predictive intervals are rescaled on a held-out conformal-calibration set to a nominal coverage of $95\%$. \textbf{d}, Robustness is assessed by injecting Gaussian noise into the scaled sensor inputs. \textbf{e}, Capability comparison across virtual-sensing method classes; filled, half-filled and open circles denote native, partial and unsupported capability.
    }
    \label{fig:overview}
\end{figure}

We formulate virtual sensing as an amortized operator that maps accessible
observations to spatially continuous interior fields. Unlike recursive state
estimation, the governing model is not solved during deployment; throughout
this work, real-time therefore refers to forward-pass latency within a
monitoring loop rather than temporal forecasting or recursive filtering.  Where an input carries a temporal history, the operator consumes the complete signal rather than the causal segment available up to the current instant, so its estimation setting is analogous to fixed-interval smoothing rather than to causal filtering. The real-time comparison is accordingly made against the seconds-to-minutes cost of the high-fidelity solver that would otherwise supply the field, not against the contemporaneous update of a recursive state estimator.
Let $\boldsymbol{u}=(u_1,\dots,u_n)\in\mathcal{U}$
denote the available heterogeneous observations, where each component $u_i\in\mathcal{F}_i$ may be a scalar operating condition, a spatially distributed function, or a time-dependent sensor history. The target is an $m$-channel physical field $\mathbf{s}(\mathbf{r})\in\mathbb{R}^m$ evaluated at query coordinates $\mathbf{r}$ on a reconstruction set $\mathcal{Y}\subset D\subset\mathbb{R}^d$. Field-level virtual sensing therefore amounts to learning the forward operator 
\begin{equation}
\mathcal{G}:\mathcal{U}\to\mathcal{S},\qquad \mathcal{G}(\boldsymbol{u})(\mathbf{r}) =\mathbf{s}(\mathbf{r}),\qquad \mathbf{r}\in\mathcal{Y},
\end{equation}
where $\mathcal{S}=L^2(\mathcal{Y};\mathbb{R}^m)$. When observations are acquired over a sensing region $\mathcal{X}$ that is disjoint from $\mathcal{Y}$, the formulation describes cross-domain inference, $\mathcal{X}\cap\mathcal{Y}=\emptyset$. Inaccessible locations remain within the spatial domain represented during training; they are inaccessible to physical instrumentation at deployment rather than necessarily outside the training distribution. Because $\mathbf{r}$ is an explicit input to the decoder, the trained operator can be queried at coordinate sets different from those used during training, although accuracy at previously unsampled coordinates remains an empirical generalization property.
MIMONet instantiates the operator $\mathcal{G}$ through modality-specific branch encoders and a coordinate-conditioned trunk network (Fig.~\ref{fig:overview}a). Each branch encoder $\phi_i:\mathcal{F}_i\to\mathbb{R}^{I} $
maps an input component $u_i$ to a latent vector of width $I$. The encoded inputs are fused element-wise into a shared latent representation,
\begin{equation}
    \psi_j = \prod_{i=1}^{n}[\phi_i(u_i)]_j, \qquad j=1,\dots,I.
\end{equation}
A trunk network maps each query coordinate $\mathbf{r}\in\mathbb{R}^d$ to per-channel decoding bases, $
\tau:\mathbb{R}^{d}\to\mathbb{R}^{I\times m}$,
with components $\tau_{j,o}(\mathbf{r})$. The prediction for output channel $o$ is
\begin{equation}
    \hat{s}_o(\mathbf{r},\{u_i\}) = \sum_{j=1}^{I} \tau_{j,o}(\mathbf{r}) \prod_{i=1}^{n}\phi_{i,j}(u_i) +\beta_o,
\end{equation}
where $\phi_{i,j}(u_i)=[\phi_i(u_i)]_j$ and $\beta_o$ is a channel-specific bias. The product over the encoded inputs introduces explicit cross-modal interactions before coordinate-conditioned decoding, while all output channels share the same fused representation $\boldsymbol{\psi}$. Channel-specific trunk bases retain the flexibility required to represent physical quantities with different spatial structures and scales.

This construction extends the branch--trunk formulation in three respects: separate branch networks accommodate heterogeneous scalar- and function-valued observations; multiplicative fusion introduces interactions among the encoded modalities; and shared-latent decoding reconstructs multiple coupled fields without learning an independent regressor for every field and spatial location. The shared representation enables the model to capture dependencies among quantities such as pressure, velocity, temperature, and turbulent kinetic energy. It does not, however, explicitly enforce incompressibility, momentum balance, energy conservation, or other governing constraints; these properties remain empirical rather than architectural guarantees. For a batch of $B$ inputs evaluated at $P$ coordinates, the branch--trunk contraction scales as $\mathcal{O}(BPIm)$, supporting dense field reconstruction within real-time latency budgets. 
Figure~\ref{fig:overview}e positions this formulation relative to established virtual-sensing and state-estimation classes. Classical observers support real-time recursive estimation when an explicit governing model is available, while regression-based virtual sensors provide rapid model-free estimates at locations generally fixed during training. Standard PINNs recover continuous fields but require instance-specific optimization for new boundary conditions or parameter configurations. Equation-discovery methods identify an explicit dynamical model rather than directly reconstructing inaccessible fields. At the method-class level, neural operators can jointly support continuous field output, coordinate-independent evaluation, cross-domain reconstruction, and amortized inference. This capability comparison motivates quantitative evaluation against methods operating in a compatible field-reconstruction regime: GeoFNO, NOMAD, and the learned-kriging model KCN.

\subsection{Boundary-Only Reconstruction of Confined Recirculating Flow}
\label{sec:ldc}

The lid-driven cavity (LDC), defined on the unit square
$\Omega=[0,1]\times[0,1]$, provides a controlled test of field-level virtual sensing under a boundary-only information constraint. Although geometrically simple, the flow develops a primary vortex, secondary corner recirculation, a high-shear layer beneath the moving lid, and spatially localized turbulent-kinetic-energy production (Fig.~\ref{fig:results1}a). These structures make the cavity a useful first test of whether boundary actuation contains sufficient information to reconstruct spatially distributed interior quantities. The CFD data are generated using the incompressible RANS equations with a $k$--$\varepsilon$ turbulence closure, as described in the Methods.

A time-dependent velocity is imposed on the top wall:
\begin{equation}
v_x(x,1,t)=V(t),
\qquad
v_y(x,1,t)=0,
\qquad x\in[0,1],
\end{equation}
while the remaining walls satisfy the no-slip condition, $\boldsymbol{v}=(0,0)$.
The system is initialized with zero velocity and uniform pressure, $
\boldsymbol{v}(x,y,0)=(0,0)$, $
p(x,y,0)=p_0$.
Consequently, the lid-velocity history $V(t)$ is the only observable input, while the complete interior flow state remains unavailable to direct measurement.
The input function $V(t)$ is randomly generated and sampled at $90$ time levels over the simulation interval (Fig.~\ref{fig:results1}b), giving
$\mathcal{U}_{\mathrm{ldc}}
=
\left\{V\in\mathbb{R}^{90}\right\}$.
The reconstruction targets are the final-time pressure $p$, velocity magnitude
$\|\boldsymbol{v}\|=\sqrt{v_x^2+v_y^2}$, and turbulent kinetic energy $k$, with output space $
\mathcal{S}_{\mathrm{ldc}}
=
\left(L^2(\Omega)\right)^3$.
The corresponding boundary-to-field operator is
\begin{equation}
\mathcal{G}_{\mathrm{ldc}}(V)(x,y)
=
\begin{bmatrix}
p(x,y)\\
\|\boldsymbol{v}(x,y)\|\\
k(x,y)
\end{bmatrix},
\qquad
(x,y)\in\Omega.
\end{equation}
The complete lid-velocity history is supplied as an input function, and the target is the spatial field at the prescribed final time. The task is therefore boundary-to-field reconstruction rather than causal forecasting or recursive temporal filtering.

The dataset comprises $4{,}937$ independently generated boundary-input/field-output pairs on a $4{,}225$-node unstructured mesh. It was divided at the trajectory level into $3{,}159$ training, $790$ validation, and $988$ held-out test samples, ensuring that complete lid-velocity histories remain confined to individual splits and preventing temporal leakage.
A single-branch MIMONet is used for this case. The $90$-dimensional lid-velocity input is encoded by a fully connected branch network with layer widths $
[90,512,512,512,256]$,
while the spatial coordinates $(x,y)$ are encoded by a trunk network with widths $
[2,256,256,256,256\times3]$.
The branch and trunk share a latent width of $I=256$, and their contraction produces the three outputs
$\{p,\|\boldsymbol{v}\|,k\}$. ReLU activations are used throughout, with no additional output-transform head.
The model is trained using an equally weighted mean-squared-error loss across the three output channels and the Adam optimizer. The initial learning rate is $10^{-3}$ with a weight decay of $10^{-7}$; the learning rate is reduced by a factor of $0.5$ when the validation loss plateaus. Training proceeds for at most $500$ epochs with a batch size of $8$ and early stopping after $10$ consecutive epochs without validation improvement. The branch inputs are min--max normalized to $[-1,1]$. Each output channel is independently transformed to the same range using scaling parameters fitted exclusively on the training set, preventing differences in physical units and magnitudes from causing one field to dominate the loss.

MIMONet achieves mean relative $\ell_2$ errors of $2.1\%$ for pressure, $5.0\%$ for velocity magnitude, and $2.8\%$ for turbulent kinetic energy, corresponding to a channel-mean error of $3.3\%$ (Fig.~\ref{fig:results1}c; complete per-channel results for all four models in Supplementary Table~S11). It attains the lowest error in every output channel. The corresponding channel-mean errors are $25.0\%$ for KCN, $41.9\%$ for NOMAD, and $68.0\%$ for GeoFNO, making the MIMONet error approximately $7.6$--$20.6\times$ lower across the three baselines. The comparatively larger error in $\|\boldsymbol{v}\|$ is consistent with the sensitivity of the thin shear layer and recirculation boundaries to small spatial displacements.
The test-set distributions in Fig.~\ref{fig:results1}c are concentrated at low error, and the sample-wise mean is unimodal without a pronounced high-error tail. Performance is therefore not driven by a small subset of favourable lid histories. In the median-error example (Fig.~\ref{fig:results1}d), MIMONet recovers the primary recirculating vortex, secondary corner structures, and the shear layer beneath the moving lid without visible spurious oscillations. Residual errors concentrate near the moving wall, corner recirculation regions, and zones of elevated turbulent kinetic energy, where the fields exhibit their strongest spatial gradients.
Errors in $p$, $\|\boldsymbol{v}\|$, and $k$ concentrate around the same dynamically sensitive structures. This behaviour is consistent with the shared-latent decoder representing common flow features across the output fields, although it does not establish exact satisfaction of the governing conservation laws. Overall, the cavity study demonstrates that spatially distributed interior fields can be reconstructed from boundary actuation alone before introducing the geometric confinement and multimodal inputs of the reactor-subchannel and heat-exchanger cases.

\subsection{Coupled-Field Reconstruction in a PWR Subchannel}
\label{sec:subchannel}

The pressurized-water-reactor (PWR) subchannel introduces geometric confinement, distributed thermal forcing, and coupled thermal--hydraulic outputs beyond the boundary-only cavity problem. Coolant passes through narrow, instrument-inaccessible regions between the fuel rods, where heating and wall confinement generate steep temperature gradients, secondary flow, and localized turbulence (Fig.~\ref{fig:results1}e). The objective is to reconstruct the transverse temperature, velocity, and turbulent-kinetic-energy fields from the known axial power distribution and accessible inlet conditions, without measurements on the target cross-section.

The input comprises the axial heat-source distribution
\begin{equation}
q(z)=A\sin\left(\frac{\pi z}{H}\right),
\qquad z\in[0,H],
\end{equation}
together with the scalar inlet temperature $T_{\mathrm{in}}$ and inlet velocity $v_{\mathrm{in}}$ (Fig.~\ref{fig:results1}f). The heat-source amplitude $A$ controls the magnitude of the distributed thermal forcing, while the domain height is fixed at $H=800$\,mm. The resulting mixed functional--parametric input space is $
\mathcal{U}_{\mathrm{sub}}
=
\left\{
(q,T_{\mathrm{in}},v_{\mathrm{in}})
:
q\in L^2([0,H]),
\;
T_{\mathrm{in}},v_{\mathrm{in}}\in\mathbb{R}
\right\}$.
The operating envelope is generated by independently sampling $
A\sim\mathcal{U}(540,660)\;[\mathrm{kW/m^2}]$, $
T_{\mathrm{in}}\sim\mathcal{U}(536.4,655.6)\;[\mathrm{K}]$,and
$v_{\mathrm{in}}\sim\mathcal{U}(4.05,4.95)\;[\mathrm{m/s}]$.
The response is evaluated on a transverse plane
$\mathcal{Y}_{\mathrm{sub}}\subset\mathbb{R}^2$ at a fixed axial position $z=z_0$. The reconstruction targets are the temperature $T(x,y)$, turbulent kinetic energy $k(x,y)$, and velocity magnitude $\|\boldsymbol{v}(x,y)\|$, giving $
\mathcal{S}_{\mathrm{sub}}
=
\left(L^2(\mathcal{Y}_{\mathrm{sub}})\right)^3$.
The corresponding operator is
\begin{equation}
\mathcal{G}_{\mathrm{sub}}
(q,T_{\mathrm{in}},v_{\mathrm{in}})(x,y)
=
\begin{bmatrix}
T(x,y)\\[3pt]
k(x,y)\\[3pt]
\|\boldsymbol{v}(x,y)\|
\end{bmatrix},
\qquad
(x,y)\in\mathcal{Y}_{\mathrm{sub}}.
\end{equation}
This formulation requires the operator to combine distributed heating with inlet-driven advection to recover three coupled fields on a geometrically inaccessible plane.

The dataset contains $5{,}000$ independently generated operating conditions, divided into $3{,}200$ training, $800$ validation, and $1{,}000$ held-out test samples. Each input is paired with the three output fields evaluated on a $1{,}733$-node unstructured mesh, giving a discretized output in $\mathbb{R}^{1733\times3}$. The three-dimensional ANSYS Fluent configuration, governing thermal--hydraulic equations, mesh, turbulence closure, and convergence requirements are described in the Methods and Supplementary Methods.
The subchannel model follows the MIMONet configuration used for the LDC case, with problem-specific modifications for its mixed functional--parametric input. The heat-source profile $q(z)$, discretized at $100$ axial locations, is encoded by a branch network with widths
$[100,512,512,512,256]$,
while the scalar pair $(T_{\mathrm{in}},v_{\mathrm{in}})$ is encoded by a second branch with widths
$[2,512,512,512,256]$.
The two latent representations are fused multiplicatively before coordinate-conditioned decoding. The trunk architecture, $
[2,256,256,256,256\times3]$,
latent width $I=256$, and ReLU activations are unchanged from the LDC configuration. The three decoded outputs are
$\{T,k,\|\boldsymbol{v}\|\}$.
Unless otherwise stated, the optimization and output-scaling procedures are also the same as in the LDC case: training uses an equally weighted mean-squared-error loss, the Adam optimizer with an initial learning rate of $10^{-3}$, a learning-rate reduction factor of $0.5$, at most $500$ epochs, and independent output scaling to $[-1,1]$. For the subchannel case, the weight decay is $10^{-6}$, the learning rate is reduced after ten consecutive epochs without validation-loss improvement, and the batch size is $4$. Unlike the min--max-normalized LDC input, the heat-source and inlet-condition inputs are standardized using means and variances fitted on the subchannel training split.

MIMONet achieves mean relative $\ell_2$ errors of $0.27\%$ for temperature, $2.2\%$ for velocity magnitude, and $4.2\%$ for turbulent kinetic energy, corresponding to a channel-mean error of $2.2\%$ (Fig.~\ref{fig:results1}g; complete per-channel results for all four models in Supplementary Table~S11). It attains the lowest error for every output channel. The corresponding channel-mean errors are $53.0\%$ for KCN, $67.8\%$ for NOMAD, and $72.2\%$ for GeoFNO, making the MIMONet error approximately $24$--$33\times$ lower across the three baselines.
The test-set distributions in Fig.~\ref{fig:results1}g are sharply concentrated at low error. The sample-wise mean distribution is centered near $2.2\%$ with a narrow spread and no pronounced high-error tail, indicating consistent performance across unseen combinations of heat-source amplitude and inlet conditions. The error hierarchy is also physically interpretable: temperature is comparatively smooth and diffusion-regularized, velocity is more sensitive to advective transport and wall confinement, and turbulent kinetic energy contains sharper localized structures and consequently exhibits the largest error.
In the median-error example (Fig.~\ref{fig:results1}h), MIMONet recovers the high-speed interior flow, its attenuation near the no-slip fuel-rod surfaces, the wall-to-core temperature variation, and the localized turbulent-kinetic-energy peaks adjacent to heated boundaries. Residual errors remain low through most of the flow interior and increase primarily near wall-adjacent shear layers and steep thermal gradients, where small spatial discrepancies produce larger pointwise errors. No obvious systematic directional bias is visible in the residual maps.
The co-location of residuals across $T$, $\|\boldsymbol{v}\|$, and $k$ is consistent with the shared latent representation capturing common thermal--hydraulic structures, although it does not establish exact satisfaction of the governing equations. Overall, the subchannel study shows that heterogeneous distributed and scalar inputs can reconstruct coupled fields on an inaccessible cross-section, extending the boundary-only result to a reactor-relevant geometry with multimodal forcing.

\subsection{Pressure--Velocity Reconstruction in a Compact Heat Exchanger}
\label{sec:hx}

The third configuration considers a periodic compact heat exchanger, in which geometric confinement and elevated operating conditions preclude direct instrumentation of the internal pressure and velocity fields~\cite{ahmed2024enhancing}. Compared with the PWR subchannel, this problem introduces a fully three-dimensional velocity field, stronger pressure--velocity coupling, and recirculation generated by an internal fin. The virtual-sensing task is to reconstruct these coupled fields on an inaccessible interior plane using only the applied wall heat flux and the scalar inlet temperature and velocity.

As in the subchannel case, the axial heat-source profile is prescribed as
\begin{equation}
q(z)=A\sin\left(\frac{\pi z}{H}\right),
\qquad z\in[0,H],
\end{equation}
where \(A\) denotes the heat-flux amplitude. This profile is extruded uniformly along the streamwise coordinate \(x\) to define the wall forcing,
\begin{equation}
q_{\mathrm{wall}}(x,z):=q(z),
\qquad x\in[0,L].
\end{equation}
Together with the scalar inlet temperature \(T_{\mathrm{in}}\) and axial velocity magnitude \(u_{\mathrm{in}}\), the wall profile defines the mixed functional--parametric input$
\boldsymbol{u}_{\mathrm{hx}}
=
\left(q_{\mathrm{wall}},T_{\mathrm{in}},u_{\mathrm{in}}\right)
\in\mathcal{U}_{\mathrm{hx}}$.
The response is evaluated on a vertical plane
\(\mathcal{Y}_{\mathrm{hx}}\subset\mathbb{R}^{2}\) at the fixed transverse location \(x=x_{0}\), spanning the \((z,y)\) coordinates. The output comprises the pressure and all three velocity components, $
\mathcal{S}_{\mathrm{hx}}
=
\left(L^{2}(\mathcal{Y}_{\mathrm{hx}})\right)^{4}$,
and the corresponding operator is
\begin{equation}
\mathcal{G}_{\mathrm{hx}}
\left(q_{\mathrm{wall}},T_{\mathrm{in}},u_{\mathrm{in}}\right)(z,y)
=
\begin{bmatrix}
p(z,y)\\[2pt]
u_x(z,y)\\[2pt]
u_y(z,y)\\[2pt]
u_z(z,y)
\end{bmatrix},
\qquad (z,y)\in\mathcal{Y}_{\mathrm{hx}}.
\end{equation}
The model must therefore propagate a distributed thermal input and scalar inlet conditions into a spatially resolved pressure field and three-component velocity field on a domain disjoint from the accessible inputs.

\begin{figure}[t]
    \centering
    \includegraphics[width=\textwidth]{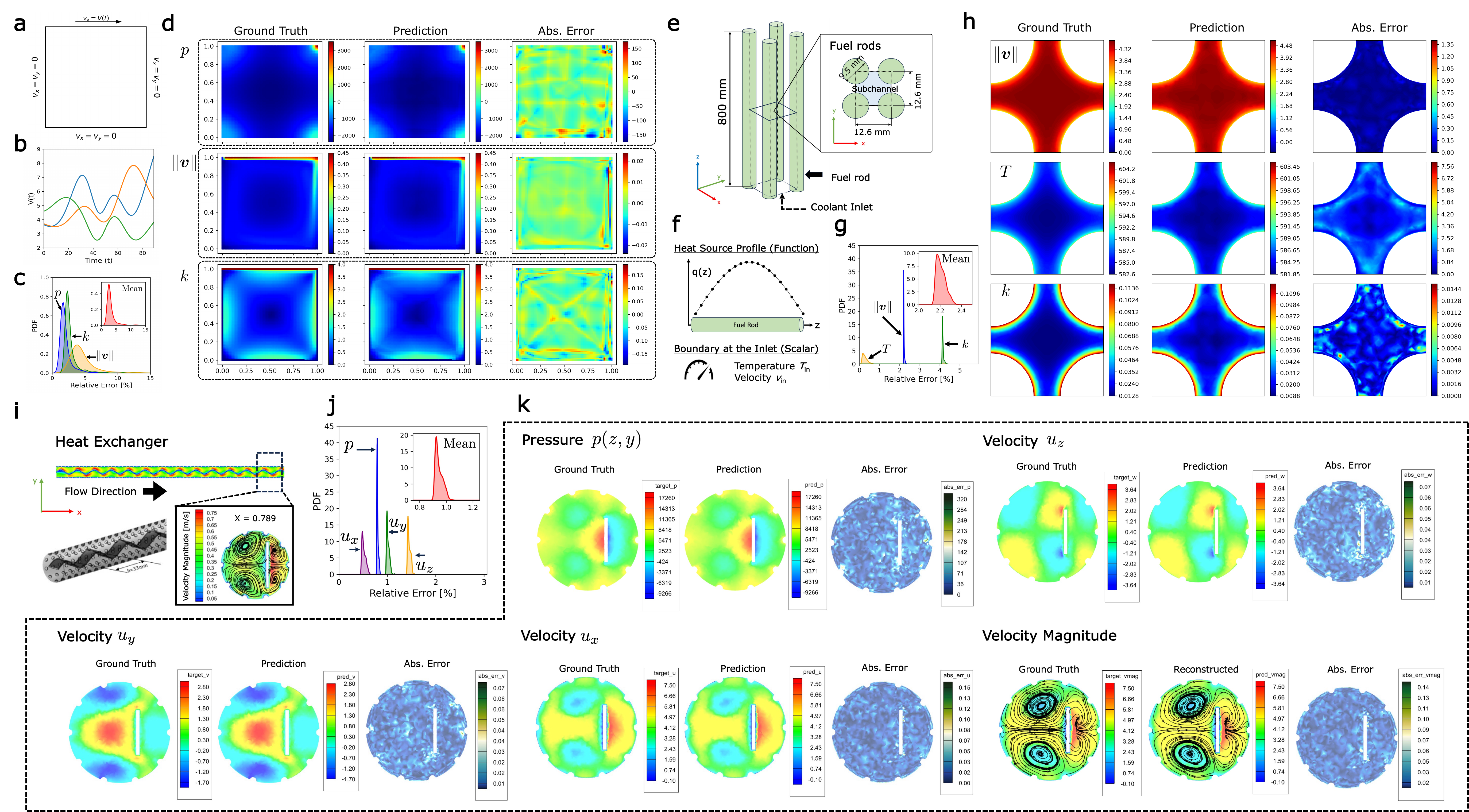}
    \caption
    {
    \textbf{Full-field virtual sensing across three case studies.} \textbf{a}--\textbf{d}, Lid-driven cavity. \textbf{e}--\textbf{h}, PWR subchannel. \textbf{i}--\textbf{k}, Periodic compact heat exchanger. \textbf{a}, Cavity configuration and boundary conditions, with the top wall actuated by the time-dependent lid velocity $V(t)$. \textbf{b}, Representative lid-velocity histories sampled from the dataset. \textbf{e}, Subchannel geometry, fuel rods and transverse reconstruction plane, driven by an axial heat-source profile and scalar inlet conditions. \textbf{f}, Representative axial heat-source functions $q(z)$ and the accessible inlet measurements. \textbf{i}, Heat-exchanger geometry, target plane at $x=0.789$ and representative internal flow structure. \textbf{c}, \textbf{g}, \textbf{j}, Test-set distributions of the relative $\ell_2$ error for each reconstructed field; insets show the distribution of the sample-wise mean across fields. \textbf{d}, \textbf{h}, \textbf{k}, Ground truth, MIMONet prediction and pointwise absolute error for a representative test sample, taken at the median sample-wise mean relative $\ell_2$ error (\textbf{d}, \textbf{h}) and at the lowest mean error (\textbf{k}). \textbf{k} additionally shows the velocity magnitude with superimposed streamlines.
    }
    \label{fig:results1}
\end{figure}

For each operating condition, the converged CFD solution was sampled on the plane \(x=0.789\), producing a two-dimensional unstructured mesh containing \(3{,}977\) nodes. The plane intersects developing-flow regions, near-wall shear layers, and the recirculation zones surrounding the internal fin (Fig.~\ref{fig:results1}i). The dataset comprises \(1{,}546\) simulations generated by varying the wall-flux amplitude, inlet temperature, and inlet axial velocity. Of these, \(1{,}236\) samples were used for training and \(310\) were retained as a held-out test set; a validation subset was drawn from the training data during optimization. Each input is paired with an output array in \(\mathbb{R}^{3977\times4}\) containing \(\{p,u_x,u_y,u_z\}\). The governing equations, turbulence treatment, boundary conditions, and CFD solution procedure are provided in the Methods.
The MIMONet architecture mirrors that used for the PWR subchannel, with the output trunk expanded from three to four channels. The 100-point wall-flux profile is encoded by a functional branch with widths
\([100,512,512,512,256]\), while the inlet pair
\((T_{\mathrm{in}},u_{\mathrm{in}})\) is encoded by a scalar branch
\([2,512,512,512,256]\). Coordinates \((z,y)\) on the target plane are processed by a trunk network
\([2,256,256,256,256\times4]\). The two 256-dimensional branch representations are fused multiplicatively and contracted with the trunk representation to predict \(\{p,u_x,u_y,u_z\}\) simultaneously.
The optimization and normalization procedures were otherwise the same as for the PWR subchannel. The model was trained using an equally weighted mean-squared error over the four output channels and the Adam optimizer, with an initial learning rate of \(10^{-3}\) and a weight decay of \(10^{-6}\). The learning rate was reduced by a factor of \(0.5\) after ten epochs without improvement in the validation loss, and training continued for at most \(500\) epochs with a batch size of \(4\). The functional and scalar inputs were standardized using statistics computed from the training set, and each output channel was independently mapped to \([-1,1]\) using training-set extrema.

The test-error distributions in Fig.~\ref{fig:results1}j are sharply concentrated at low values for all four outputs. The mean relative \(\ell_{2}\) errors are \(0.82\%\) for pressure, \(1.45\%\) for \(u_z\), \(1.02\%\) for \(u_y\), and \(0.52\%\) for \(u_x\), giving a channel-averaged error of \(0.95\%\). The corresponding distribution of sample-wise mean errors is narrow and centered near \(1\%\), indicating consistent performance across the sampled combinations of wall heating and inlet conditions. 
The representative fields in Fig.~\ref{fig:results1}k show that MIMONet recovers the pressure variation and the distinct spatial structures of all three velocity components. Residuals remain small over most of the target plane and increase primarily near the internal fin, geometric junctions, and regions containing steep shear gradients. Their localization in these regions is consistent with the greater spatial complexity and sensitivity of the associated flow structures.
The reconstructed velocity magnitude and streamlines in Fig.~\ref{fig:results1}k preserve the two dominant counter-rotating recirculation cells adjacent to the internal fin, together with the secondary structures near the upper and lower junctions. The predicted vortex cores, separatrices, and stagnation regions closely follow those in the reference solution, while the largest residuals remain localized near the fin and regions of elevated shear. The agreement between the component fields and the reconstructed streamline topology indicates that the low field-wise errors also preserve the principal internal flow organization, which is relevant for identifying maldistribution and flow-induced loading in locations that cannot be instrumented directly.

\subsection{Calibrated Uncertainty for Inaccessible-Field Reconstruction}
\label{sec:uq}

Pointwise reconstruction accuracy does not, by itself, indicate whether an individual field prediction should be trusted. This distinction is particularly important in virtual sensing because the inferred quantities cannot be checked against interior measurements during operation. We therefore augment MIMONet with spatially resolved predictive uncertainty and calibrate the resulting intervals on held-out data.
For the heat-exchanger case, dropout layers with probability \(p=0.10\) were added to every hidden layer of the two branch networks and the trunk network. All other architectural and optimization settings were retained from the deterministic model. The dataset was repartitioned into \(1{,}082\) training samples, \(154\) calibration samples, and the unchanged \(310\)-sample test set; a validation subset was drawn from the training partition. During inference, dropout remains active and \(N=20\) stochastic forward evaluations are performed for each input. For output channel \(c\), the resulting operator realizations are
\begin{equation}
s_{c}^{(e)}(\mathbf{r})
=
\mathcal{G}_{c}^{(e)}(\boldsymbol{u})(\mathbf{r}),
\qquad e=1,\ldots,N,
\end{equation}
from which the predictive mean and variance are estimated as
\begin{align}
\bar{s}_{c}(\mathbf{r})
&=
\frac{1}{N}
\sum_{e=1}^{N}s_{c}^{(e)}(\mathbf{r}),
\\
\sigma_{\mathcal{G},c}^{2}(\mathbf{r})
&=
\frac{1}{N-1}
\sum_{e=1}^{N}
\left[
s_{c}^{(e)}(\mathbf{r})
-
\bar{s}_{c}(\mathbf{r})
\right]^{2}.
\end{align}
The stochastic dispersion \(\sigma_{\mathcal{G},c}(\mathbf{r})\) provides a spatially resolved proxy for model uncertainty. Its magnitude, however, is not intrinsically calibrated and therefore cannot be interpreted directly as a statistically valid prediction interval.
Calibration is performed in two stages. First, normalized residuals are computed on the held-out calibration set,
\begin{equation}
z_{i,c}(\mathbf{r})
=
\frac{
s_{i,c}^{\mathrm{true}}(\mathbf{r})
-
\bar{s}_{i,c}(\mathbf{r})
}{
\sigma_{\mathcal{G},i,c}(\mathbf{r})+\epsilon
},
\qquad i=1,\ldots,N_{\mathrm{cal}},
\end{equation}
where \(\epsilon>0\) prevents numerical instability in regions of vanishing ensemble variance. A channel-specific scale factor, $
\lambda_c
=
\operatorname{std}\!\left(z_{i,c}\right)$
is then used to rescale the predictive standard deviation,
\begin{equation}
\sigma'_{\mathcal{G},i,c}(\mathbf{r})
=
\lambda_c\,
\sigma_{\mathcal{G},i,c}(\mathbf{r}).
\end{equation}

Second, the absolute normalized calibration scores are defined as
\begin{equation}
a_{i,c}(\mathbf{r})
=
\frac{
\left|
s_{i,c}^{\mathrm{true}}(\mathbf{r})
-
\bar{s}_{i,c}(\mathbf{r})
\right|
}{
\sigma'_{\mathcal{G},i,c}(\mathbf{r})+\epsilon
}.
\end{equation}
Let \(q_c^{1-\alpha}\) denote the split-conformal quantile of these scores for channel \(c\), with \(\alpha=0.05\). The resulting \(95\%\) prediction interval is
\begin{equation}
\mathcal{I}_{c}^{95\%}(\mathbf{r})
=
\left[
\bar{s}_{c}(\mathbf{r})
-
q_c^{0.95}\sigma'_{\mathcal{G},c}(\mathbf{r}),
\;
\bar{s}_{c}(\mathbf{r})
+
q_c^{0.95}\sigma'_{\mathcal{G},c}(\mathbf{r})
\right].
\end{equation}
This construction retains the spatial variation supplied by the stochastic operator while using distribution-free conformal calibration to control empirical marginal coverage under exchangeability~\cite{garg2025distributionfree}.

\begin{figure}[t]
    \centering
    \includegraphics[width=\textwidth]{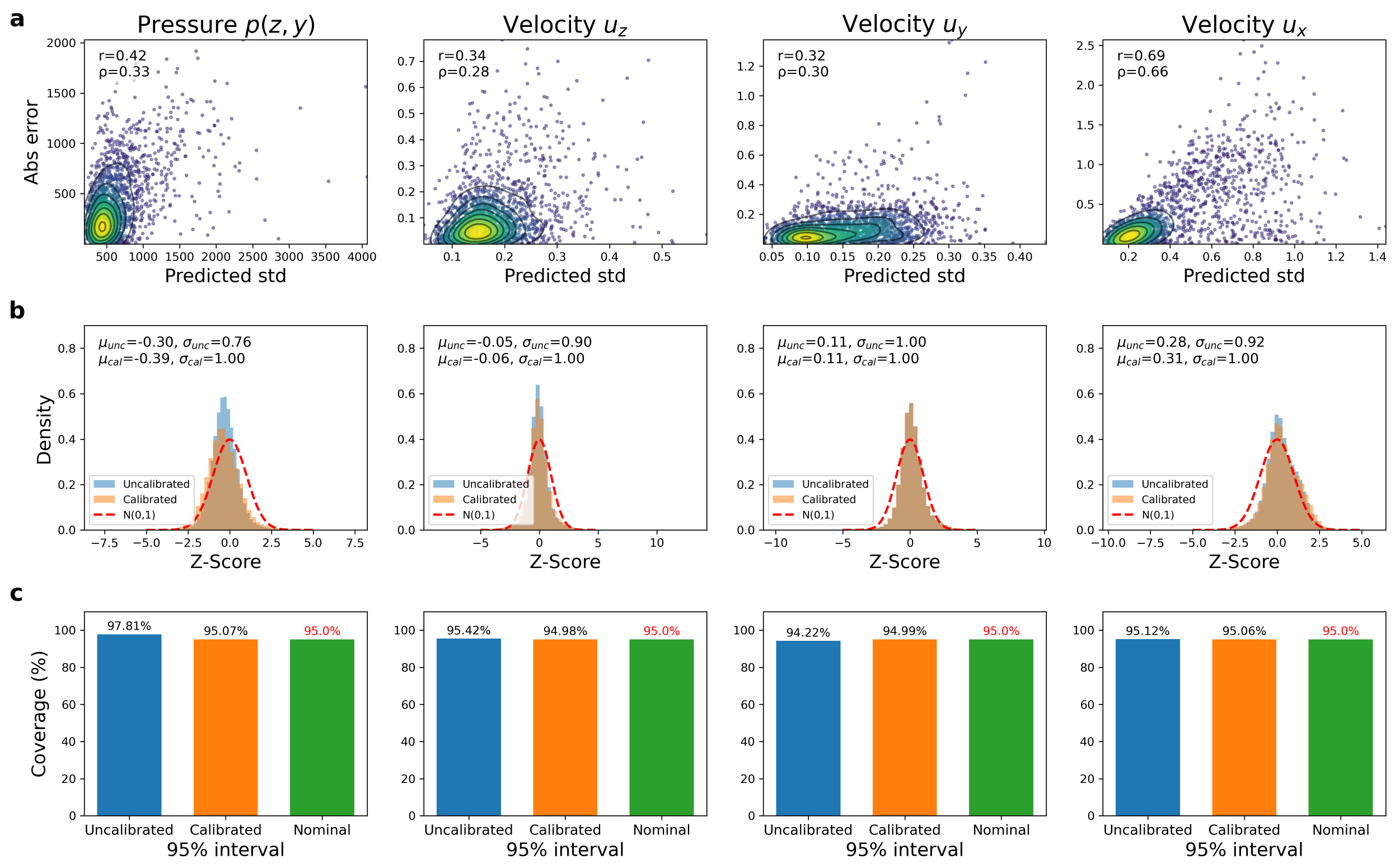}
    \caption{
    \textbf{Uncertainty informativeness and calibration for heat-exchanger field reconstruction.}
    \textbf{a}, Predicted standard deviation versus absolute reconstruction error for pressure and the three velocity components, with density contours. Pearson \(r\) and Spearman \(\rho\) quantify the association between uncertainty magnitude and realized error.
    \textbf{b}, Distributions of normalized residuals before and after calibration, compared with the standard normal density. The uncalibrated \(z\)-score distributions are narrower than \(\mathcal{N}(0,1)\), indicating conservative raw MC-dropout intervals, whereas calibration brings their means and standard deviations close to \(0\) and \(1\), respectively. The normal density is included as a diagnostic reference and is not assumed by the conformal procedure.
    \textbf{c}, Empirical coverage of nominal \(95\%\) prediction intervals before and after calibration. The uncalibrated intervals are conservative on pressure ($97.8\%$) but close to nominal on the velocity channels ($95.4$, $94.2$ and $95.1\%$), so the miscalibration is channel-dependent rather than a uniform excess; conformal calibration brings all four within $0.25$ percentage points of the $95\%$ target.
    }
    \label{fig:uq}
\end{figure}

Figure~\ref{fig:uq}a examines uncertainty informativeness, namely, whether larger stochastic dispersion is associated with larger reconstruction error. Positive associations are observed across the four output channels, with the strongest relationship for \(u_x\) (Pearson \(r=0.69\), Spearman \(\rho=0.66\)). Although the estimated uncertainty does not determine the pointwise error exactly, it provides a meaningful ranking of comparatively reliable and unreliable spatial predictions.
The normalized-residual distributions in Fig.~\ref{fig:uq}b show that the raw MC-dropout uncertainties are miscalibrated by channel rather than uniformly conservative. The normalised residual scale is below unity for pressure ($0.77$), $u_z$ ($0.90$) and $u_x$ ($0.93$), so the predicted standard deviations exceed the residual scale on those channels, but it is $1.00$ for $u_y$, where the raw spread is already correctly scaled. This is consistent with the uncalibrated \(95\%\) intervals over-covering on pressure ($97.8\%$) while slightly under-covering on $u_y$ ($94.2\%$) (Fig.~\ref{fig:uq}c). Channel-wise rescaling and conformal calibration reduce this excess width, bringing the normalized residuals close to unit scale and the empirical coverage close to the nominal \(95\%\) level.
The calibrated intervals therefore satisfy two complementary requirements: their spatial variation is informative about reconstruction difficulty, and their aggregate coverage has a direct empirical interpretation. The calibration result applies to test conditions exchangeable with the calibration data and does not guarantee coverage under arbitrary distribution shift. Within this scope, stochastic operator inference combined with conformal calibration provides quantitatively interpretable confidence bounds for fields reconstructed at locations where direct operational validation is unavailable. Having established both deterministic reconstruction accuracy and calibrated stochastic inference, we next compare MIMONet with alternative architectures in terms of nominal accuracy, model size, and robustness to degraded inputs.

\subsection{Benchmarking Against Operator-Learning Baselines}
\label{sec:benchmarking}

Having established field-reconstruction accuracy across the three physical
configurations and calibrated uncertainty for the heat-exchanger case, we next
compare MIMONet with GeoFNO, NOMAD, and KCN in terms of nominal accuracy,
model size, and robustness to degraded inputs. GeoFNO represents
geometry-aware spectral operator learning, NOMAD provides a
coordinate-conditioned neural-operator baseline, and KCN represents learned
kriging on spatial graphs. All models were evaluated using the same case-wise
training and test partitions, accessible input information, output fields, and
channel-wise relative $\ell_{2}$ metric. Parameter counts vary across cases
because the input branches and output dimensions are problem-dependent.
\begin{figure}[t]
    \centering
    \includegraphics[width=\textwidth]{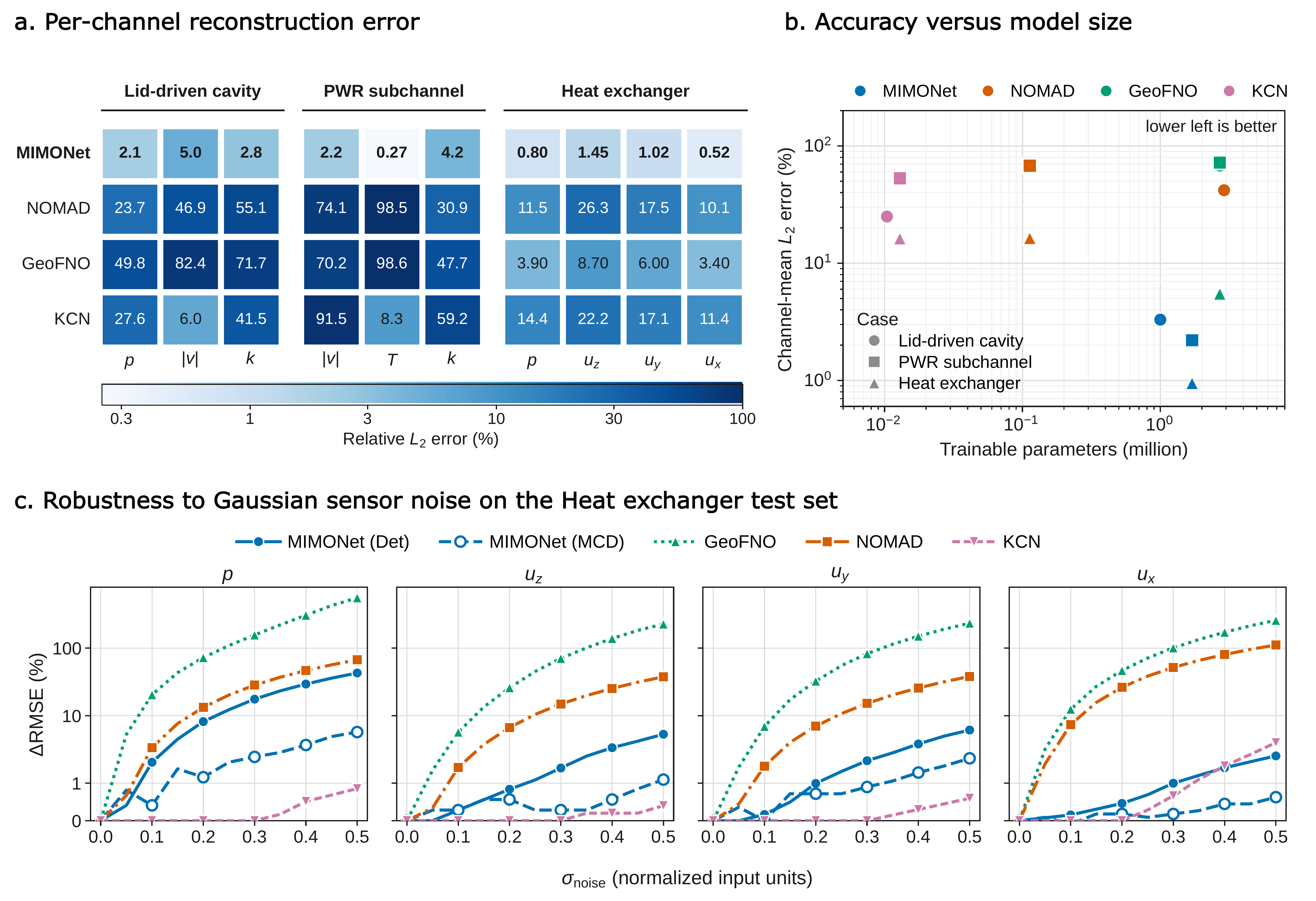}
    \caption{\textbf{Benchmarking MIMONet against neural-operator and learned-interpolation baselines.} \textbf{a}, Mean relative $\ell_2$ error (\%) for every reconstructed field in the lid-driven cavity, PWR subchannel and heat-exchanger test sets. The colour scale is logarithmic. \textbf{b}, Channel-averaged error versus the number of trainable parameters; colour identifies the architecture and marker shape the case study. \textbf{c}, Per-channel relative RMSE inflation over each model's own noise-free baseline, under additive Gaussian noise applied to the normalized heat-exchanger inputs. All baselines use deterministic single-pass inference. MIMONet-MCD is the calibrated dropout-enabled variant of the same model (Section~\ref{sec:uq}), averaging $N=20$ stochastic evaluations, and is shown to quantify the robustness gained through stochastic inference rather than as a separate architecture. The vertical axis is linear below 1\% and logarithmic above.} 
    \label{fig:benchmarking}
\end{figure}

MIMONet attains the lowest error for every reconstructed quantity in all three cases (Fig.~\ref{fig:benchmarking}a). Its channel-averaged errors are \(3.3\%\) for the lid-driven cavity, \(2.2\%\) for the PWR subchannel, and \(0.95\%\) for the heat exchanger. Relative to the strongest baseline in each case, these values correspond to reductions of approximately \(7.6\)-fold, \(24.1\)-fold, and \(5.8\)-fold, respectively. The advantage is not limited to a particular physical quantity: MIMONet remains the most accurate method for pressure, temperature, velocity, and turbulent kinetic energy.
The channel-resolved comparison also reveals an important distinction between isolated and coupled-field accuracy. KCN approaches MIMONet for velocity magnitude in the cavity (\(6.0\%\) compared with \(5.0\%\)) and provides a comparatively low temperature error in the PWR subchannel (\(8.3\%\)). However, its errors increase substantially for the remaining coupled quantities, reaching \(27.6\%\) and \(41.5\%\) for cavity pressure and turbulent kinetic energy and \(91.5\%\) and \(59.2\%\) for subchannel velocity and turbulent kinetic energy. GeoFNO and NOMAD exhibit similarly uneven performance across channels. In contrast, MIMONet maintains low errors across all jointly reconstructed outputs, indicating that its advantage arises from balanced multichannel reconstruction rather than exceptional performance on a single smooth field.

The accuracy differences cannot be explained by model size alone (Fig.~\ref{fig:benchmarking}b). MIMONet uses \(1.0\) million parameters for the cavity and \(1.7\) million for the subchannel and heat exchanger, compared with \(2.7\) million for GeoFNO. Despite its smaller parameter count, MIMONet reduces the channel-averaged error from \(68.0\%\) to \(3.3\%\) in the cavity, from \(72.2\%\) to \(2.2\%\) in the subchannel, and from \(5.5\%\) to \(0.95\%\) in the heat exchanger. Conversely, the substantially smaller NOMAD and KCN models do not provide a competitive accuracy--size trade-off. These results indicate that the performance gain is associated with the multimodal branch encoding, multiplicative latent fusion, and shared multichannel decoding rather than parameter count alone.

To examine whether this advantage persists under degraded measurements, Gaussian perturbations were applied to the concatenated heat-exchanger inputs,
\begin{equation}
\widetilde{\boldsymbol{u}}
=
\boldsymbol{u}+\boldsymbol{\eta},
\qquad
\boldsymbol{\eta}
\sim
\mathcal{N}\!\left(
\boldsymbol{0},
\sigma_{\mathrm{noise}}^{2}\boldsymbol{I}
\right),
\end{equation}
where \(\sigma_{\mathrm{noise}}\in[0,0.5]\) is expressed in normalized input units. Robustness is measured through the channel-wise relative RMSE inflation
\begin{equation}
\Delta\mathrm{RMSE}(\sigma_{\mathrm{noise}})
=
100
\frac{
\mathrm{RMSE}(\sigma_{\mathrm{noise}})
-
\mathrm{RMSE}(0)
}{
\mathrm{RMSE}(0)
}
\;(\%).
\end{equation}
All architectures were evaluated on the same $310$-sample heat-exchanger test
set without retraining on noisy inputs. MIMONet-Det, GeoFNO, NOMAD, and KCN
use one deterministic forward pass for each perturbed input. MIMONet-MCD uses
the $N=20$ stochastic evaluations defined in Section~\ref{sec:uq}. Because the
MCD variant additionally performs posterior averaging, it is reported as an
uncertainty-aware extension rather than as a directly capacity-matched
single-pass baseline.
At $\sigma_{\mathrm{noise}}=0.5$, deterministic MIMONet exhibits a
channel-averaged RMSE inflation of approximately $14\%$, compared with
approximately $64\%$ for NOMAD and $319\%$ for GeoFNO (Fig.~\ref{fig:benchmarking}c). Thus, among the
deterministic architectures that propagate the perturbed branch inputs,
MIMONet provides the lowest error inflation. The calibrated MIMONet-MCD
variant further reduces the channel-averaged inflation to approximately
$2.45\%$, with per-channel values of \(5.71\%\) for pressure and \(1.13\%\), \(2.33\%\), and \(0.63\%\) for \(u_z\), \(u_y\), and \(u_x\), respectively. This additional reduction reflects
the benefit of stochastic averaging rather than a purely architectural
single-pass advantage.
The larger sensitivity of pressure is consistent with its globally coupled character, whereas the velocity components remain comparatively stable under the imposed perturbations.
KCN displays an apparently small channel-averaged inflation of approximately \(1.5\%\), but this behaviour does not imply superior sensing robustness. Its graph and edge weights are constructed from fixed mesh coordinates, so perturbations applied to the functional and scalar branch inputs are largely excluded from its inference pathway. The resulting insensitivity must therefore be considered together with its noise-free heat-exchanger error of \(16.3\%\), compared with \(0.95\%\) for MIMONet.

To complement the finite-noise evaluation, we characterized the local
sensitivity of MIMONet using an empirical Lipschitz proxy. For each output
channel, \(L_{99}\) was defined as the \(99\)th percentile of
\begin{equation}
    \frac{
\|\mathcal{G}_{\theta}(\boldsymbol{u}+\boldsymbol{\delta})
-\mathcal{G}_{\theta}(\boldsymbol{u})\|_{2}
}{
\|\boldsymbol{\delta}\|_{2}
}
\end{equation}
over \(1{,}000\) random perturbation directions at each of \(20\) magnitudes
in normalized input space. The resulting values were approximately
\(0.88\), \(0.37\), \(0.64\), and \(1.02\) for \(p\), \(u_z\), \(u_y\), and
\(u_x\), respectively. Thus, no large local amplification is observed over
the sampled heat-exchanger test distribution, consistent with the controlled
error growth under noisy inputs in Fig.~\ref{fig:benchmarking}c. These values
are empirical quantiles over sampled inputs and perturbation directions, not
global Lipschitz constants or formal stability guarantees; the complete
sensitivity curves are provided in Supplementary Material. Taken together, the nominal and perturbed-input evaluations show that MIMONet combines consistently low multichannel reconstruction error with stable propagation of noisy boundary information into inaccessible interior fields.

\subsection{Local Out-of-Distribution Generalization}
\label{sec:ood}

A virtual sensor may encounter operating conditions outside the range represented in its training data. We therefore evaluated whether the learned operator degrades gradually under moderate parameter extrapolation. For the heat-exchanger configuration, \(50\) additional ANSYS Fluent simulations were generated with the inlet velocity \(u_{\mathrm{in}}\), inlet temperature \(T_{\mathrm{in}}\), and wall heat-flux amplitude placed approximately \(15\%\) outside their respective training ranges. Twenty-five cases were sampled below the lower training boundary and \(25\) above the upper boundary. The geometry, solver configuration, convergence criteria, data-processing pipeline, and target plane at \(x=0.789\) were unchanged. All trained models were evaluated on these cases without retraining or fine-tuning.
\begin{figure}[t]
    \centering
    \includegraphics[width=\textwidth]{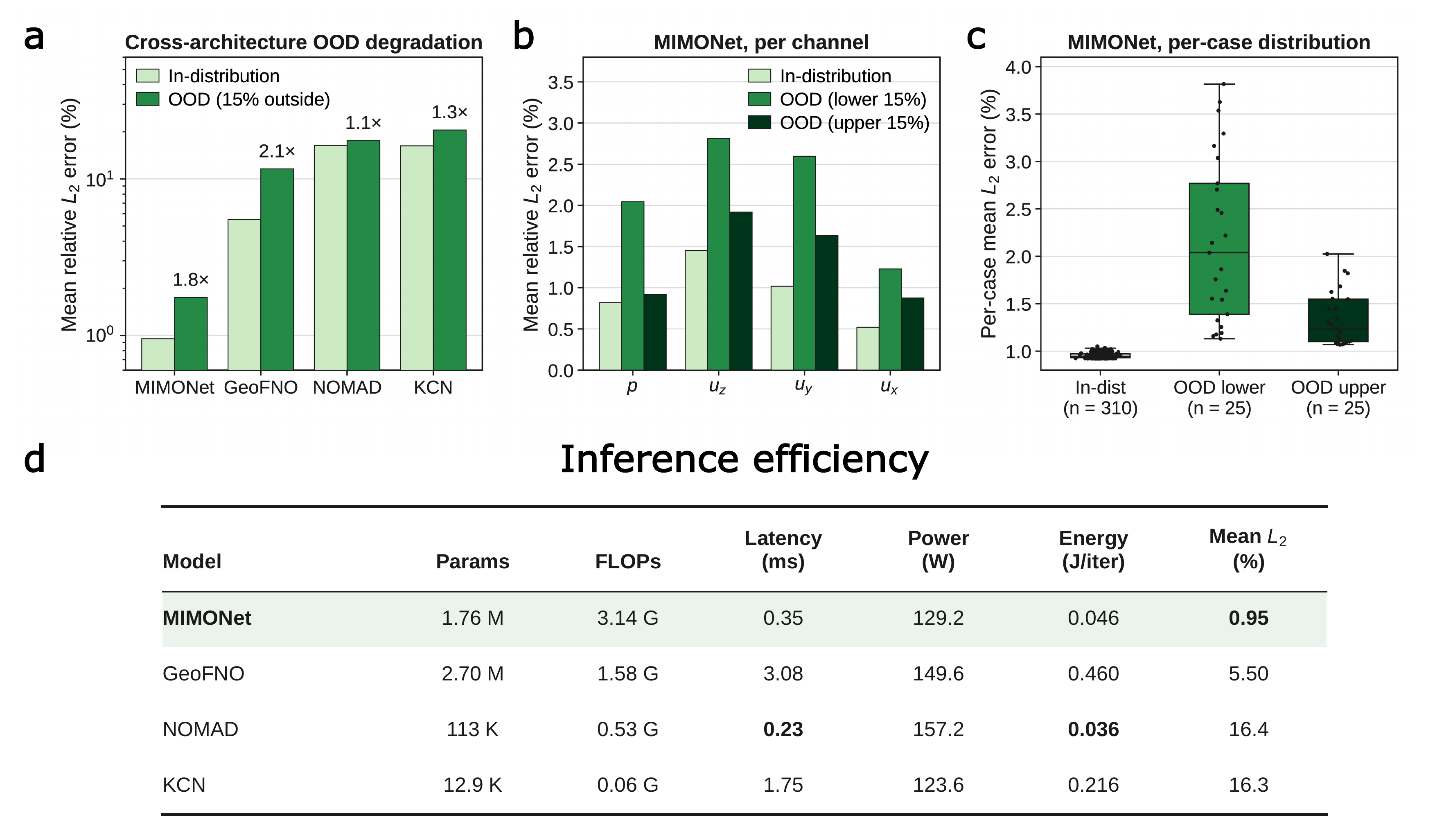}
    \caption{\textbf{Local out-of-distribution generalization and inference efficiency on the heat exchanger.} \textbf{a}--\textbf{c}, MIMONet and the operator-learning baselines evaluated on 50 additional CFD cases with operating parameters approximately 15\% outside the training ranges (25 below and 25 above the training boundaries). \textbf{a}, Channel-averaged relative $L_2$ error under in-distribution and out-of-distribution conditions; labels give the OOD-to-ID error ratio. Note the logarithmic axis. \textbf{b}, Per-channel MIMONet errors for the in-distribution test set and for operating conditions below and above the training boundaries. \textbf{c}, Distribution of sample-wise MIMONet errors, with individual cases superimposed. Boxes show the median and interquartile range and whiskers the full range; the largest sample-wise OOD error is 3.82\%. \textbf{d}, Inference efficiency, profiled on a single NVIDIA H200 GPU with $B=1$, $P=3{,}977$ query points, $1{,}000$ forward evaluations and float32 precision. Energy is calculated from total board power without idle-power subtraction. Bold entries mark the lowest latency, energy and reconstruction error.} 
    \label{fig:ood_generalization}
\end{figure}

MIMONet’s channel-averaged relative \(\ell_{2}\) error increases from \(0.95\%\) on the in-distribution test set to \(1.75\%\) on the OOD cases, corresponding to a \(1.8\)-fold increase (Fig.~\ref{fig:ood_generalization}a). The OOD errors are \(1.48\%\) for pressure, \(2.37\%\) for \(u_z\), \(2.11\%\) for \(u_y\), and \(1.05\%\) for \(u_x\) (Fig.~\ref{fig:ood_generalization}b). Cases below the lower training boundary exhibit a larger channel-averaged error (\(2.17\%\)) than cases above the upper boundary (\(1.34\%\)), indicating asymmetric sensitivity to the direction of the parameter shift.
MIMONet also retains the lowest OOD error among the evaluated architectures. GeoFNO increases from \(5.5\%\) to \(11.6\%\), NOMAD from \(16.4\%\) to \(17.6\%\), and KCN from \(16.3\%\) to \(20.5\%\). MIMONet’s OOD error is therefore approximately \(6.6\)--\(11.7\) times lower than those of the three baselines. The relative degradation factors differ across architectures: GeoFNO exhibits the largest proportional increase, whereas NOMAD changes comparatively little but starts from a substantially higher in-distribution error.

The sample-wise distribution in Fig.~\ref{fig:ood_generalization}c remains compact under the tested parameter shift. The largest OOD error is \(3.82\%\), compared with \(1.05\%\) for the largest in-distribution error. Thus, the extrapolation cases increase both the typical error and its spread, but no abrupt loss of accuracy is observed within the evaluated \(15\%\) range.
These results provide evidence of local parameter-extrapolation robustness under a fixed geometry, CFD solver, turbulence model, and preprocessing pipeline. They do not establish generalization to new geometries, different physical models, substantially larger parameter shifts, experimental data, or independently generated solver datasets. Cross-solver and experimental validation would therefore be required before interpreting this result as broad out-of-distribution generalization.

\subsection{Real-Time Inference and Energy Cost}
\label{sec:latency}

The operational value of virtual sensing depends not only on reconstruction accuracy but also on whether complete field estimates can be produced within the monitoring latency budget. We therefore profiled MIMONet, GeoFNO, NOMAD, and KCN on the heat-exchanger task using a single NVIDIA H200 GPU under an identical workload: batch size \(B=1\), \(P=3{,}977\) query points, \(1{,}000\) forward evaluations, and float32 precision. Ten warm-up evaluations were performed before measurement.
Floating-point operations were measured using \texttt{torch.profiler}, and latency was recorded using \texttt{time.perf\_counter} with \texttt{torch.cuda.synchronize} before and after each timed evaluation. Energy per inference was estimated as, $
E_i=P_i t_i$,
where \(P_i\) is the mean NVML-reported board power and \(t_i\) is the measured inference latency. Because idle-board power was not subtracted, the energy values should be interpreted as workload-level comparative measurements on the same GPU rather than isolated model energy consumption. Reconstruction errors were evaluated on the same \(310\)-sample heat-exchanger test set used in Section~\ref{sec:hx}.

MIMONet reconstructs all four fields at \(3{,}977\) query points in \(0.35\,\mathrm{ms}\), with an estimated energy cost of \(46\,\mathrm{mJ}\) per inference and a channel-averaged error of \(0.95\%\) (Fig.~\ref{fig:ood_generalization}d). This latency is substantially below the millisecond-scale monitoring target considered here and is more than five orders of magnitude lower than the hundreds of seconds required by the corresponding high-fidelity CFD solution. The comparison concerns inference after offline training and does not include sensor acquisition, data transfer, or preprocessing latency.
GeoFNO is the closest baseline in reconstruction accuracy, with a mean error of \(5.5\%\), but requires \(3.08\,\mathrm{ms}\) and \(460\,\mathrm{mJ}\) per inference. Relative to GeoFNO, MIMONet is approximately \(8.8\) times faster and \(10\) times less energy-intensive while reducing the mean error by a factor of \(5.8\). The difference between the measured latency and reported FLOP counts also shows that FLOPs alone do not determine execution time; memory access, kernel structure, and hardware utilization contribute to the observed architecture-level performance.
NOMAD attains the lowest latency (\(0.23\,\mathrm{ms}\)) and energy cost (\(36\,\mathrm{mJ}\)) but has a mean reconstruction error of \(16.4\%\), approximately \(17\) times that of MIMONet. It therefore represents a different operating point on the accuracy--efficiency trade-off rather than being uniformly dominated by MIMONet. KCN combines the smallest parameter count and FLOP estimate with a latency of \(1.75\,\mathrm{ms}\), an energy cost of \(216\,\mathrm{mJ}\), and an error of \(16.3\%\), illustrating that parameter count and nominal FLOPs do not translate directly into end-to-end inference efficiency.

The branch--trunk contraction used by MIMONet scales linearly with the number of query points, \(\mathcal{O}(BPIm)\), where \(I\) is the latent width and \(m\) the number of output channels. This scaling supports dense field reconstruction while retaining sub-millisecond latency for the evaluated heat-exchanger mesh. Complementary profiling across NVIDIA A40, A100, H200, and GH200 GPUs is reported in Supplementary Fig.~S9 and Supplementary Table~S23, where MIMONet remains below \(1\,\mathrm{ms}\) on the A40, A100 and H200 under the same workload; the slowest measurement anywhere is \(1.17\,\mathrm{ms}\) on the GH200.
The profiling results establish real-time computational feasibility for the evaluated hardware and query resolution. They should not be interpreted as a hardware-independent latency guarantee, because total deployment latency will additionally depend on sensor communication, preprocessing, batching, device scheduling, and the target computing platform.

\section{Validation Beyond Solver-Generated Data}
\label{sec:realworld}

The simulation-based studies establish that operator-based virtual sensing can
reconstruct coupled fields under controlled, physically demanding conditions,
with the reactor subchannel as the safety-critical case. We next evaluate
whether the same framework remains effective beyond solver-generated training
data, using two measured datasets and one observation-constrained reanalysis
product. These datasets provide interior
ground truth that can be withheld during evaluation, enabling direct comparison
with practical interpolation, extrapolation, and learned sensing baselines.
The three tasks progressively reduce direct observational support for the
target. In the fuel-cell case, sparse measurements of both target quantities
are available, but most interior locations are withheld. In the wind-tower
case, the target lies above the highest input sensor and must be inferred by
vertical extrapolation. In the ocean case, temperature and salinity are
observed on disjoint sparse networks, whereas sea-surface height has no direct
input observations and must be recovered through cross-modal coupling. This
progression tests sparse interpolation, physical extrapolation, and zero-sensor
cross-modal reconstruction within a common operator framework.

\subsection{Sparse Internal-Field Reconstruction in a Fuel Cell}
\label{sec:fuelcell}

A proton-exchange-membrane fuel cell (PEMFC) develops a spatially varying current-density field that governs local power generation, heat production, and degradation. Complete current-density maps can be obtained experimentally using a current-density-mapping (CDM) plate inserted between the cell layers, but this intrusive arrangement is unsuitable for routine operation. We therefore test whether the internal current-density and temperature fields can be reconstructed from measurements retained at only a small subset of spatial locations.
We use the publicly available PEMFC dataset from the University of Seville~\cite{suarez2023cdm}. The experiments were conducted on an approximately \(50\,\mathrm{cm}^{2}\) parallel-serpentine cell under eight operating protocols comprising four gas inlet--outlet configurations and two load profiles: a polarization curve and a dynamic load cycle. Current density \(J\) was recorded on an \(18\times18\) grid containing \(324\) locations, while temperature \(T\) was measured on a \(9\times9\) grid, both at \(1\,\mathrm{Hz}\). Approximately \(8{,}000\) measured states are retained. The temperature measurements are bilinearly mapped to the \(18\times18\) current-density grid, and the states are partitioned into \(70\%\) training, \(15\%\) validation, and \(15\%\) test samples by a uniform random permutation of frame index. This is a random split rather than a chronological one, so a test frame can be temporally adjacent to a training frame; the reported errors accordingly characterise sparse-sensor reconstruction on operating states close to those seen in training, and are not a measure of temporal extrapolation. The protocol and its consequences are set out in Supplementary Material.

Only \(32\) of the \(324\) grid locations are supplied to the model, corresponding to approximately \(10\%\) spatial coverage. Let
\(\mathcal{S}_{\mathrm{obs}}\subset\Omega_{\mathrm{grid}}\)
denote this set of observed locations, with
\(|\mathcal{S}_{\mathrm{obs}}|=32\). The virtual-sensing operator is
\begin{equation}
\mathcal{G}_{\mathrm{fc}}
\left(
\left\{
J_i,T_i,\mathbf{r}_i
\right\}_{\mathbf{r}_i\in\mathcal{S}_{\mathrm{obs}}}
\right)(\mathbf{r})
=
\begin{bmatrix}
J(\mathbf{r})\\
T(\mathbf{r})
\end{bmatrix},
\qquad
\mathbf{r}\in\Omega_{\mathrm{grid}}.
\end{equation}
The complete fields are retained as reconstruction targets, but the remaining \(292\) locations receive no input observations. Because the sparse inputs are sampled from laboratory CDM maps, this experiment evaluates sparse reconstruction using measured physical data; it does not imply that the complete CDM apparatus is itself deployable.

A two-branch MIMONet is used. The first branch encodes the \(32\) observed current-density values together with Fourier representations of their coordinates, while the second encodes the corresponding temperature observations and coordinates. Their latent representations are fused multiplicatively, and a Fourier-feature trunk jointly decodes current density and temperature at arbitrary grid coordinates.
MIMONet is evaluated against KCN and the classical IDW and \(k\)-nearest-neighbour (\(k\)NN) interpolants using identical sensor locations and data splits. KCN accepts scattered observations directly and provides a learned locality-based control. FNO is not treated as a native baseline because its spectral layers require a gridded input field; applying it to the \(32\) scattered observations would require a separate interpolation front end that is itself part of the sensing pipeline. Complete five-seed numerical results are reported in Supplementary Material.

Across five random seeds, MIMONet reconstructs the two-channel field with a channel-averaged relative \(\ell_{2}\) error of
\(1.90\pm0.51\%\), compared with \(6.17\pm1.96\%\) for KCN,
\(11.67\pm0.81\%\) for IDW, and \(11.99\pm1.03\%\) for \(k\)NN. Relative to IDW, the strongest classical baseline, this corresponds to an \(84\%\) reduction in channel-averaged error.
The current-density field, the principal inaccessible quantity, is reconstructed with an error of \(3.77\pm1.02\%\), compared with
\(12.29\pm3.93\%\) for KCN, \(23.10\pm1.60\%\) for IDW, and
\(23.74\pm2.04\%\) for \(k\)NN. Temperature is considerably smoother and is reconstructed with an error of \(0.04\pm0.00\%\). This exceptionally small temperature error should be interpreted in light of the original \(9\times9\) measurements being bilinearly mapped to the finer grid; current density therefore provides the more discriminating assessment of sparse reconstruction.
The representative result in Fig.~\ref{fig:real-world}a--f recovers the high-current peripheral region and the lower-current structure associated with the serpentine flow path. Residuals are concentrated near channel bends and regions containing sharper current-density gradients, whereas the smoother temperature field is reproduced almost uniformly. The measured-data experiment therefore shows that MIMONet reconstructs sparse internal energy-device fields substantially more accurately than classical interpolation and locality-based learned estimation.

\subsection{Vertical Extrapolation Beyond Wind-Tower Sensors}
\label{sec:wind}

The second measured-data task introduces a different form of inaccessibility: the target lies above the highest available sensor. Wind-energy assessment requires wind speed near turbine hub height, typically around \(80\,\mathrm{m}\), whereas operational monitoring may rely on measurements from lower levels. Standard engineering methods use a power law or logarithmic profile for this extrapolation, but their fixed-form assumptions can become inaccurate under atmospheric stratification, low-level jets, and site-dependent flow conditions. We therefore test whether an operator can infer the vertical profile above the instrumented range from lower-level measurements and contextual information.
We use the BSC/IEA-Wind Tall Tower Dataset~\cite{ramon2020iea}, a quality-controlled collection of in-situ meteorological-mast measurements. Seven towers containing near-surface measurements and instrumentation extending to at least \(80\,\mathrm{m}\) are retained: Boseong (KR), Lindenberg (DE), the NWTC M2, M4, and M5 towers (US-CO), Hegyh\'ats\'al (HU), and Wallaby Creek (AU). The measurements are resampled to hourly resolution, yielding approximately \(6\times10^{5}\) vertical profiles spanning five climatic regions. Data from each tower are partitioned chronologically into \(70\%\) training, \(15\%\) validation, and \(15\%\) test samples.
Only wind speeds at \(10\), \(25\), and \(50\,\mathrm{m}\) are supplied as measurements. The operator reconstructs wind speed at query heights above the highest input sensor:
\begin{equation}
\mathcal{G}_{\mathrm{wind}}
\left(
\{U(z_i),z_i\}_{z_i\in\{10,25,50\}\,\mathrm{m}},
\mathbf{c}
\right)(z)
=
U(z),
\qquad z>50\,\mathrm{m},
\end{equation}
where
\(\mathbf{c}=[\mathrm{lat},\mathrm{lon},\sin h,\cos h]\)
contains the tower location and hour-of-day encoding. Performance is evaluated at \(60\), \(70\), \(80\), and \(100\,\mathrm{m}\); the \(100\,\mathrm{m}\) evaluation uses the five towers with measurements reaching that height. Because no measurements above \(50\,\mathrm{m}\) are included in the input, evaluation at these heights constitutes vertical extrapolation beyond the instrumented range.

A two-branch MIMONet is used. The first branch encodes the three lower-level wind-speed measurements together with their heights, while the second encodes the site and diurnal context vector \(\mathbf{c}\). These representations are fused multiplicatively, and a Fourier-feature trunk encodes the query height \(z\). The context branch allows the operator to represent site- and time-dependent variations in vertical profile shape, although it does not explicitly measure atmospheric stability.
MIMONet is compared with FNO, KCN, the power law, a two-point log-height extrapolation, and a nearest-sensor baseline that returns the wind speed measured at \(50\,\mathrm{m}\). Unlike the scattered-input fuel-cell case, FNO is directly applicable because the vertical wind profile forms an ordered one-dimensional sequence. The learned models are evaluated using the same chronological splits. They are not capacity-matched: MIMONet carries $207{,}617$ parameters against $68{,}289$ for FNO and $66{,}818$ for KCN, so the two baselines are matched to each other while MIMONet is roughly three times larger. Complete results at all four query heights are reported in Supplementary Material.

At hub height (\(80\,\mathrm{m}\)), MIMONet achieves a relative \(\ell_{2}\) error of \(4.71\pm0.15\%\), compared with \(7.30\pm1.16\%\) for FNO, \(25.28\pm0.57\%\) for KCN, \(27.86\%\) for the nearest-sensor baseline, \(28.95\%\) for the two-point log-height extrapolation, and \(34.30\%\) for the power law. Relative to the nearest-sensor estimate, the strongest classical baseline at this height, MIMONet reduces the error by \(83\%\).
The error increases with extrapolation distance for every method. At \(100\,\mathrm{m}\), MIMONet attains \(14.70\pm0.59\%\), compared with \(18.88\pm1.00\%\) for FNO, \(43.06\pm1.14\%\) for KCN, \(47.73\%\) for the nearest-sensor estimate, \(52.49\%\) for the two-point log-height extrapolation, and \(95.92\%\) for the power law. The degradation of the fixed-form profiles reflects the application of idealized atmospheric-profile assumptions to heterogeneous measured conditions in which those assumptions need not hold.

The representative profiles in Fig.~\ref{fig:real-world}g,~h show that MIMONet follows variations in measured profile shape within the uninstrumented region, including cases where the profile flattens or accelerates differently from the fitted power law. FNO is the strongest comparison method, confirming that operator learning is also effective when the input has a natural one-dimensional structure. MIMONet nevertheless retains lower error at every evaluated height (Fig.~\ref{fig:real-world}i), consistent with the benefit of jointly encoding the measured profile, site information, diurnal context, and continuous query coordinate. The results extend virtual sensing from spatial interpolation within an instrumented domain to reconstruction beyond the physical range of the available sensors.

\subsection{Cross-Modal Ocean-State Reconstruction from Disjoint Observations}
\label{sec:sst}

The third task combines disjoint sensing, coupled multi-field reconstruction, and a target variable with no direct input observations. Temperature and salinity are supplied at different spatial locations, while sea-surface height (SSH) is withheld completely from the operator input. The model must therefore reconstruct the full temperature, salinity, and SSH fields from two sparse, non-overlapping observation networks. Unlike the preceding measurement-based cases, this experiment uses an observation-constrained ocean reanalysis and tests cross-modal reconstruction under a controlled sensor-withholding protocol.

We use the GLORYS12V1 global ocean reanalysis~\cite{lellouche2021glorys}, a \(1/12^{\circ}\)-resolution data-assimilative product distributed by the Copernicus Marine Service. GLORYS12 combines an ocean circulation model with in-situ and satellite observations, including Argo profiles, satellite altimetry, and sea-surface-temperature retrievals. We extract potential temperature \(T\), salinity \(S\), and SSH over the North Atlantic
(\(35\)--\(50^{\circ}\mathrm{N}\), \(60\)--\(20^{\circ}\mathrm{W}\)) and block-average the fields to approximately \(0.25^{\circ}\) resolution. The resulting dataset contains $730$ daily snapshots over $9{,}457$ ocean cells, partitioned into $610$ training, $60$ validation and $60$ test days by a random permutation of days. Because daily ocean fields are strongly autocorrelated, this protocol places most test days adjacent to training days; a forward-in-time split is therefore reported as a robustness check in Supplementary Material.

Two fixed, disjoint sensor networks are constructed: \(400\) temperature locations
\(\mathcal{S}_{T}\) and \(400\) salinity locations
\(\mathcal{S}_{S}\), with
\(\mathcal{S}_{T}\cap\mathcal{S}_{S}=\emptyset\). No SSH values are supplied to the model. For day \(d\), the cross-modal reconstruction operator is
\begin{equation}
\mathcal{G}_{\mathrm{ocean}}
\left(
\{T_i^{(d)},\mathbf{r}_i\}_{\mathbf{r}_i\in\mathcal{S}_{T}},
\{S_j^{(d)},\mathbf{r}_j\}_{\mathbf{r}_j\in\mathcal{S}_{S}}
\right)(\mathbf{r})
=
\begin{bmatrix}
T^{(d)}(\mathbf{r})\\
S^{(d)}(\mathbf{r})\\
\mathrm{SSH}^{(d)}(\mathbf{r})
\end{bmatrix},
\qquad
\mathbf{r}\in\Omega_{\mathrm{ocean}}.
\end{equation}
Temperature and salinity must therefore be reconstructed beyond their respective sparse networks, while SSH must be inferred entirely from relationships learned between the input \(T\) and \(S\) fields and the SSH target represented in the training data. Although SSH has zero coverage in the operator input, the GLORYS12 reference field itself is influenced by assimilated altimetry and should not be interpreted as independent in-situ ground truth.

For this case, we use MIMONet-NL, a nonlinear-decoder variant in which an MLP operating on the concatenated trunk and fused branch representations replaces the linear contraction used in the preceding cases. One branch encodes the \(400\) temperature measurements and their Fourier-encoded coordinates, while the second encodes the \(400\) salinity measurements at the disjoint locations. The branch representations are fused multiplicatively, and a Fourier-feature trunk provides the query-coordinate representation for all three output fields. Temperature and salinity are learned as residual corrections to an IDW estimate, whereas SSH is predicted without an IDW prior because no SSH observations are supplied.
MIMONet-NL is compared with KCN, IDW, \(k\)NN, and climatology. KCN accepts the two scattered networks directly and provides a learned locality-based comparison. IDW and \(k\)NN reconstruct the observed \(T\) and \(S\) channels independently. Because neither method can interpolate a variable with no observations, their SSH estimates are obtained by ridge regression from their reconstructed temperature and salinity fields. Climatology returns the grid-wise training mean and provides the strongest classical SSH predictor.

FNO is not treated as a native end-to-end baseline because the two scattered observation networks must first be mapped to a regular grid. As a reference, an IDW-gridded, capacity-matched FNO reconstructs all three channels with a channel-mean error of approximately \(4.4\%\) (\(3.10\%\) for \(T\), \(0.47\%\) for \(S\) and \(9.58\%\) for SSH), close to MIMONet-NL on every channel. The rasterisation front end, not the accuracy, is why it is not treated as a native baseline here: it requires the scattered observations to be interpolated onto a regular grid before the operator sees them, which reintroduces exactly the spatial-interpolation step the operator formulation avoids. Complete five-seed comparisons are reported in Supplementary Material.

\begin{figure}[htbp]
\centering
\includegraphics[width=\textwidth]{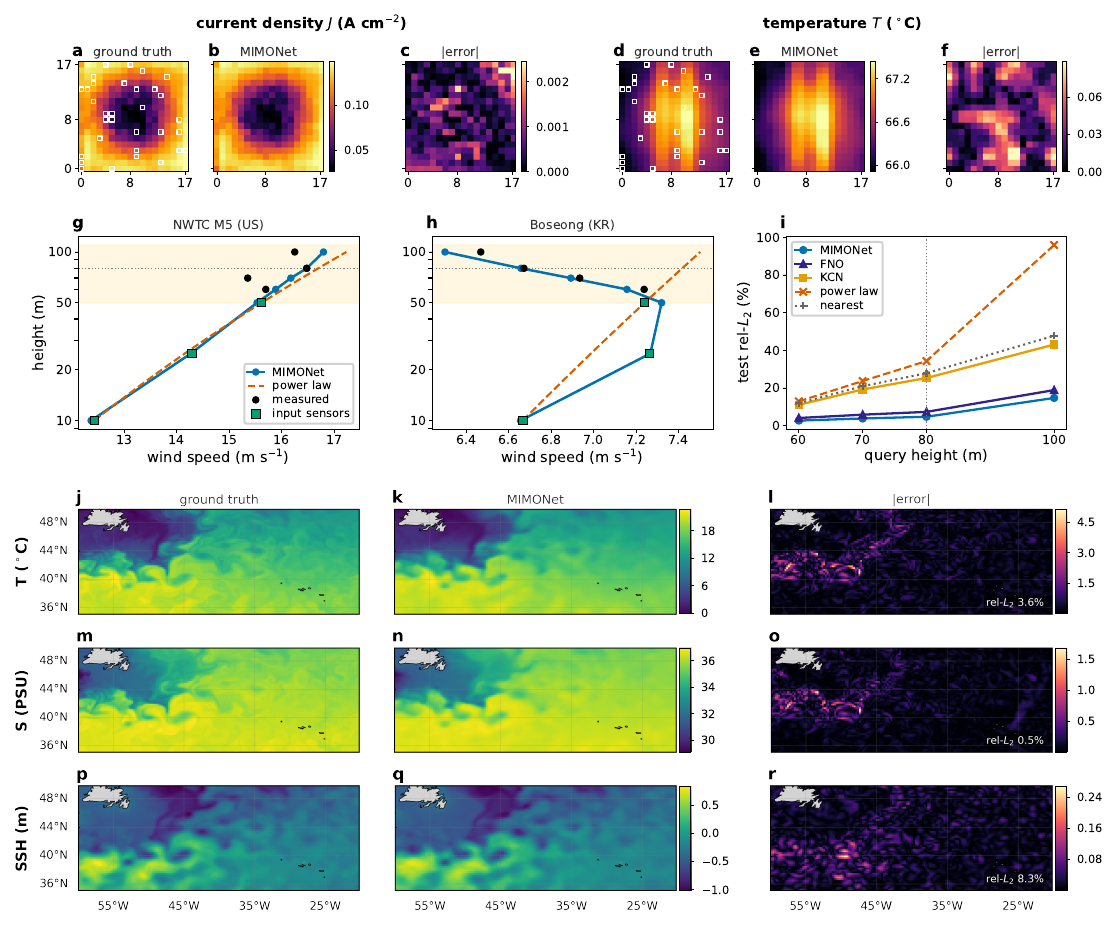}
\caption{\textbf{Operator-based virtual sensing on three real-world datasets.} Each block reconstructs a field that no deployable instrument returns, from sparse observations of other quantities. \textbf{a}--\textbf{f}, Proton-exchange-membrane fuel cell (Problem~4): current density $J$ (\textbf{a}--\textbf{c}) and temperature $T$ (\textbf{d}--\textbf{f}) for a held-out operating state, showing ground truth (\textbf{a},~\textbf{d}), MIMONet reconstruction (\textbf{b},~\textbf{e}) and absolute error (\textbf{c},~\textbf{f}). White squares mark the 32 of 324 observed segments of the $18\times18$ mapping grid; ground truth and reconstruction share a colour scale per channel. \textbf{g}--\textbf{i}, IEA-Wind tall towers (Problem~5). \textbf{g}, \textbf{h}, Vertical profiles on held-out timestamps: input sensors at $10/25/50\,$m (green), measurements (black), MIMONet (blue) and a power law fitted to the lowest and top sensors (dashed). The shaded band carries no input sensor. \textbf{i}, Test relative $L_2$ error against query height, five-seed mean; dotted line, hub height. \textbf{j}--\textbf{r}, GLORYS12 North Atlantic (Problem~6): temperature $T$ (\textbf{j}--\textbf{l}), salinity $S$ (\textbf{m}--\textbf{o}) and sea-surface height (\textbf{p}--\textbf{r}) on a held-out day, from 400 temperature and 400 salinity sensors at disjoint locations. Ground truth (\textbf{j},~\textbf{m},~\textbf{p}) and reconstruction (\textbf{k},~\textbf{n},~\textbf{q}) share a colour scale; error panels (\textbf{l},~\textbf{o},~\textbf{r}) are inset with the per-field relative $L_2$ error. Sea-surface height is unobserved in the input and is recovered from the learned temperature--salinity coupling. Land, grey.
}
\label{fig:real-world}
\end{figure}

Across five seeds, MIMONet-NL reconstructs the three fields with a channel-averaged relative \(\ell_{2}\) error of
\(4.22\pm0.06\%\), compared with \(14.82\pm0.15\%\) for KCN,
\(23.35\%\) for the IDW-based pipeline, \(21.99\%\) for the \(k\)NN-based pipeline, and \(18.83\%\) for climatology. Using the strongest applicable classical predictor for each channel gives an average error of \(15.16\%\); MIMONet-NL reduces this error by \(72\%\).
For the directly observed modalities, MIMONet-NL attains errors of
\(3.10\pm0.06\%\) for temperature and \(0.46\pm0.01\%\) for salinity, compared with the strongest classical results of \(6.79\pm0.32\%\) and \(0.99\pm0.04\%\), respectively. KCN achieves \(5.77\pm0.23\%\) for temperature and \(0.86\pm0.01\%\) for salinity, showing that locality-based learning improves reconstruction when the corresponding target variable is represented among the inputs.
The more distinctive result is SSH reconstruction. From temperature and salinity observations alone, MIMONet-NL achieves an SSH error of
\(9.11\pm0.17\%\), compared with \(37.7\%\) for climatology,
\(37.84\pm0.28\%\) for KCN, and approximately \(59\%\) for ridge regression applied to the classically reconstructed \(T\) and \(S\) fields. This corresponds to a \(4.1\)-fold reduction relative to climatology. KCN’s climatology-level SSH performance, despite its stronger results for temperature and salinity, indicates that local spatial reconstruction alone is insufficient for recovering the withheld channel. MIMONet-NL instead performs cross-modal field recovery, using the joint, nonlocal statistical and physical relationships among temperature, salinity, and SSH represented in the reanalysis data to reconstruct a channel that carries no direct observations.
Figure~\ref{fig:real-world}j--r shows that the reconstructed fields preserve the Gulf Stream front and prominent mesoscale structures. Residuals are largest near frontal regions and sharp SSH gradients, where small spatial displacements produce comparatively large pointwise discrepancies. Within the stated reanalysis and sensor-withholding protocol, the results extend virtual sensing from sparse reconstruction and vertical extrapolation to recovery of a target field with no direct observations in the operator input.

\section{Discussion}
\label{sec:discussion}

This study establishes operator-based virtual sensing as a viable framework for reconstructing spatially continuous, coupled interior fields when direct instrumentation is sparse, physically restricted, or entirely unavailable for a target quantity. The simulation-based studies demonstrate feasibility under controlled conditions across increasingly complex, reactor-relevant flow configurations, while the measurement-based and observation-constrained datasets show that the framework remains effective beyond solver-generated data. These latter applications span three distinct sensing regimes: reconstruction within a sparsely instrumented domain, extrapolation beyond the available sensor range, and cross-modal recovery of an unobserved target field. MIMONet reduces error by \(72\)--\(84\%\) relative to the strongest applicable classical baselines across these datasets. The most distinctive result is cross-modal field recovery: the reconstruction of sea-surface height, a field observed by no sensor, from temperature and salinity observations alone, achieving \(9.11\pm0.17\%\) error compared with \(37.7\%\) for climatology. The overall evidence therefore supports operator learning not simply as a faster surrogate for numerical simulation, but as a general inference framework for recovering inaccessible physical states under progressively weaker observational support.

\paragraph{Why the operator formulation enables virtual sensing.}
The observed performance follows from treating virtual sensing as an operator-learning problem rather than as pointwise interpolation or conventional field-to-field surrogacy. MIMONet assigns separate branch encoders to heterogeneous measurements, allowing functional inputs, scalar operating conditions, and observations collected on distinct sensor networks to be represented within a shared latent space. Multiplicative fusion introduces cross-modal interactions before decoding, while the coordinate-conditioned trunk evaluates the reconstructed fields at arbitrary query locations. The resulting map can therefore transfer information across measurement modalities and spatially separated regions instead of relying only on observations near each query point. This distinction becomes increasingly important as the task progresses from interpolation within a sparsely observed domain to extrapolation and zero-sensor reconstruction. FNO remains effective when the measurements possess a regular grid structure, as in the vertical wind profiles, but requires an additional gridding stage for scattered sensor networks. KCN accepts scattered observations directly and performs well for target quantities represented among its inputs, but its climatology-level SSH result indicates that its locality-based formulation does not recover the required cross-modal relationship under the present zero-sensor protocol. MIMONet-NL addresses the greater complexity of this task through nonlinear decoding while retaining the same multimodal, coordinate-conditioned operator structure. The formulation provides an inductive bias for coupled field reconstruction, but it does not explicitly enforce conservation laws, incompressibility, or other governing constraints; the physical consistency observed here is learned from the training data and evaluated empirically rather than guaranteed by the architecture.

\paragraph{Real-time inference, uncertainty, and robustness.}
A second advantage of the operator formulation is that the computational cost of solving the inverse reconstruction problem is shifted to an offline training stage. Once trained, MIMONet evaluates the complete field through a single forward pass, reconstructing the heat-exchanger fields in \(0.35\,\mathrm{ms}\) at an energy cost of \(46\,\mathrm{mJ}\) on an H200 and retaining sub-millisecond inference on the evaluated A40, A100 and H200 platforms, with a worst case of \(1.17\,\mathrm{ms}\) on the GH200. These measurements refer specifically to model inference; the latency of sensor acquisition, communication, preprocessing, and downstream decision-making would also contribute to the response time of a deployed monitoring system. Uncertainty estimation is therefore essential because rapid predictions are useful only when their reliability can be assessed. MC-Dropout provides spatially resolved predictive uncertainty, while conformal calibration adjusts the resulting intervals to attain the prescribed empirical coverage on exchangeable held-out data. The uncertainty maps generally widen in regions that are weakly constrained by the available measurements or contain sharper field gradients, providing an indicator of where reconstructions should be treated cautiously. Under the prescribed sensor-noise perturbations, the stochastic model exhibits less than \(3\%\) channel-mean RMSE inflation, with a maximum of \(5.7\%\) on the pressure channel, and is substantially less sensitive than its deterministic counterpart and the evaluated GeoFNO and NOMAD baselines. These results jointly support fast, uncertainty-aware inference under the tested conditions, although calibration and robustness under sustained distribution shift, sensor drift, or structured hardware failures remain to be established.

\paragraph{Limitations and path to deployment.}
The present results define a promising but bounded validation envelope. The reactor-relevant studies use fields generated with a single CFD solver and RANS \(k\)--\(\varepsilon\) closure over fixed geometries and prescribed parameter ranges; performance under alternative solvers, turbulence closures, discretizations, or substantial geometric changes has not been established. Their targets are steady two-dimensional planes exported from three-dimensional simulations, rather than transient volumetric fields, and experimental interior thermal--hydraulic measurements were unavailable for independent validation. The three additional datasets provide important evidence beyond solver-generated training data, but they introduce their own qualifications. The PEMFC temperature field is mapped from a coarser measurement grid, the wind evaluation is limited to the retained towers and observed atmospheric conditions, and GLORYS12 is an observation-constrained reanalysis rather than independent ocean-sensor ground truth. Moreover, the experiments largely use fixed sensor configurations and do not fully represent sensor relocation, progressive calibration drift, correlated failures, or distribution shifts in operating conditions. Deployment therefore requires cross-solver verification, validation against controlled experiments, systematic testing under changing sensor networks, and evaluation of the complete monitoring loop rather than model inference alone. Extending the framework to spatiotemporal and geometry-aware operators would enable transient three-dimensional reconstruction across changing configurations, while physics-informed constraints and adaptive uncertainty calibration could improve consistency and reliability outside the training distribution. These developments would provide the necessary bridge from the demonstrated reconstruction capability to operational virtual sensing in energy and other safety-critical systems.

Taken together, the validated cases span a $50\,\mathrm{cm}^2$ electrochemical cell, atmospheric boundary layers at instrumented towers, and the North Atlantic basin, six orders of magnitude in spatial scale and four distinct governing-physics regimes. That a single operator formulation recovers inaccessible fields across this range, without architectural change and with the cross-modal recovery of an entirely unobserved channel, indicates that field-level observability is better viewed as a general inference capability than as a set of domain-specific techniques. Operator-based virtual sensing thus offers a route to real-time observability in precisely those systems where the states that matter most cannot be directly measured.

\section{Methods}
\label{sec:methods}

This section describes the operator formulation,
training and normalization procedures, uncertainty calibration and robustness
evaluation, and the metrics and computational profiling used to assess
reconstruction accuracy and real-time feasibility.
\subsection*{Operator Formulation and MIMONet Architecture}
\label{sec:operator_formulation}

Virtual sensing is formulated as learning an operator that maps available
measurements and operating conditions to spatially continuous fields in regions
where the target quantities cannot be measured directly. Let
\(\boldsymbol{u}=(u_1,\ldots,u_n)\in\mathcal{U}\) denote \(n\) heterogeneous
input modalities, which may include scalar operating parameters, sampled boundary
functions, actuation histories, or sparse field measurements. The objective is to
approximate
\begin{equation}
\mathcal{G}:\mathcal{U}\rightarrow
L^2(\Omega;\mathbb{R}^{m}),
\qquad
\mathcal{G}(\boldsymbol{u})(\mathbf{r})
=
\mathbf{s}(\mathbf{r})
=
\begin{bmatrix}
s_1(\mathbf{r}) & \cdots & s_m(\mathbf{r})
\end{bmatrix}^{\mathsf T},
\label{eq:virtual_sensing_operator}
\end{equation}
where \(\mathbf{r}\in\Omega\) is a query coordinate and
\(\mathbf{s}(\mathbf{r})\) contains the \(m\) coupled target quantities. Unlike
fixed-output regression, the coordinate-conditioned formulation can be evaluated
at arbitrary query locations within the represented physical domain, including
locations not used as input sensors.

The proposed MIMONet implements Eq.~\eqref{eq:virtual_sensing_operator} using modality-specific
branch encoders, multiplicative latent fusion, and shared-latent multi-field
decoding. Each input modality is encoded independently as
\begin{equation}
\mathbf{z}_i=\mathcal{B}_i(u_i;\boldsymbol{\theta}_i)
\in\mathbb{R}^{q},
\qquad i=1,\ldots,n,
\end{equation}
allowing inputs with different dimensions and physical meanings to be processed
without first mapping them to a common grid. The branch representations are fused
elementwise:
\begin{equation}
\boldsymbol{\psi}
=
\bigodot_{i=1}^{n}\mathbf{z}_i,
\qquad
\psi_j=\prod_{i=1}^{n}z_{ij},
\quad j=1,\ldots,q.
\label{eq:mimonet_fusion}
\end{equation}
This multiplicative operation introduces interactions among the input modalities
before spatial decoding and provides a separable inductive bias for learning
coupled responses. It does not, by itself, impose the bilinear terms or
conservation laws of the governing equations.
A trunk network encodes the query coordinate, optionally after a Fourier-feature
mapping, into channel-specific basis vectors
\(\mathbf{t}_o(\mathbf{r})\in\mathbb{R}^{q}\). In the standard MIMONet decoder,
the prediction for output channel \(o\) is
\begin{equation}
\widehat{s}_o(\mathbf{r})
=
\left\langle
\boldsymbol{\psi},
\mathbf{t}_o(\mathbf{r})
\right\rangle+b_o,
\qquad o=1,\ldots,m.
\label{eq:mimonet_decoder}
\end{equation}
All target fields are therefore decoded from the same fused latent state
\(\boldsymbol{\psi}\), while the channel-specific basis vectors allow each output
to retain its own spatial structure. Sharing the fused representation encourages
coordinated multi-field prediction but does not guarantee incompressibility,
conservation, or other forms of physical consistency; these properties are
assessed empirically.

For the ocean reconstruction task, the greater complexity of predicting a channel
with no direct observations motivated a nonlinear-decoder variant, denoted
MIMONet-NL. It retains the same modality-specific branches, multiplicative fusion,
and coordinate trunk, but replaces the linear contraction in
Eq.~\eqref{eq:mimonet_decoder} with
\begin{equation}
\widehat{\mathbf{s}}(\mathbf{r})
=
\mathcal{D}_{\mathrm{NL}}
\left(
\left[
\boldsymbol{\psi};
\mathcal{T}(\mathbf{r})
\right]
\right),
\label{eq:mimonet_nl}
\end{equation}
where \([\cdot\,;\cdot]\) denotes concatenation and
\(\mathcal{D}_{\mathrm{NL}}\) is a multilayer perceptron. This decoder permits
nonlinear interaction between the fused measurement representation and the query
coordinate while preserving the common operator formulation used across the six
evaluation problems.
The formulation assumes that the supplied measurements and operating conditions
contain sufficient information to identify the target fields over the evaluated
data distribution. Observability is therefore assessed empirically through
held-out reconstruction accuracy, uncertainty, and robustness experiments rather
than asserted as a formal guarantee.

\subsection*{Training, Normalization, and Baseline Implementation}
\label{sec:training}

In all cases, models were trained to minimize the mean-squared error between predicted and
reference fields over the training samples, query locations, and output channels:
\begin{equation}
\mathcal{L}(\boldsymbol{\theta})
=
\frac{1}{N_{\mathrm{tr}}Pm}
\sum_{b=1}^{N_{\mathrm{tr}}}
\sum_{p=1}^{P}
\sum_{o=1}^{m}
\left[
\widehat{\widetilde{s}}_{bpo}
-
\widetilde{s}_{bpo}
\right]^2,
\label{eq:training_loss}
\end{equation}
where \(N_{\mathrm{tr}}\), \(P\), and \(m\) denote the numbers of training
samples, query points, and output channels, respectively. Model selection was
performed using the validation loss, and the test partition was used only for
final evaluation. MIMONet-NL was used only for the ocean reconstruction task; all
other cases used the standard decoder in Eq.~\eqref{eq:mimonet_decoder}.
Because the target fields have different physical units and numerical ranges,
each output channel was scaled independently to \([-1,1]\). For channel \(o\),
the transformation was
\begin{equation}
\widetilde{s}_{bpo}
=
-1+
2\,
\frac{s_{bpo}-s_o^{\min}}
{s_o^{\max}-s_o^{\min}+\varepsilon},
\qquad
s_o^{\min}=\min_{b,p}s_{bpo},
\quad
s_o^{\max}=\max_{b,p}s_{bpo},
\label{eq:target_scaling}
\end{equation}
where the extrema were calculated from the training partition only and
\(\varepsilon\) prevents division by zero. The same transformation was applied
unchanged to validation and test targets. Predictions were returned to physical
units before calculating errors, physical residuals, or uncertainty intervals.
Channel-wise scaling prevents fields with larger numerical magnitudes, such as
pressure or temperature, from dominating the joint optimization objective.
Learned baselines were trained using the same input information, target fields,
and data partitions as MIMONet. Model capacities were matched where permitted by
the respective architectures, and hyperparameters were selected using validation
performance without access to the test partition. FNO was evaluated directly when
the input possessed a regular ordered representation and through an explicitly
reported gridding front end where scattered observations required preprocessing.
KCN was applied directly to scattered sensor inputs. Classical methods used the
same retained sensor locations as the learned models and were fitted using only
information available within the corresponding training or input protocol.
Results from multiple initializations are reported as the mean and standard
deviation where applicable.

The models were implemented in PyTorch and trained on NVIDIA A40, A100, H200, or
GH200 GPUs available through the DeltaAI computing
system~\cite{deltaai_doc}. Architecture dimensions, parameter counts, optimizer
settings, learning-rate schedules, batch sizes, stopping criteria, random seeds,
and baseline-specific configurations are reported in Supplementary
Material.

\subsection*{Uncertainty Calibration and Robustness Evaluation}
\label{sec:uncertainty}

Predictive uncertainty was estimated using Monte Carlo dropout
(MC-Dropout). At inference, the trained weights were held fixed while dropout
layers in the branch and trunk networks remained active with probability
\(p=0.10\). For an input \(\boldsymbol{u}\), \(E=20\) stochastic operator
evaluations were performed. The predictive mean and variance at query coordinate
\(\mathbf{r}\) were estimated as
\begin{equation}
\widehat{\boldsymbol{\mu}}(\mathbf{r})
=
\frac{1}{E}\sum_{e=1}^{E}
\widehat{\mathbf{s}}^{(e)}(\mathbf{r}),
\qquad
\widehat{\boldsymbol{\sigma}}^{\,2}(\mathbf{r})
=
\frac{1}{E-1}\sum_{e=1}^{E}
\left[
\widehat{\mathbf{s}}^{(e)}(\mathbf{r})
-
\widehat{\boldsymbol{\mu}}(\mathbf{r})
\right]^2 .
\label{eq:mc_dropout}
\end{equation}
The resulting spread provides a computationally tractable approximation of
epistemic uncertainty induced by the stochastic subnetworks; it is not equivalent
to exact Bayesian posterior inference and does not, by itself, represent all
sources of measurement or model-form uncertainty.

Because raw MC-Dropout intervals need not attain their nominal coverage, they were
calibrated using split conformal prediction. For output channel \(o\), normalized
nonconformity scores were calculated on a held-out calibration partition as
\begin{equation}
a_{bpo}
=
\frac{
\left|
s_{bpo}-\widehat{\mu}_{bpo}
\right|
}{
\widehat{\sigma}_{bpo}+\varepsilon
},
\label{eq:conformal_score}
\end{equation}
where \(\varepsilon\) prevents instability at locations with negligible stochastic
variance. Let \(q_o^{(1-\alpha)}\) denote the finite-sample conformal quantile of
the calibration scores for channel \(o\). The calibrated interval at miscoverage
level \(\alpha\) is
\begin{equation}
\mathcal{C}_{o}(\mathbf{r})
=
\left[
\widehat{\mu}_{o}(\mathbf{r})
-
q_o^{(1-\alpha)}
\bigl(\widehat{\sigma}_{o}(\mathbf{r})+\varepsilon\bigr),
\;
\widehat{\mu}_{o}(\mathbf{r})
+
q_o^{(1-\alpha)}
\bigl(\widehat{\sigma}_{o}(\mathbf{r})+\varepsilon\bigr)
\right].
\label{eq:conformal_interval}
\end{equation}
We used \(\alpha=0.05\) to construct nominal \(95\%\) intervals. Coverage and
interval width were evaluated only on the test partition. Under the standard
exchangeability assumption between calibration and test samples, this procedure
provides marginal finite-sample coverage rather than simultaneous coverage at
every spatial location.

Robustness to sensor corruption was evaluated by perturbing the input
measurements with zero-mean Gaussian noise. For modality \(i\), the corrupted
input was generated as
\begin{equation}
u_i^{(\rho)}
=
u_i+\rho\,\sigma_{u_i}\boldsymbol{\epsilon},
\qquad
\boldsymbol{\epsilon}\sim\mathcal{N}(\mathbf{0},\mathbf{I}),
\label{eq:sensor_noise}
\end{equation}
where \(\sigma_{u_i}\) is the training-set standard deviation of the corresponding
input and \(\rho\) controls the prescribed noise magnitude. The reference target
fields were left unchanged, so the experiment measures sensitivity to input
corruption rather than changes in the underlying physical state. Identical noisy
inputs were supplied to deterministic MIMONet, stochastic MIMONet, GeoFNO, and
NOMAD, and robustness was quantified through the relative increase in RMSE from
the clean-input condition. The evaluated noise levels, calibration-set
construction, conformal quantile convention, and channel-wise coverage results
are reported in Supplementary materials.

\subsection*{Evaluation Metrics and Computational Profiling}
\label{sec:evaluation}

Reconstruction accuracy was evaluated using the channel-wise relative
\(\ell_{2}\) error. For test sample \(b\) and output channel \(o\), evaluated at
query points \(\{\mathbf{r}_p\}_{p=1}^{P}\), the error is
\begin{equation}
e_{\mathrm{rel}}^{(b,o)}
=
\frac{
\left[
\sum_{p=1}^{P}
\left(
\widehat{s}_{o}^{(b)}(\mathbf{r}_p)
-
s_{o}^{(b)}(\mathbf{r}_p)
\right)^2
\right]^{1/2}
}{
\left[
\sum_{p=1}^{P}
s_{o}^{(b)}(\mathbf{r}_p)^2
\right]^{1/2}
}.
\label{eq:relative_l2}
\end{equation}
Per-channel results are the arithmetic mean of
\(e_{\mathrm{rel}}^{(b,o)}\) over all test samples and are reported as
percentages. When a single multi-field error is reported, the per-channel errors
are averaged with equal weight so that fields with larger physical magnitudes do
not dominate the aggregate. Results over repeated model initializations are
reported as the mean and standard deviation. Robustness is evaluated through the
percentage change in root-mean-square error relative to the corresponding
clean-input condition, while uncertainty performance is assessed using empirical
coverage and mean interval width.

Forward-pass latency and energy use were profiled on NVIDIA A40, A100, H200, and
GH200 GPUs using the trained heat-exchanger model. Each evaluation used a batch
size of one and \(P=3{,}977\) query points, corresponding to one complete
multi-field reconstruction. Gradient tracking was disabled, and ten preliminary
iterations were used to stabilize memory allocation, kernel caching, and GPU
clock states before \(N=1{,}000\) timed evaluations. CUDA synchronization was
performed immediately before and after every forward pass to prevent asynchronous
kernel execution from biasing the measurement. Latency was measured using
\texttt{time.perf\_counter()}, and the reported value is the mean with one
standard deviation over the \(1{,}000\) evaluations. Throughput was calculated as
the reciprocal of mean latency.

GPU board power was sampled through NVIDIA's NVML interface immediately before
and after each synchronized forward pass and averaged to obtain \(P_i\). The
estimated board energy associated with inference \(i\) was then calculated as
\(E_i=P_i t_i\), giving
\begin{equation}
\overline{E}
=
\frac{1}{N}
\sum_{i=1}^{N}P_i t_i .
\label{eq:inference_energy}
\end{equation}
The reported power represents total board draw without idle-baseline subtraction;
the resulting quantity should therefore be interpreted as an estimate of
per-inference board energy rather than a direct measurement of incremental model
energy. Throughout this work, real-time inference refers specifically to
the operator forward-pass latency. Sensor acquisition, communication,
preprocessing, and downstream decision-making were not included in the timing
measurements and would contribute to the latency of a complete deployed
monitoring loop.

\section*{Data Availability} The simulation data generated in this study --- the transient lid-driven cavity, the pressurized-water-reactor subchannel and the compact heat-exchanger cases --- have been deposited in the Zenodo database under accession code \href{https://doi.org/10.5281/zenodo.21224751}{10.5281/zenodo.21224751}. The processed arrays for the three measured-data problems, the sensor draws and data splits for all five random seeds, the trained model checkpoints and the per-seed result files underlying every figure and table in this Article and its Supplementary Information are available at the same deposit. The proton-exchange-membrane fuel-cell data used in this study are available in the idUS repository of the Universidad de Sevilla under accession code \href{https://idus.us.es/handle/11441/160750}{11441/160750}, and are described in ref.~\cite{toharias2024dib}; the accompanying experimental study is ref.~\cite{suarez2023cdm}. The tall-tower wind data used in this study are available in the EUDAT B2SHARE database under accession code \href{https://doi.org/10.23728/b2share.136ecdeee31a45a7906a773095656ddb}{10.23728/b2share.136ecdeee31a45a7906a773095656ddb}, and are described in ref.~\cite{ramon2020iea}. The ocean reanalysis data used in this study are available in the Copernicus Marine Service database under accession code \href{https://doi.org/10.48670/moi-00021}{GLOBAL\_MULTIYEAR\_PHY\_001\_030}; we use surface potential temperature, salinity and sea-surface height over $35$--$50^{\circ}$N, $60$--$20^{\circ}$W for the $730$ daily fields spanning 1 January 2017 to 31 December 2018. Access to the source product requires free registration with the Copernicus Marine Service.
\section*{Data Availability}
The three real-world datasets analysed in this study are publicly available. The proton-exchange-membrane fuel-cell current-density-mapping dataset is available from the University of Seville idUS repository under a CC~BY-NC licence~\cite{suarez2023cdm}. The BSC/IEA-Wind Tall Tower dataset is available on EUDAT B2SHARE under a CC~BY-NC licence~\cite{ramon2020iea}. The GLORYS12V1 global ocean reanalysis is distributed by the Copernicus Marine Service as product GLOBAL\_MULTIYEAR\_PHY\_001\_030~\cite{lellouche2021glorys}. The three preprocessed simulation datasets (lid-driven cavity, pressurized-water-reactor subchannel, and heat exchanger) generated in this study are available with the code at \url{https://github.com/kkazuma19/MIMONet} and in an archived Zenodo release (https://doi.org/10.5281/zenodo.21224751). 

\section*{Code Availability}
The full implementation of MIMONet, including source code, training and evaluation scripts, dataset configurations, the three preprocessed simulation datasets, and the trained model checkpoints for every reported result, is publicly available at \url{https://github.com/kkazuma19/MIMONet} and archived on Zenodo under accession code \href{https://doi.org/10.5281/zenodo.21224751}{10.5281/zenodo.21224751}~\cite{kobayashi2026mimonet_code}. 

\bibliographystyle{unsrt}
\bibliography{references}

\section*{Acknowledgements}
This research used both the DeltaAI advanced computing and data resource, supported by the National Science Foundation (award OAC 2320345) and the State of Illinois, and the Delta advanced computing and data resource, supported by the National Science Foundation (award OAC 2005572) and the State of Illinois. Delta and DeltaAI are joint efforts of the University of Illinois Urbana-Champaign and its National Center for Supercomputing Applications. During the preparation of this work, the authors used a large language model solely for language editing and refinement. The authors reviewed and edited the content as needed and take full responsibility for the content of the publication.

\section*{Funding statement}
S. B. Alam acknowledges support from the U.S. Nuclear Regulatory Commission under grant numbers 31310025M0012 and 31310024M0041. S. Chakraborty acknowledges Fulbright-Nehru Academic and Professional Excellence Fellowship for supporting his visit to the University of Illinois Urbana-Champaign.

\section*{Author Contributions Statement} 
K.K. and S.B.A. contributed to the conceptualization of the study. Methodology was developed by K.K., F.A., S.S. and S.C. Software was implemented by K.K., F.A., J.P. and S.K. Formal analysis was carried out by K.K., J.P., S.S. and S.C. Resources were provided by S.K. and S.B.A. The original draft was written by K.K., S.C. and S.B.A. Writing -- review and editing were performed by K.K., S.K., S.C. and S.B.A. Visualization was conducted by K.K. Supervision and funding acquisition were undertaken by S.B.A.

\section*{Competing Interests Statement}
The authors declare no competing interests.

\section*{Supplementary Information}
Supplementary Information accompanies this paper and contains the operator formulation and its theoretical basis; complete dataset preparation, architectures, training configurations and baseline definitions for all six problems; complete per-channel and five-seed results, including a forward-in-time robustness check for the ocean problem; conformal calibration, Lipschitz stability, sample-complexity and sensor-noise studies; and complexity and cross-GPU inference benchmarking. Supplementary Figs.~S1--S9 and Supplementary Tables~S1--S23.

\clearpage
\appendix

\setcounter{section}{0}
\setcounter{figure}{0}
\setcounter{table}{0}
\setcounter{theorem}{0}
\setcounter{assumption}{0}
\setcounter{equation}{0}
\renewcommand{\thesection}{S\arabic{section}}
\renewcommand{\thesubsection}{S\arabic{section}.\arabic{subsection}}
\renewcommand{\thefigure}{S\arabic{figure}}
\renewcommand{\thetable}{S\arabic{table}}
\renewcommand{\theequation}{S\arabic{equation}}

\begin{center}
{\LARGE\bfseries Supplementary Information}\\[0.6em]
{\large Virtual Sensing to Enable Real-Time Monitoring of Inaccessible Locations
\& Unmeasurable Parameters}
\end{center}
\vspace{1em}

\section*{Notation and Glossary}
\label{sec:supp_notation}

Symbols used throughout the main text and this Supplementary Material are summarised in Table~\ref{tab:supp_notation} for convenience. Quantities defined only within a single section retain their context-specific meaning given there.

\begin{table}[h!]
\centering
\caption{\textbf{Notation and glossary.} Domain symbols are reused across the main text and Supplementary Material.}
\label{tab:supp_notation}
\small
\begin{tabular}{@{}p{0.18\textwidth}p{0.76\textwidth}@{}}
\toprule
\textbf{Symbol} & \textbf{Definition} \\
\midrule
\multicolumn{2}{l}{\emph{Domains and spaces}} \\
$\Omega \subset \mathbb R^d$ & Spatial domain of the governing PDE; $d=2$ throughout the case studies. \\
$\mathcal Y \subset \mathbb R^d$ & Set of admissible query coordinates (interior + boundary). \\
$\mathcal U = \prod_{i=1}^n \mathcal F_i$ & Heterogeneous input space; each $\mathcal F_i$ is a finite- or infinite-dimensional space of one input modality (scalars, function-valued profiles, time series). \\
$\mathcal S \subset L^2(\mathcal Y;\mathbb R^m)$ & Output (solution) space of $m$-channel fields over $\mathcal Y$. \\
$\mathcal X,\,\mathcal Y_{\!q}$ & Subsets of the domain where data are observed (sensors) and where reconstruction is requested (virtual queries) respectively. The disjoint case $\mathcal X\cap\mathcal Y_{\!q}=\varnothing$ is the cross-domain inference regime that Problems~5 and~6 instantiate; the symbols are used descriptively here and not in any derivation. \\
\midrule
\multicolumn{2}{l}{\emph{Operator and architecture}} \\
$\mathcal G:\mathcal U\to\mathcal S$ & True (continuous) nonlinear forward operator mapping boundary inputs to interior fields. \\
$\hat{\mathcal G}_\theta$ & MIMONet surrogate operator parameterised by trainable parameters $\theta$. \\
$\boldsymbol u = (\boldsymbol u_1,\dots,\boldsymbol u_n)$ & Multi-modal input drawn from $\mathcal U$. \\
$\mathbf s(\mathbf r) = \mathcal G(\boldsymbol u)(\mathbf r)$ & Multi-channel output field at coordinate $\mathbf r\in\mathcal Y$. \\
$\phi_i:\mathcal F_i\to\mathbb R^{I}$ & Branch encoder for modality $i$; produces a width-$I$ latent vector. \\
$\tau:\mathbb R^d\to\mathbb R^{I\times m}$ & Trunk encoder mapping a query coordinate to per-channel decoding weights. \\
$\psi_j(\boldsymbol u) = \prod_{i=1}^n [\phi_i(\boldsymbol u_i)]_j$ & Fused branch latent (multiplicative aggregation; see Supplementary Note~1 in Section~\ref{sec:supp_math}). \\
$\boldsymbol\beta \in \mathbb R^m$ & Per-channel bias term applied after trunk--branch contraction. \\
$\hat{s}_o(\mathbf r\,|\,\boldsymbol u)=\sum_{j=1}^{I}\psi_j(\boldsymbol u)\,\tau_{j,o}(\mathbf r)+\beta_o,\ o=1,\dots,m$ & MIMONet's per-channel pointwise prediction at coordinate $\mathbf r$ (equivalent vector form $\hat{\mathbf s}=\sum_j \psi_j\,\boldsymbol\tau_j+\boldsymbol\beta$ with $\boldsymbol\tau_j=\tau_{\cdot,j,\cdot}\in\mathbb{R}^m$). \\
$I$ & Branch--trunk latent width ($256$ for Problems~1--4, $128$ for Problem~5, $384$ for Problem~6; see Section~\ref{sec:supp_architectures}). \\
$B,\;P$ & Batch size and number of trunk query points per sample. \\
\midrule
\multicolumn{2}{l}{\emph{Training and inference}} \\
$N_{\text{train}}$ & Training-pool size used in the sample-complexity sweep. \\
$n$ & Number of input modalities (e.g.\ $n=1$ for LDC, $n=2$ for Subchannel and HX). Distinct from $n_{\text{NOMAD}}$ (NOMAD latent dimension) and from $n_x$ (state-space dimension of classical observers). \\
$\sigma_{\text{noise}}$ & Gaussian noise standard deviation in the standardised input space. \\
$\Delta\mathrm{RMSE}(\sigma_{\text{noise}})\,(\%)$ & Relative RMSE inflation at noise level $\sigma_{\text{noise}}$ vs.\ the $\sigma=0$ baseline: $100\cdot[\mathrm{RMSE}(\sigma)-\mathrm{RMSE}(0)]/\mathrm{RMSE}(0)$. \\
$L_{99,o},\;L\equiv\max_o L_{99,o}$ & Per-channel and global empirical Lipschitz proxies (99th percentile of $\|\Delta\mathcal G_\theta\|_2/\|\boldsymbol\delta\|_2$). \\
$\gamma$ & Power-law exponent in the sample-complexity fit $\epsilon(N)\approx a\,N^{-\gamma}$. Distinct from the per-channel bias $\boldsymbol\beta\in\mathbb{R}^m$ above. \\
$\alpha,\;s_c$ & $\alpha$ = conformal miscoverage level (e.g.\ $1-\alpha=0.95$ for $95\%$ intervals). $s_c=\mathrm{std}(z_c)$ = per-channel post-hoc variance-calibration factor, where $z_c$ is the vector of raw calibration-set z-scores $(\mathbf s-\hat{\mathbf s})/\sigma_G$ for channel $c$; renamed from $\alpha_c$ to avoid collision with the conformal level. Two unrelated uses of $\alpha$ occur locally in Section~\ref{sec:supp_baseline_defs}: the wind-shear power-law exponent and the ridge regularisation strength. \\
$N$ (in MC-Dropout) & Number of stochastic forward passes per inference; $N=20$ throughout. \\
\midrule
\multicolumn{2}{l}{\emph{Field-channel labels (per case)}} \\
LDC: $p$, $\|\boldsymbol v\|$, $k$ & Pressure, velocity-magnitude, turbulent kinetic energy. \\
PWR subchannel: $\|\boldsymbol v\|$, $T$, $k$ & Velocity-magnitude, temperature, turbulent kinetic energy. \\
Heat exchanger: $p$, $u_z$, $u_y$, $u_x$ & Pressure and three velocity components. \\
\midrule
\multicolumn{2}{l}{\emph{Constants used in the case studies}} \\
$P_{\text{Sub}}=1{,}733$ & Trunk-mesh size for the PWR subchannel test set. \\
$P_{\text{LDC}}=4{,}225$ & Trunk-mesh size for the lid-driven cavity test set. \\
$P_{\text{HX}}=3{,}977$ & Trunk-mesh size for the heat-exchanger test set. \\
\bottomrule
\end{tabular}
\end{table}

\appendix
\setcounter{section}{0}

\makeatletter
\renewcommand*\l@section[2]{\@dottedtocline{1}{0em}{2.6em}{\bfseries #1}{\bfseries #2}}
\renewcommand*\l@subsection{\@dottedtocline{2}{2.6em}{3.4em}}
\makeatother
\tableofcontents
\clearpage

\section{Overview of the six problems}
\label{sec:supp_overview}

\paragraph{What this Supplementary is arguing, and what it is not.} The claim under test throughout is the one set out in the Introduction, and it is a claim about deployment regime rather than about what is recoverable in principle. Classical state estimators can reconstruct spatially distributed fields from boundary data; what they do is propagate a state recursively under an explicit governing model and observation operator, on a prescribed grid or spatial basis whose coefficients are estimated, so that the governing model is part of online inference and evaluation at a new coordinate is tied to that representation. Established data-driven virtual sensors avoid the online model but return values at a set of locations fixed at training time. The regime addressed here is different from both: a map from sparse, heterogeneous, scalar- and function-valued measurements on a sensing region to a spatially continuous, coupled field over a reconstruction region that may be disjoint from it, evaluated at arbitrary query coordinates in a single amortised forward pass, with no PDE solved online and no retraining per operating condition. Where this Supplementary says that a classical estimator has no mechanism for a particular task, the statement is always about the estimator classes actually compared, which are spatial interpolants and fixed analytic profiles, and never a claim that no classical method could recover the field by other means. The comparisons that bear on that claim are against the classical estimators, and they are reported for Problems~4--6 in Section~\ref{sec:supp_realworld_results}. They are not reported for Problems~1--3, and the reason is structural rather than an omission: those problems map boundary data to interior fields, so there are no samples of the output field to interpolate and no classical interpolant is defined. The simulation problems establish accuracy against other operator formulations on coupled multiphysics fields; the measured-data problems carry the comparison against what is deployed in practice. Comparisons among neural-operator architectures appear in several places, and their role is different: they document design decisions, establish that a reported effect is not an artefact of one particular architecture, and bound how much of a result depends on the choice of operator. On the measured-data problems those differences are small next to the gap between the operator formulation and the classical alternatives. On the simulation problems they are not: the simulation-case results in Section~\ref{sec:supp_sim_results} span a factor of twenty between operator architectures, which is larger than any operator-versus-classical gap reported here. We do not read that spread as evidence of architectural superiority, because the baselines were not tuned per problem and carry very different capacities; it is reported because the main text quotes it, and it is not load-bearing. No conclusion in this work depends on a particular operator architecture being the best of its family.

Table~\ref{tab:supp_overview} summarises what each problem asks of the operator, where the data comes from, and how it is divided. The three simulation problems (1--3) are generated with ANSYS Fluent and share a common structure: a boundary function together with scalar operating conditions is mapped to interior fields on a fixed CFD mesh. The three measured-data problems (4--6) differ from them in kind, not only in origin: their inputs are point sensors rather than boundary data, so the operator is asked to complete a partially observed state rather than to map boundary conditions to an interior. The three differ in how far that completion has to reach. In Problem~4 the sensors and the targets lie on the same grid, so the task is spatial infilling. In Problem~5 every target height lies above every input sensor, so the operator extrapolates beyond the instrumented range. In Problem~6 one of the three output channels is observed by no sensor of any kind, so the operator must infer it across modalities; that case is the sharpest form of the disjoint sensing and reconstruction domains this work addresses.

\begin{table}[h!]
\centering
\caption{\textbf{The six virtual-sensing problems.} $P$ is the number of query coordinates at which the output field is evaluated. ``Sensor regime'' states what the operator receives at inference. Split sizes are given as training / validation / test in the units of the split column; the split protocol is stated because it differs across problems and is essential to reading the reported errors.}
\label{tab:supp_overview}
\adjustbox{max width=\textwidth}{%
\begin{tabular}{@{}llllrl@{}}
\toprule
\# & Problem & Inputs & Outputs & $P$ & Split (unit; protocol) \\
\midrule
1 & Lid-driven cavity & Lid velocity $V(t)$, $90$ time levels & $p$, $\|\boldsymbol v\|$, $k$ & $4{,}225$ & $3{,}159/790/988$ (samples; by trajectory) \\
2 & PWR subchannel & $q(z)$ at $100$ axial points; $T_{\text{in}}, v_{\text{in}}$ & $T$, $\|\boldsymbol v\|$, $k$ & $1{,}733$ & $3{,}200/800/1{,}000$ (samples; random) \\
3 & Heat exchanger & $q_{\text{wall}}$ at $100$ points; $T_{\text{in}}, u_{\text{in}}$ & $p$, $u_z$, $u_y$, $u_x$ & $3{,}977$ & $1{,}082/154/310$ (samples; random) \\
\midrule
4 & PEM fuel cell & $J$ and $T$ at $n_{\text{obs}}$ of $324$ segments & $J$, $T$ at all $324$ & $324$ & $5{,}600/1{,}200/1{,}200$ (frames; random) \\
5 & IEA-Wind towers & $U$ at $10/25/50\,$m; site and hour & $U$ at $60$--$100\,$m & $7$ & $421{,}013/90{,}219/90{,}222$ (profiles; chronological per tower) \\
6 & GLORYS12 ocean & $T$ at $400$ and $S$ at $400$ disjoint cells & $T$, $S$, SSH at all cells & $9{,}457$ & $610/60/60$ (days; random) \\
\bottomrule
\end{tabular}}
\end{table}

\begin{table}[h!]
\centering
\caption{\textbf{Data provenance and sensor regime.} ``Observed fraction'' is the share of the reconstruction domain that the operator sees at inference; it is not defined for Problems~1--3, whose inputs are boundary data rather than samples of the output field. Problem~5 reconstructs a one-dimensional profile, so its observed fraction refers to heights rather than to area.}
\label{tab:supp_provenance}
\adjustbox{max width=\textwidth}{%
\begin{tabular}{@{}llll@{}}
\toprule
\# & Source & Sensor regime at inference & Observed fraction \\
\midrule
1 & ANSYS Fluent, transient RANS & Boundary actuation history only & --- \\
2 & ANSYS Fluent, steady RANS & Axial heat source and two inlet scalars & --- \\
3 & ANSYS Fluent, steady RANS & Wall heat flux and two inlet scalars & --- \\
\midrule
4 & Segmented-cell laboratory measurements & $n_{\text{obs}}$ point sensors on the mapping grid & $1.9$--$50\%$ of $324$ \\
5 & BSC/IEA-Wind Tall Tower Dataset~\cite{ramon2020iea} & Three lower-level anemometers & $3$ of $7$ heights \\
6 & GLORYS12V1 reanalysis & $800$ point sensors, two disjoint networks & $8.5\%$ of $9{,}457$ \\
\bottomrule
\end{tabular}}
\end{table}

Two entries in Tables~\ref{tab:supp_overview} and~\ref{tab:supp_provenance} deserve emphasis because they qualify how the corresponding results should be read, and both are documented in full in Section~\ref{sec:supp_datasets}. The split protocol is not uniform: Problem~5 is split chronologically within each tower, Problem~1 is split at the level of complete trajectories so that a lid-velocity history cannot straddle the split, and Problems~2, 3, 4 and~6 are split by a random permutation of samples, frames or days. For Problems~4 and~6 in particular, where consecutive samples are temporally adjacent recordings of an autocorrelated system, a random split places test states close to training states; Section~\ref{sec:supp_ocean_chrono} quantifies what that means for the ocean problem. Separately, the observed fraction for Problem~4 refers to the current-density channel; the temperature channel is measured on a $9\times9$ grid and bilinearly upsampled to the $18\times18$ reconstruction grid, so it is not observed at $324$ points in any configuration.

\section{Mathematical formulation: the neural operator as a virtual sensor}
\label{sec:supp_math}
This section states formally the map that the Introduction describes in words, and fixes the notation used throughout. The formulation is not incidental to the claim: it is what makes the output a function rather than a vector of values at pre-chosen locations, and that is the property the deployment regime turns on.

A class of learning problems is considered in which the objective is to approximate nonlinear operators that map combinations of scalar and function-valued inputs to spatially distributed physical outputs. Such operators are encountered naturally in systems governed by partial differential equations (PDEs), particularly in thermal--fluid domains where boundary conditions, internal heat generation, and flow control determine the system response. Formally, the learned map is treated as an operator on function spaces rather than as a finite-dimensional regressor, ensuring that inference is defined for any admissible input function within a prescribed set.

This formulation encompasses mixed-modal input spaces, including both low-dimensional parametric variables (such as operating setpoints or control inputs) and high-dimensional function-valued fields (such as inlet velocity profiles or spatially varying heat sources). The objective is to learn an operator that maps these inputs to physically meaningful output fields (typically vector- or tensor-valued)
distributed over a spatial domain of interest. These output fields may represent temperature, velocity, or turbulence kinetic energy, which are often difficult to measure directly in reactor-relevant environments. Crucially, by learning the \emph{field-valued} response, the framework supports virtual sensing at \emph{inaccessible locations} (points with no instrumentation) and recovery of \emph{unmeasurable parameters}.

Across all scenarios, the inputs represent operational conditions derived from design specifications, actuator configurations, or sensor-informed system states. The learned operator thus enables high-resolution virtual diagnostics, reconstructing internal fields from sparse, heterogeneous, or boundary-only measurements. The emphasis is on \emph{amortised} evaluation: once trained, the operator can be queried on arbitrary coordinate sets without retraining, decoupling inference cost from mesh resolution.

Let $D\subset\mathbb{R}^d$ denote the spatial domain and $\mathcal{Y}\subset D$ a lower-dimensional subset (e.g., a measurement or reconstruction plane). The forward operator is
\[
\mathcal{G}:\mathcal{U}\to\mathcal{S},\qquad
\mathcal{G}(\boldsymbol{u})(\mathbf{r})=\mathbf{s}(\mathbf{r}),\ \ \mathbf{r}\in\mathcal{Y},
\]
where $\boldsymbol{u}=(u_1,\dots,u_n)\in\mathcal{U}$ with $\mathcal{U}=\prod_{i=1}^n \mathcal{F}_i$, and each $u_i\in\mathcal{F}_i$ is either $\mathbb{R}$ or a function space such as $L^2(\Omega_i)$ with $\Omega_i\subseteq\overline{D}$. The output $\mathbf{s}(\mathbf{r})\in\mathbb{R}^m$ encodes $m$ physical quantities on $\mathcal{Y}$, and the output space is $\mathcal{S}\subset L^2(\mathcal{Y};\mathbb{R}^m)$. On compact subsets $\mathcal{K}\subset\mathcal{U}$, we seek $\hat{\mathcal{G}}$ that is uniformly close to $\mathcal{G}$ in the Bochner norm $\|\cdot\|_{L^2(\mathcal{Y};\mathbb{R}^m)}$, i.e., $\sup_{\boldsymbol{u}\in\mathcal{K}}\|\hat{\mathcal{G}}(\boldsymbol{u})-\mathcal{G}(\boldsymbol{u})\|\le \varepsilon$.

In practice, inputs and outputs are observed on discrete grids. Let $\Pi_{\!U}$ and $\Pi_{\!Y}$ denote sampling operators for inputs and outputs, respectively. Given a dataset $\mathcal{D}=\{(\Pi_{\!U}\boldsymbol{u}^{(i)}, \Pi_{\!Y}\mathbf{s}^{(i)})\}_{i=1}^N$ drawn from an operating distribution on $\mathcal{K}\times\mathcal{S}$, we minimise empirical loss
\[
\min_{\theta}\ \frac{1}{N}\sum_{i=1}^N \ell\!\left( \Pi_{\!Y}\hat{\mathcal{G}}_\theta(\boldsymbol{u}^{(i)}),\ \Pi_{\!Y}\mathbf{s}^{(i)} \right),
\]
with $\ell$ a channel-wise $L^2$ loss. Because $\hat{\mathcal{G}}_\theta(\boldsymbol{u})$ is defined for any query set, evaluation on \emph{inaccessible} coordinates is obtained by replacing $\Pi_{\!Y}$ with arbitrary point sets.

To approximate the nonlinear operator $\mathcal{G}:\mathcal{U}\rightarrow\mathcal{S}$, a mixed-input, mixed-output branch--trunk architecture (MIMONet) is employed. This design accommodates both scalar and function-valued inputs and produces multichannel fields at arbitrary query locations, thereby aligning the learning task with the operator viewpoint. The representation is resolution-invariant: the same parameters decode fields on different meshes simply by changing the set of query coordinates.

Each input component $u_i\in\mathcal{F}_i$ is first encoded into a fixed-width latent vector by a feedforward encoder $\phi_i:\mathcal{F}_i\to\mathbb{R}^{I}$. The resulting latents are then fused element-wise to form a single representation
\[
\psi_j=\prod_{i=1}^{n}[\phi_i(u_i)]_j,\ \ j=1,\dots,I.
\]
This multiplicative fusion preserves permutation invariance across modalities and yields a compact, modality-agnostic summary of the inputs. Physically, the product structure serves as a low-rank surrogate for multiplicative couplings in the governing PDEs (e.g., convective transport), encouraging a shared latent $\psi$ that is decoded into all output channels through channel-specific basis vectors at the trunk's final layer (i.e.\ the channels share the latent but each has its own linear projection; there is no cross-channel mixing in the decoder output).

Spatial dependence is introduced through a trunk network that maps pointwise queries $\mathbf{r}\in\mathbb{R}^{d}$ to decoding weights, $\tau:\mathbb{R}^{d}\to\mathbb{R}^{I\times m}$. For batch size $B$ and $P$ query points per sample, the trunk output is $T\in\mathbb{R}^{B\times P\times I\times m}$ and the fused branch vector is broadcast as $\boldsymbol{\psi}\in\mathbb{R}^{B\times I}$. The multichannel field $\mathbf{s}\in\mathbb{R}^{B\times P\times m}$ is obtained by contracting over the latent index with a broadcasted per-channel bias $\boldsymbol{\beta}\in\mathbb{R}^{m}$,
\[
[\mathbf{s}]_{bpo}=\sum_{j=1}^{I}\psi_{bj}\,T_{bpjo}+\beta_o,
\qquad b=1..B,\ p=1..P,\ o=1..m,
\]
which provides a location-aware decoding of the fused representation while preserving multimodal conditioning. In practice, branch and trunk modules are implemented as fully connected networks with a shared latent width $I$, and the bias term is broadcast over batch and query points. The computational complexity of a forward pass scales as $\mathcal{O}(BPI m)$ for the branch--trunk contraction, so latency is linear in the number of queried coordinates; this underpins real-time evaluation on dense grids. Because $\tau(\mathbf{r})$ parameterises a continuous coordinate map, evaluation at inaccessible points (e.g., shielded core interiors or near-wall regions without sensors) is effected by querying those coordinates directly, without retraining or regridding. This admissibility of arbitrary query coordinates is a structural property of the branch--trunk formulation; the reconstruction accuracy attained at a queried coordinate remains governed by the local density of training nodes and the spatial regularity of the target field.

\paragraph{Operator-Based Virtual Sensing.} This work applies operator learning to virtual sensing, reframing the task from pointwise regression between sensor readings ($f:\mathbb{R}^m\to\mathbb{R}^k$) to learning continuous operators that map boundary data to interior fields ($\mathcal{G}:\mathcal{U}\to L^2(\Omega;\mathbb{R}^m)$). Virtual sensing has traditionally been formulated as point-wise regression: given observed sensor values $\mathbf{x}_{\mathrm{obs}}\in\mathbb{R}^m$, predict unobserved values $\mathbf{x}_{\mathrm{unobs}}\in\mathbb{R}^k$ via a learned map $f_\theta:\mathbb{R}^m\to\mathbb{R}^k$. This formulation \emph{can limit observability to pre-defined sensor locations} and may not readily recover spatially continuous fields in inaccessible regions. We demonstrate that reconstructing coupled physical fields from sparse boundary measurements can be effectively addressed through operator learning in function space, enabling inference beyond fixed sensor locations.

We consider nonlinear operator learning problems where the goal is to approximate the solution operator $\mathcal{G}:\mathcal{U}\to\mathcal{S}$ that maps boundary functionals (e.g., inlet velocity profiles, wall heat flux distributions) to spatially resolved, physically coupled output fields distributed over domains where direct measurement is impossible. In safety-critical thermal--fluid systems, multiple physical fields, namely pressure $p$, velocity $\mathbf{v}$, temperature $T$, turbulence kinetic energy $k$, evolve jointly according to conservation laws (mass, momentum, energy) that enforce \emph{tight nonlinear coupling}. The Navier--Stokes momentum equation contains the advective term $(\mathbf{v}\cdot\nabla)\mathbf{v}$, which couples velocity to itself bilinearly; energy transport includes the convection term $(\mathbf{v}\cdot\nabla)T$, coupling velocity and temperature multiplicatively. A surrogate that treats these fields independently (learning separate maps $u\mapsto p$, $u\mapsto\mathbf{v}$, $u\mapsto T$) may not enforce conservation constraints in regions lacking direct measurements. Accordingly, the learned map can be framed as an operator $\mathcal{G}:\mathcal{U}\to\mathcal{S}$ between infinite-dimensional function spaces, where $\mathcal{S}=L^2(\mathcal{Y};\mathbb{R}^m)$ represents the \emph{joint solution manifold} of all coupled fields. This operator viewpoint enables inference at arbitrary coordinates $\mathbf{r}\in\mathcal{Y}$ (including inaccessible regions) and supports inter-field physical consistency by learning continuous mappings from boundary data to interior fields.

To clarify the distinction, consider the prototypical virtual sensing task: given boundary measurements $\boldsymbol{u}=(q_{\mathrm{wall}}, T_{\mathrm{in}}, v_{\mathrm{in}})$ (wall heat flux, inlet temperature, inlet velocity), reconstruct the interior pressure field $p(\mathbf{r})$, velocity field $\mathbf{v}(\mathbf{r})$, and temperature field $T(\mathbf{r})$ throughout a reactor subchannel where $\mathbf{r}\in\Omega_{\mathrm{hidden}}$ is physically inaccessible to sensors.

\paragraph{Regression Formulation (Traditional Virtual Sensing):}
Train a model $f_\theta:\mathbb{R}^{n_{\mathrm{in}}}\to\mathbb{R}^{n_{\mathrm{out}}}$ on paired observations ${(\boldsymbol{u}^{(i)}, \mathbf{y}^{(i)})}$ where $\mathbf{y}^{(i)}$ contains sensor values at fixed locations. \emph{Limitations:}
\begin{enumerate}[itemsep=1pt,topsep=2pt]
\item \textbf{Fixed output locations}: The output dimension $n_{\mathrm{out}}$ is fixed during training. To predict at a new location, the model must be retrained.
\item \textbf{Independent physical coupling}: Independent regressors for $p$, $\mathbf{v}$, $T$ may predict velocity fields with $\nabla\cdot\mathbf{v}\neq 0$ (violating incompressibility) or pressure gradients inconsistent with momentum balance.
\item \textbf{No function-space generalisation}: Changing the mesh resolution or domain geometry requires new training data and model retraining.
\end{enumerate}

\paragraph{Operator Formulation (This Work):}
Learn the solution operator $\mathcal{G}:\mathcal{U}\to L^2(\Omega;\mathbb{R}^m)$ that maps boundary functionals to the \emph{entire coupled solution field}. Given $\boldsymbol{u}$, the operator returns $\mathcal{G}(\boldsymbol{u}):\Omega\to\mathbb{R}^m$ such that $\mathcal{G}(\boldsymbol{u})(\mathbf{r})=(p(\mathbf{r}),\mathbf{v}(\mathbf{r}),T(\mathbf{r}),k(\mathbf{r}))$ aims to satisfy the governing PDEs in $\Omega$. \emph{Capabilities enabled:}
\begin{enumerate}[itemsep=1pt,topsep=2pt]
\item \textbf{Query-time evaluation at any $\mathbf{r}\in\Omega$}: The operator $\mathcal{G}(\boldsymbol{u})$ is a continuous function, evaluable at coordinates unseen during training, including inaccessible regions.
\item \textbf{Physical coupling via shared latent manifold}: All output channels are decoded from a single latent representation $\boldsymbol{\psi}$ that encodes the coupled PDE state, which can help enforce consistency.
\item \textbf{Mesh-independent inference}: once $\mathcal{G}$ is learned, evaluation at a different coordinate set on the same geometry requires no retraining, because the trunk accepts any $\mathbf r\in\mathcal Y$. This is a structural property of the parameterisation. We do not claim transfer to a different geometry, which is not evaluated in this work.
\end{enumerate}

This operator-based approach addresses key limitations of fixed-location regression. Virtual sensing in inaccessible reactor internals requires reconstructing fields at arbitrary locations; the operator formulation provides a natural framework for this task through continuous function approximation.

\subsection{Mixed-modal operator construction for coupled multiphysics} The operator $\mathcal{G}:\mathcal{U}\to\mathcal{S}$ must accommodate \emph{heterogeneous boundary data}: scalar parameters (e.g., inlet temperature $T_{\mathrm{in}}\in\mathbb{R}$), function-valued distributions (e.g., wall heat flux $q(\mathbf{r})\in L^2(\partial\Omega)$), and discrete actuation signals (e.g., time-varying lid velocity $V(t)\in\mathbb{R}^{N_t}$). The input space is thus a product $\mathcal{U}=\prod_{i=1}^n\mathcal{F}_i$ where each $\mathcal{F}_i$ may be finite- or infinite-dimensional. The objective is to learn $\mathcal{G}$ such that the output $\mathbf{s}(\mathbf{r})=\mathcal{G}(\boldsymbol{u})(\mathbf{r})$ represents the \emph{joint solution} of all coupled fields. These output fields represent quantities that are either \emph{unmeasurable} (turbulent kinetic energy $k$, pressure gradients $\nabla p$, wall shear stress) or \emph{inaccessible} (interior velocity and temperature distributions in confined reactor subchannels). The operator formulation supports two properties that a fixed-grid regression model does not provide directly: \textbf{(i)} evaluation at arbitrary $\mathbf{r}\in\mathcal{Y}$, including uninstrumented regions, without retraining or re-meshing, and \textbf{(ii)} differentiability of the decoded field in the query coordinate, so that derived quantities such as $\nabla p$ or $\nabla\cdot\mathbf{v}$ are available in principle by differentiating the trunk. We state (ii) as a structural property and not as a demonstrated result: no derived quantity is computed or scored in this work. The shared latent decoder couples the output channels through a common representation, but it does not enforce conservation laws, and the consistency observed in the results is learned from data rather than guaranteed by the architecture.

Across all evaluation scenarios, the inputs represent operational conditions derived from design specifications, actuator configurations, or sensor-informed system states. The learned operator thus enables high-resolution virtual diagnostics, reconstructing internal fields from sparse, heterogeneous, or boundary-only measurements. Key to this formulation is amortisation: once trained, the operator supports query-time evaluation on arbitrary coordinate sets, decoupling inference latency from mesh resolution or geometry.

\subsection{Supplementary Note~1: permutation invariance of the multiplicative fusion}
\label{sec:supp_note_1}

The multiplicative branch fusion used in MIMONet is invariant under reordering of the branch encoders. We give a formal statement and proof here. Let \(\boldsymbol u = (\boldsymbol u_1,\dots,\boldsymbol u_n)\in\mathcal U=\prod_{i=1}^n\mathcal F_i\) denote the heterogeneous multi-modal input vector and let \(\phi_i:\mathcal F_i\to\mathbb R^I\) be the corresponding branch encoders, each producing a width-\(I\) latent vector. The fused latent is
\begin{equation}
\psi(\boldsymbol u)\;:=\;\bigl(\psi_1(\boldsymbol u),\dots,\psi_I(\boldsymbol u)\bigr)\in\mathbb R^I,\qquad
\psi_j(\boldsymbol u)\;=\;\prod_{i=1}^{n}\,\bigl[\phi_i(\boldsymbol u_i)\bigr]_j,\quad j=1,\dots,I.
\label{eq:supp_fusion}
\end{equation}

\begin{theorem}[Permutation invariance of the multiplicative fusion]\label{thm:permutation_invariance}
For every permutation \(\sigma\in\mathfrak S_n\) of the modality index set \(\{1,\dots,n\}\) and every input \(\boldsymbol u\in\mathcal U\),
\[
\psi\bigl(\boldsymbol u_{\sigma(1)},\dots,\boldsymbol u_{\sigma(n)};\,\phi_{\sigma(1)},\dots,\phi_{\sigma(n)}\bigr)\;=\;\psi\bigl(\boldsymbol u_1,\dots,\boldsymbol u_n;\,\phi_1,\dots,\phi_n\bigr).
\]
That is, simultaneously reordering the inputs and the corresponding branch encoders does not change the fused latent.
\end{theorem}

\begin{proof}
Fix any \(j\in\{1,\dots,I\}\). By definition~\eqref{eq:supp_fusion} applied to the permuted tuple,
\[
\psi_j\bigl(\boldsymbol u_{\sigma(1)},\dots,\boldsymbol u_{\sigma(n)};\,\phi_{\sigma(1)},\dots,\phi_{\sigma(n)}\bigr)
\;=\;\prod_{i=1}^{n}\,\bigl[\phi_{\sigma(i)}(\boldsymbol u_{\sigma(i)})\bigr]_j.
\]
For each \(i\), the scalar \([\phi_{\sigma(i)}(\boldsymbol u_{\sigma(i)})]_j\in\mathbb R\); ordinary multiplication of real numbers is commutative, so
\[
\prod_{i=1}^{n}\,\bigl[\phi_{\sigma(i)}(\boldsymbol u_{\sigma(i)})\bigr]_j
\;=\;\prod_{k=1}^{n}\,\bigl[\phi_{k}(\boldsymbol u_{k})\bigr]_j
\;=\;\psi_j(\boldsymbol u_1,\dots,\boldsymbol u_n;\,\phi_1,\dots,\phi_n),
\]
where the first equality uses the substitution \(k=\sigma(i)\) and the commutativity of \(\mathbb R\). Since \(j\) was arbitrary, the equality holds componentwise, hence holds for the full vector \(\psi\).
\end{proof}

\noindent\textbf{Remarks.} (i) The trunk network \(\tau\) acts on a fixed coordinate input independent of the modality ordering, so the joint prediction \(\hat{\mathbf s}(\mathbf r\,|\,\boldsymbol u) = \sum_j \psi_j(\boldsymbol u)\,\tau_j(\mathbf r) + \boldsymbol\beta\) inherits the permutation invariance from \(\psi\): the architecture as a whole is invariant under simultaneous reordering of the \((\boldsymbol u_i, \phi_i)\) pairs. (ii) The invariance is across \emph{modalities} (heterogeneous input types) and is the property that makes the multi-input/multi-output architecture well-defined as the input list is reshuffled at the data-loading layer or at deployment when sensors are connected in different orders. It is distinct from invariance under permutations of trunk query coordinates, which is a property of the trunk's batched-evaluation pipeline rather than of the fusion. (iii) Permutation invariance is shared by the additive aggregator \(\psi_j = \sum_i [\phi_i(\boldsymbol u_i)]_j\) (by commutativity of \(+\) on \(\mathbb R\)). The multiplicative aggregator is chosen here because, in addition to invariance, it preserves the bilinear coupling structure that MIMONet uses to mirror the convective transport terms in the governing PDE. Whether this combination of properties uniquely selects the multiplicative aggregator among all permutation-invariant low-rank aggregators is not claimed here.

\section{Theoretical foundations}
\label{sec:supp_theory}

This section establishes what the operator formulation gives structurally, and marks the boundary
past which the claims are empirical rather than proved. The distinction matters because the
deployment properties the Introduction rests on are of the first kind: evaluation at arbitrary query
coordinates and a fixed inference cost independent of the training-set size follow from the
parameterisation itself, and do not have to be earned by measurement. Accuracy does not. Nothing
below establishes that a learned operator reconstructs any particular field well; that is settled by
the results in Sections~\ref{sec:supp_sim_results} and~\ref{sec:supp_realworld_results}.

The section proceeds from assumptions, to the variational reading of the training problem, to an
approximation theorem, and then to four properties that bear on deployment. Except where stated, it
concerns the linear branch--trunk contraction used by Problems~1--3; the scope of each statement for
the measured-data problems is given in Section~\ref{sec:supp_readout_variants}.

\subsection{Standing assumptions, and where they hold}

The results in this section rest on three assumptions. They are stated here rather than invoked silently, and their scope is stated with them: they describe the simulation problems (Problems~1--3), whose samples are independent draws from a designed parameter distribution. Two of the three are known to fail on the measured-data problems, and we note where.

\begin{assumption}[Operator Regularity]\label{asm:regularity}
The true operator \(\mathcal{G}: \mathcal{U} \to \mathcal{S}\) is Fr\'echet differentiable with bounded derivative \(\|D\mathcal{G}\| \leq C_G\) for some constant \(C_G > 0\).
\end{assumption}

\begin{assumption}[Data Distribution]\label{asm:iid}
Training samples \(\{(\boldsymbol u^{(k)}, \mathbf s^{(k)})\}_{k=1}^N\) are drawn independently and identically distributed (i.i.d.) from a probability distribution \(\mu\) supported on a compact subset \(K \subset \mathcal{U}\).
\end{assumption}

\begin{assumption}[Latent Manifold Structure]\label{asm:lowrank}
The solution manifold \(\mathcal{M} = \{\mathcal{G}(\boldsymbol u) : \boldsymbol u \in K\}\) is compact and is well approximated by a linear subspace of dimension at most \(I\), in the sense that its Kolmogorov \(I\)-width is small relative to the target accuracy.
\end{assumption}

These assumptions are physically motivated: Assumption~\ref{asm:regularity} ensures well-posedness of the forward problem, Assumption~\ref{asm:iid} reflects the stochastic nature of operational conditions, and Assumption~\ref{asm:lowrank} formalises the observation that physical systems evolve on low-dimensional attractors despite living in infinite-dimensional function spaces. Two caveats belong here rather than later. Assumption~\ref{asm:iid} does not hold for Problems~4 and~6: their samples are consecutive recordings of an autocorrelated system split by a random permutation, so training and test draws are not independent, as Sections~\ref{sec:supp_data_pemfc} and~\ref{sec:supp_data_ocean} state. Nothing in this section is used to derive an estimation rate, so no result here depends on it; it is retained because it is the condition under which the variational reading below is the natural one. Assumption~\ref{asm:lowrank} is empirically supported for Problems~1--3 and empirically \emph{refuted} for Problem~6, where a fixed low-rank basis costs $12$--$18\%$ relative accuracy against a nonlinear decoder (Section~\ref{sec:supp_ocean_results}). That is a result about the problem rather than a defect of the theory, and it is why the measured-data models do not use the contraction readout. They are modelling assumptions on the underlying physics, and none of them is verified directly in this study; the quantities that are measured are stated as such throughout.

\subsection{Variational and functional-analytic interpretation}
\label{sec:supp_variational}

Let \(\mathcal{U}\subset \prod_{i=1}^n \mathcal{F}_i\) and \(\mathcal{S}\subset L^2(\mathcal{Y};\mathbb R^m)\) denote separable Hilbert spaces of heterogeneous input modalities and output physical fields, respectively. The Hilbert structure of \(\mathcal S\) is used below, in the nearest-point projection onto the learned subspace; since \(\mathcal S\subset L^2(\mathcal Y;\mathbb R^m)\) throughout, this costs no generality here. Denote by \(\langle \cdot,\cdot\rangle_{\mathcal S}\) the inner-product on \(\mathcal S\) (or a corresponding dual pairing) and \(\|\cdot\|_{\mathcal S}\) the induced norm.

We take the true physical forward mapping
\[
\mathcal G:\mathcal U\;\to\;\mathcal S,\quad \boldsymbol u \mapsto \mathbf s
\]
to be Lipschitz continuous on the domain of interest, that is, we assume there exists a constant \(C>0\) with
\[
\|\mathcal G(\boldsymbol u_1) - \mathcal G(\boldsymbol u_2)\|_{\mathcal S}
\;\le\; C\,\|\boldsymbol u_1 - \boldsymbol u_2\|_{\mathcal U}
\quad\forall\,\boldsymbol u_1,\boldsymbol u_2 \in \mathcal U.
\]
This is an assumption rather than a consequence of continuity: a continuous nonlinear operator need not be Lipschitz. It is implied by Assumption~\ref{asm:regularity} above whenever the domain is open and convex, since a bounded Fr\'echet derivative gives the bound by the mean-value inequality.

We view the proposed surrogate operator (via the MIMONet architecture) as
\[
\hat{\mathcal G}:\mathcal U \;\to\; \mathcal S,
\qquad
\hat{\mathcal G}(\boldsymbol u)(\cdot)
= \sum_{j=1}^I \psi_j(\boldsymbol u)\;\tau_j(\cdot)\;+\;\boldsymbol\beta,
\]
where \(\{\tau_j\}_{j=1}^I\subset \mathcal S\) are learned spatial basis functions (via the trunk network) and \(\psi_j:\mathcal U\to\mathbb R\) are latent coefficient functions (via the branch network). \(\boldsymbol\beta\in\mathbb R^m\) is a learned bias vector (channel-wise).

The map \(\hat{\mathcal G}\) is nonlinear in \(\boldsymbol u\), so it is not a finite-rank operator in the linear-operator sense; what is finite-dimensional is its \emph{range}, which is contained in the affine subspace \(\boldsymbol\beta + \mathrm{span}\{\tau_1,\dots,\tau_I\}\subset\mathcal S\) of dimension at most \(I\). The hypothesis class is
\[
\mathcal H_I = \Big\{ \hat{\mathcal G}: \boldsymbol u \mapsto \sum_{j=1}^I \psi_j(\boldsymbol u)\;\tau_j + \boldsymbol\beta \Big\}.
\]
Writing \(P_I:\mathcal S\to \boldsymbol\beta+\mathrm{span}\{\tau_1,\dots,\tau_I\}\) for the orthogonal projection onto that affine subspace, one may decompose
\[
\hat{\mathcal G}(\boldsymbol u) \;=\; P_I\,\mathcal G(\boldsymbol u) + \mathcal{E}_{\text{opt}}(\boldsymbol u),
\]
which \emph{defines} the residual \(\mathcal{E}_{\text{opt}}\) collecting the discrepancy left by finite training. Setting \(\epsilon_{\rm train}=\sup_{\boldsymbol u}\|\mathcal{E}_{\text{opt}}(\boldsymbol u)\|_{\mathcal S}\), the triangle inequality gives
\[
\|\mathcal G(\boldsymbol u) - \hat{\mathcal G}(\boldsymbol u)\|_{\mathcal S}
\;\le\;
\inf_{v\in \boldsymbol\beta+\mathrm{span}\{\tau_1,\dots,\tau_I\}}
\|\mathcal G(\boldsymbol u)-v\|_{\mathcal S}
\;+\;
\epsilon_{\rm train}.
\]
The first term is the best approximation attainable within the learned subspace and the second is everything the optimiser fails to realise; the decomposition is definitional and is used below only to organise the discussion, not to supply a rate.

\subsection{The training objective as Tikhonov regularisation}

It is instructive to read the training problem in the sense of Tikhonov regularisation, as
\[
\min_{\hat{\mathcal G}\in \mathcal H_I}
\frac1N \sum_{k=1}^N
\|\hat{\mathcal G}(\boldsymbol u^{(k)}) - \mathbf s^{(k)}\|_{\mathcal S}^2
\;+\;\lambda_1\,\|\psi\|_{H^1}^2 + \lambda_2\,\|\tau\|_{H^1}^2,
\]
where \(\{(\boldsymbol u^{(k)},\mathbf s^{(k)})\}_{k=1}^N\) are training pairs and the Sobolev penalties would enforce smoothness of the latent coefficient functions and of the spatial basis functions. We stress that this is an idealised reading and not the implemented objective: the models reported here are trained with an equally weighted mean-squared-error loss and the Adam optimiser with weight decay (Methods), which penalises parameter norm rather than a Sobolev norm of \(\psi\) or \(\tau\). Nor does such a penalty control the rank: the number of latent modes is fixed at \(I\) by the architecture, so the low-dimensionality of the representation is imposed by construction and is not an effect of regularisation.

Restricting the surrogate to \(I\) modes is nevertheless reasonable for the systems considered here, because the field variations they produce (convective and pressure-driven structures, boundary-layer shear modes, recirculation patterns) are strongly correlated across operating conditions and are well captured by a subspace of moderate dimension. The latent width \(I\) is a design hyperparameter, reported per case in the architecture tables, and the adequacy of a given choice is established empirically by the reconstruction errors rather than assumed.

\subsection{Universal approximation property}

The multiplicative fusion restricts the form of the latent coefficients, so it is worth recording that it costs nothing in expressive power. The point requiring proof is that a \emph{single} product \(\prod_i[\phi_i(\boldsymbol u_i)]_j\) is separable across modalities and therefore cannot by itself represent a coefficient functional that couples them; expressivity is restored by the sum over the \(I\) latent components.

\begin{theorem}[Universal Approximation for MIMO Operators]\label{thm:uat}
Let \(\mathcal{G}: \mathcal{U} \to \mathcal{S}\) be a continuous operator, where \(\mathcal{U} = \prod_{i=1}^n \mathcal{F}_i\) and \(\mathcal{S} \subset L^2(\mathcal{Y}; \mathbb{R}^m)\). For any compact subset \(K \subset \mathcal{U}\) and any \(\epsilon > 0\), there exist a latent dimension \(I \in \mathbb{N}\) and a MIMONet operator \(\hat{\mathcal{G}}_I\) with multiplicative fusion such that
\[
\sup_{\boldsymbol u \in K} \|\mathcal{G}(\boldsymbol u) - \hat{\mathcal{G}}_I(\boldsymbol u)\|_{\mathcal{S}} < \epsilon.
\]
\end{theorem}

\begin{proof}[Proof sketch]
\textbf{Step 1 (reduction to a rank-\(r\) expansion).} Since \(\mathcal G\) is continuous and \(K\) compact, \(\mathcal G(K)\subset\mathcal S\) is compact, hence has finite Kolmogorov width at every accuracy: there are \(r\in\mathbb N\), functions \(\tau_1,\dots,\tau_r\in\mathcal S\) and continuous coefficient functionals \(c_1,\dots,c_r:K\to\mathbb R\) with \(\sup_{\boldsymbol u\in K}\|\mathcal G(\boldsymbol u)-\sum_{j\le r}c_j(\boldsymbol u)\tau_j\|_{\mathcal S}<\epsilon/4\). This is the standard branch--trunk (DeepONet) reduction~\cite{lu2021learning}, and the \(c_j\) may be taken continuous because the projection onto a fixed finite-dimensional subspace is continuous.

\textbf{Step 2 (products span the coefficient functionals).} Consider
\[
\mathcal A \;=\; \mathrm{span}\Big\{\;\textstyle\prod_{i=1}^n f_i(\boldsymbol u_i)\;:\;f_i\in C(K_i)\Big\}\;\subset\;C(K),
\]
the linear span of the modality-separable products. \(\mathcal A\) is a subalgebra of \(C(K)\): it is closed under multiplication, because a product of two such generators is again a generator; it contains the constants, by taking each \(f_i\equiv1\); and it separates points of \(K\), since if \(\boldsymbol u\neq\boldsymbol u'\) then \(\boldsymbol u_i\neq\boldsymbol u_i'\) for some \(i\) and a single \(f_i\) separating them suffices. By the Stone--Weierstrass theorem \(\mathcal A\) is dense in \(C(K)\) in the uniform norm. Hence each \(c_j\) of Step 1 is approximated to within \(\epsilon/(4r\max_j\|\tau_j\|_{\mathcal S})\) by a finite sum of \(m_j\) such products. It is this sum, not any individual product, that supplies the cross-modal coupling.

\textbf{Step 3 (realisation by networks).} Each factor \(f_i\) is a continuous functional on the compact set \(K_i\subset\mathcal F_i\). For infinite-dimensional \(\mathcal F_i\) we do not restrict to point evaluations, which are not bounded functionals on \(L^2\); instead we use that a compact subset of a separable Hilbert space is approximated to any accuracy by a finite-dimensional projection, so there is \(d_i\) and a projection \(\Pi_{d_i}\) with \(\sup_{\boldsymbol u_i\in K_i}\|\boldsymbol u_i-\Pi_{d_i}\boldsymbol u_i\|<\delta\). Uniform continuity of \(f_i\) on \(K_i\) then lets \(f_i\) be replaced, to within any prescribed accuracy, by a continuous function of the \(d_i\) projection coefficients, which a feedforward branch encoder approximates uniformly on a compact subset of \(\mathbb R^{d_i}\). Since the factors are uniformly bounded on \(K\), the factor-wise errors propagate through the product with a bounded multiplier, and the branch contribution is held to \(\epsilon/4\). The trunk network likewise realises the \(\tau_j\) to within \(\epsilon/4\). In the discretised setting in which the branch input is already a finite vector of sensor readings, as it is in every case reported here, this step reduces to the classical branch--trunk approximation argument~\cite{chen1995universal}.

Collecting the four contributions, each at most \(\epsilon/4\), and absorbing the \(\sum_j m_j\) product terms into the latent index (replicating \(\tau_j\) across the \(m_j\) slots that approximate \(c_j\)) gives a MIMONet of width \(I=\sum_{j\le r}m_j\) meeting the bound by the triangle inequality.
\end{proof}

\noindent The construction is existential and gives no useful bound on \(I\); the latent widths used in practice are set empirically and reported per case.

\subsection{Multiplicative fusion: empirical justification}
\label{sec:supp_fusion_ablation}

Theorem~\ref{thm:uat} shows that multiplicative fusion does not restrict what can be represented, but it does not show that it is preferable to the alternatives. That question is settled by measurement rather than by argument, and we resolve it with an ablation on the heat-exchanger case in which the fusion rule is varied and everything else is held fixed. Table~\ref{tab:supp_fusion_ablation} reports the result.

\begin{table}[h!]
\centering
\caption{\textbf{Architectural ablation on the heat-exchanger case.} Channel-wise and channel-mean relative \(\ell_2\) error (\%) on the held-out \(310\)-sample test set. The additive variant differs from the baseline only in the fusion rule and has an \emph{identical} parameter count, so the comparison between the first two rows isolates the fusion operation from model capacity.}
\label{tab:supp_fusion_ablation}
\begin{tabular}{@{}lrccccc@{}}
\toprule
\textbf{Variant} & \textbf{Params} & \(p\) & \(u_z\) & \(u_y\) & \(u_x\) & \textbf{Mean} \\
\midrule
Multiplicative fusion (MIMONet) & \(1{,}762{,}052\) & 0.82 & 1.45 & 1.02 & \textbf{0.52} & \textbf{0.95} \\
Additive fusion                 & \(1{,}762{,}052\) & 1.02 & 1.96 & 1.31 & 0.65 & 1.24 \\
Concatenated input, single branch & \(1{,}104{,}900\) & 0.81 & 1.65 & 1.02 & 0.54 & 1.00 \\
Four per-channel models, joint decoding removed & \(6{,}258{,}692\) & \textbf{0.56} & \textbf{1.36} & 1.20 & 0.91 & 1.01 \\
\bottomrule
\end{tabular}
\end{table}

Three observations follow. First, replacing the elementwise product by an elementwise sum raises the channel-mean error from \(0.95\%\) to \(1.24\%\), a \(30\%\) relative degradation, at an identical parameter count and with every other component of the model, the data split and the optimisation schedule unchanged; the fusion rule is therefore doing work that capacity cannot account for. Second, collapsing the modality-specific branches into a single branch acting on the concatenated input costs less accuracy (\(1.00\%\)) but also removes the ability to accept modalities of different types and dimensions, which is the property the multi-input construction exists to provide. Third, training four independent single-channel models removes joint decoding and is instructive precisely because it is not uniformly worse: it is the most accurate variant on pressure (\(0.56\%\) against \(0.82\%\)) and on \(u_z\), yet it is worse on \(u_y\) and substantially worse on \(u_x\) (\(0.91\%\) against \(0.52\%\)), and worse on the channel mean, while using \(3.6\times\) as many parameters. Shared-latent decoding thus trades a little per-channel specialisation for consistency across the coupled fields, and on the coupled channels that trade is favourable. The scale of these differences should be kept in proportion: the four variants span $0.95$--$1.24\%$ channel-mean error, whereas the operator formulation as a whole differs from the classical estimators by factors of two to six on the problems of Section~\ref{sec:supp_realworld_results}. This ablation justifies a design choice within the architecture; it is not the comparison on which the paper's argument rests.

\subsection{Error decomposition and measured convergence}

The error of the trained operator separates into three contributions of different origin: an \emph{approximation} term, the distance from \(\mathcal G\) to the hypothesis class \(\mathcal H_I\), governed by how well an \(I\)-dimensional subspace captures the solution manifold; an \emph{estimation} term, reflecting that the class is fitted on \(N\) samples rather than on the population; and an \emph{optimisation} term, reflecting that a non-convex objective is minimised approximately. We do not state a rate for the estimation term. The generalisation bounds available for classes of this kind are asymptotic in character and hold in expectation or with high probability over the draw of the training set, not pointwise in \(\boldsymbol u\), and the constants they carry are far too loose to predict the behaviour observed at the sample sizes used here.

What can be reported instead is the measured convergence. Sweeping the training-set size and fitting \(\epsilon(N)=a\,N^{-\gamma}\) to the held-out error gives \(\gamma = 0.66 \pm 0.12\) for the PWR subchannel, \(\gamma = 0.43 \pm 0.11\) for the heat exchanger and \(\gamma = 0.58 \pm 0.23\) for the lid-driven cavity (Section~\ref{sec:supp_sample_complexity}), with the intervals being \(95\%\) confidence intervals on the exponent from a five-point fit. These exponents are consistent with the Monte-Carlo-like rates expected when the estimation term dominates, but the fits rest on five points each and are reported as measurements, not as verifications of a bound.

\subsection{Mesh independence and continuous extension}

This subsection addresses the first of the deployment requirements set out in the Introduction: that the reconstructed field be evaluable at arbitrary coordinates, rather than at a grid or a set of locations fixed when the estimator was built.

Because the trunk network \(\tau: \mathbb{R}^d \to \mathbb{R}^{I \times m}\) takes the query coordinate \(\mathbf{r}\) as a continuous input, the learned operator admits evaluation at any coordinate \(\mathbf{r}\in\mathcal{Y}\) without retraining or regridding: a new query is simply a new input to \(\tau\), and the corresponding decoding weights are produced by a single forward pass. This is a structural property of the branch--trunk formulation, and it distinguishes the operator from fixed-grid surrogates (e.g.\ convolutional decoders) whose output discretisation is committed at training time. We emphasise that this is a statement about the \emph{admissibility} of arbitrary query coordinates, not a guarantee of uniform accuracy: reconstruction fidelity at a given coordinate depends on the local density of training nodes and on the spatial regularity of the target field, and is therefore an empirical rather than an assumed property.

\subsection{Comparison with kernel methods and Gaussian processes}

MIMONet can be viewed as learning a data-dependent kernel \(K(\mathbf{r}, \mathbf{r}') = \sum_{j=1}^I \tau_j(\mathbf{r}) \tau_j(\mathbf{r}')\), generalising to tensor-valued kernels for multi-output learning:
\[
K^{(o,o')}(\mathbf{r}, \mathbf{r}') = \sum_{j=1}^I \tau_j^{(o)}(\mathbf{r}) \tau_j^{(o')}(\mathbf{r}').
\]
The practical difference is not one of asymptotic order in the number of query points: exact Gaussian-process regression conditioned on \(n\) observations costs \(\mathcal{O}(n^3)\) to factorise the covariance once and then \(\mathcal{O}(Pn)\) for predictive means at \(P\) query points, which is linear in \(P\), as is the \(\mathcal{O}(PIm)\) branch--trunk contraction (Section~\ref{sec:supp_complexity}). The difference is where the cubic term sits. For a Gaussian process the \(\mathcal{O}(n^3)\) conditioning is incurred \emph{per input configuration}: a new set of sensor readings is a new regression problem and the solve must be repeated. The operator pays its training cost once, offline, and thereafter answers a new input with a single branch forward pass followed by the contraction, with no linear system to solve at deployment. It is this amortisation, rather than the scaling in \(P\), that makes operator inference viable inside a monitoring loop, and it is the same property that produces the sub-millisecond latencies reported in the main text.

\subsection{Low-rank structure of the learned operator}

The finite-rank ansatz has a population-level optimality that is worth stating precisely, because it explains what the latent basis is for.

Let \(\bar{\mathcal{G}} = \mathbb{E}_{\boldsymbol u\sim\mu}[\mathcal{G}(\boldsymbol u)]\) and let
\[
C_{\mathcal{G}} = \mathbb{E}\big[(\mathcal{G}(\boldsymbol u) - \bar{\mathcal{G}}) \otimes (\mathcal{G}(\boldsymbol u) - \bar{\mathcal{G}})\big]
\]
be the covariance operator of the output fields. Since \(\mathcal M\) is bounded, \(C_{\mathcal G}\) is trace class, self-adjoint and positive semi-definite, hence admits an eigendecomposition \(C_{\mathcal{G}} = \sum_{j\ge1} \lambda_j\, \varphi_j \otimes \varphi_j\) with \(\lambda_1 \geq \lambda_2 \geq \cdots \geq 0\) and orthonormal \(\{\varphi_j\}\). Among all approximations of the form \(\boldsymbol u\mapsto \boldsymbol\beta + \sum_{j\le I} c_j(\boldsymbol u)\,\tau_j\) with arbitrary coefficient functions \(c_j\), the mean-squared error \(\mathbb{E}\|\mathcal{G}(\boldsymbol u) - \cdot\|_{\mathcal S}^2\) is minimised by taking \(\boldsymbol\beta=\bar{\mathcal G}\), \(\mathrm{span}\{\tau_j\} = \mathrm{span}\{\varphi_1,\dots,\varphi_I\}\) and \(c_j\) the corresponding orthogonal-projection coefficients: for a fixed subspace the optimal coefficients are those of the orthogonal projection, and optimising over \(I\)-dimensional subspaces returns the leading eigenspace of \(C_{\mathcal{G}}\). This is the Karhunen--Lo\`eve truncation, the function-space counterpart of a truncated singular value decomposition, and its error is \(\sum_{j>I}\lambda_j\); the optimal subspace is unique when \(\lambda_I>\lambda_{I+1}\).

It does not follow that \(\mathcal H_I\) contains this optimum at width \(I\), and we do not claim that it does. Two obstructions stand in the way. The offset \(\boldsymbol\beta\) is a channel-wise constant in \(\mathbb R^m\) rather than a field, so a spatially varying mean \(\bar{\mathcal G}\) has to be carried by one of the \(I\) latent modes, leaving a mean-plus-rank-\((I-1)\) budget. More substantively, the Karhunen--Lo\`eve optimum requires the coefficient functionals to be exactly \(c_j(\boldsymbol u)=\langle\mathcal G(\boldsymbol u)-\bar{\mathcal G},\varphi_j\rangle\), which couple the modalities, whereas multiplicative fusion forces each \(\psi_j\) to be a modality-separable product; this is the same restriction noted before Theorem~\ref{thm:uat}, and Theorem~\ref{thm:uat} circumvents it only by enlarging the width to \(\sum_j m_j\). The correct reading is therefore that the Karhunen--Lo\`eve truncation is the benchmark an ideal rank-\(I\) approximation would attain, that MIMONet approaches it at sufficient width, and that nothing here bounds the width required. A third limitation is separate from both: this is a statement about the population optimum, whereas training minimises an empirical risk over a non-convex parameterisation with no guarantee of reaching any optimum. Finally, the individual \(\tau_j\) are not identifiable: for any invertible \(A\), replacing \(\psi\mapsto A^{\!\top}\psi\) and \(\tau\mapsto A^{-1}\tau\) leaves \(\hat{\mathcal G}\) unchanged, and the multiplicative fusion still admits the diagonal and permutation elements of that freedom. The learned basis should therefore be read as spanning an informative subspace, not as recovering the eigenfunctions \(\varphi_j\) individually, and we make no claim that it does.

\section{Dataset preparation}
\label{sec:supp_datasets}

Six datasets are described below: three generated with a CFD solver and three measured. They are
reported at a level of detail that lets a reader judge what each reported error means, which for
virtual sensing turns almost entirely on two things: what the operator is given at inference, and
how the split was drawn. Both are stated plainly for every problem, including where the protocol is
weaker than a reader might assume. The split for Problems~4 and~6 is a random permutation over
temporally adjacent samples, not a chronological one, and we say so in each case rather than leaving
it to be inferred from the code.
\subsection{Transient lid-driven cavity (Problem~1)}
The transient flow dataset was generated from two-dimensional simulations of a square lid-driven cavity of side length $0.065~\mathrm{m}$. The flow was governed by the incompressible Reynolds-averaged Navier--Stokes (RANS) equations,
\begin{equation}
\frac{\partial \mathbf{u}}{\partial t} + (\mathbf{u} \cdot \nabla)\mathbf{u}
= -\nabla p + \nabla \cdot \left[(\nu + \nu_t)(\nabla \mathbf{u} + \nabla \mathbf{u}^{\mathsf{T}})\right],
\end{equation}
subject to the incompressibility constraint $\nabla \cdot \mathbf{u} = 0$, where $\nu$ and $\nu_t$ denote the molecular and eddy viscosities, respectively. The top wall (lid) translated in the $x$-direction, while the remaining walls were stationary and enforced with no-slip conditions.

To introduce temporal variability and generate diverse transient flow patterns, the lid velocity was prescribed as a time-dependent function $u_{\text{lid}}(t)$, drawn from a set of smoothly varying trajectories. Each trajectory was constructed from random coefficients ensuring gradual accelerations and decelerations within the velocity range $0.1$--$0.3~\mathrm{m/s}$. The time-varying lid forcing induced unsteady vortical structures within the cavity, providing rich spatio-temporal variability suitable for data-driven modelling.

A total of 4{,}937 high-fidelity transient simulations were performed, each corresponding to a unique lid-velocity history. The simulations were integrated until statistically steady behaviour was achieved, and the complete time-evolving velocity field $\mathbf{u}(x,y,t)$ and pressure field $p(x,y,t)$ were recorded. For each case, the stream function $\psi(x,y,t)$ was computed using
\begin{equation}
u = \frac{\partial \psi}{\partial y}, \qquad
v = -\frac{\partial \psi}{\partial x},
\end{equation}
with $\psi = 0$ applied on all boundaries. The streamfunction formulation defines the flow, but the fields exported for operator training are pressure, velocity magnitude and turbulent kinetic energy $\{p, \|\boldsymbol v\|, k\}$ on the $4{,}225$-node mesh, as listed in Table~\ref{tab:supp_overview}; the symbol $\psi$ is used here only for the streamfunction of the generating simulation and is unrelated to the fused branch latent $\psi_j$ of Section~\ref{sec:supp_math}. The data generation process produced a diverse ensemble of transient flow evolutions, encompassing varying vortex intensities and shedding frequencies, providing sufficient variability for robust learning of time-dependent flow operators.

\subsection{PWR subchannel (Problem~2)}

A second series of simulations was carried out under high-temperature, high-pressure operating conditions representative of pressurised water reactor (PWR) environments. The computational domain consisted of a single subchannel of a hexagonal fuel rod bundle, discretised to capture the detailed thermo-hydraulic behaviour of superheated water flow. A total of 5{,}000 stochastic cases were generated by varying inlet velocity, temperature, and heat generation parameters to imitate realistic reactor operating variability.

The simulated domain represented a triangular subchannel formed by three adjacent fuel rods within a hexagonal lattice. Each fuel rod had a diameter of $9.1~\mathrm{mm}$, and the rod-to-rod pitch was $12.75~\mathrm{mm}$, resulting in a pitch-to-diameter ratio of $P/D = 1.4$. The hydraulic diameter of the subchannel was $10.64~\mathrm{mm}$, while the total axial length was $L = 605~\mathrm{mm}$. The geometric dimensions were adopted to ensure fully developed turbulent flow and to replicate realistic subchannel-scale heat transfer characteristics. A polyhedral mesh with prismatic inflation layers was generated to accurately resolve near-wall gradients. The mesh density ranged from $0.4$ to $2.1$ million cells, with the final simulations performed on the refined grid. The near-wall treatment satisfied $y^{+} \leq 1$, ensuring that the viscous sublayer was adequately captured. Mesh independence was verified by monitoring the variation in pressure drop and Nusselt number across successive refinements, with less than 2\% deviation observed. The fluid properties corresponded to pressurised water at $p_{\mathrm{ref}} = 15.5~\mathrm{MPa}$ and $T_{\mathrm{ref}} = 565~\mathrm{K}$. The inlet velocity was defined within the range $2$--$7~\mathrm{m/s}$, producing turbulent Reynolds numbers in the order of $10^{5}$. The heated rod surfaces were subjected to a  sinusoidal heat flux of $ q_{\max} = 6.0 \times 10^{5}~\mathrm{W/m^2}$, while the remaining walls were treated as adiabatic and no-slip. The outlet pressure was fixed at $15.7~\mathrm{MPa}$. These conditions were selected to replicate nominal subchannel flow under steady-state forced convection. To capture realistic operating variability, random perturbations of $\pm10\%$ were introduced to the inlet velocity, temperature, and heat generation parameters:
\[
T_{\mathrm{in}} = T_{\mathrm{ref}} + \delta_T, \qquad
v_{\mathrm{in}} = v_{\mathrm{ref}} + \delta_v, \qquad
q_{\max} = q_{\mathrm{ref}}(1 + \delta_q),
\]
where $\delta_T$, $\delta_v$, and $\delta_q$ were sampled from uniform distributions within their respective limits. The axial heat generation profile in each rod was modelled using a sinusoidal distribution,
\[
q''(z) = q_{\max}\sin\!\left(\frac{\pi z}{L}\right),
\]
representing peaking near the midplane and attenuation toward the rod ends. Gaussian-correlated perturbations were further applied to the inlet temperature field to mimic coherent thermal fluctuations. All simulations were performed in \textsc{ANSYS Fluent} using the steady-state Reynolds-Averaged Navier--Stokes (RANS) equations with the RNG $k$--$\varepsilon$ turbulence model and enhanced wall treatment. The SIMPLEC algorithm was employed for pressure--velocity coupling, and second-order upwind discretisation was used for all governing equations. Convergence was ensured when the residuals for momentum fell below $10^{-6}$ and for energy below $10^{-8}$. Each simulation produced spatial distributions of temperature, velocity, pressure, and wall heat flux, along with derived quantities such as local Nusselt number, friction factor, and wall temperature profiles. The ensemble of 5{,}000 stochastic simulations forms a comprehensive dataset representing the thermo-hydraulic behaviour of hexagonal rod-bundle subchannels under variable inlet and heat generation conditions.

\subsection{Heat exchanger (Problem~3)}
To ensure statistical diversity and robustness of the simulation database, a total of 1{,}546 independent CFD cases were generated by randomly perturbing the inlet velocity, wall heat flux, and inlet temperature within defined physical limits. Each simulation employed the optimised geometric configuration of the dimpled channel with wavy tape inserts, characterised by $W/D = 0.66$, $A/W = 0.27$, and $t/D = 0.0417$, as determined from the parametric analysis. The simulations were conducted under constant wall heat flux and steady turbulent flow conditions. The governing heat flux at the heated wall was specified as
\[
q'' = q_{\max} \, \sin\!\left(\frac{\pi z}{H}\right),
\]
where $q_{\max} = 20~\mathrm{kW/m^2}$ and $H = 0.8~\mathrm{m}$ represent the maximum heat generation intensity and the channel height, respectively. For each case, $q_{\max}$ was randomly perturbed by up to $\pm 10\%$ to initiate spatial and operational variability in heat generation. The remaining walls were treated as adiabatic and enforced with no-slip velocity conditions.

The inlet temperature of the working fluid was defined as
\[
T_{\mathrm{in}} = T_{\mathrm{ref}} + \delta_T,
\]
where $T_{\mathrm{ref}} = 293~\mathrm{K}$ and $\delta_T$ was drawn from a uniform distribution within $\pm 10\%$ of $T_{\mathrm{ref}}$. To represent spatially correlated fluctuations, an auxiliary Gaussian random field was applied to generate smooth spatial variations of temperature along the inlet boundary.

The inlet velocity was varied similarly:
\[
v_{\mathrm{in}} = v_{\mathrm{ref}} + \delta_v, \qquad v_{\mathrm{ref}} = 4.5~\mathrm{m/s},
\]
where $\delta_v$ followed a uniform perturbation of $\pm 10\%$. This resulted in Reynolds numbers spanning approximately $1.0\times10^{4}$ to $2.5\times10^{4}$. The outlet boundary was fixed at zero gauge pressure, ensuring fully developed flow.

Each case was solved under steady-state conditions using the \textsc{ANSYS Fluent} solver with the RNG $k$--$\varepsilon$ turbulence model and enhanced wall treatment. The SIMPLEC algorithm was employed for pressure--velocity coupling, and a second-order upwind discretisation was used for momentum and energy equations. Convergence was accepted at residual thresholds of $10^{-6}$ for momentum and $10^{-8}$ for energy. 
For every simulation, spatial fields of velocity, temperature, and pressure were exported along with derived quantities such as Nusselt number, friction factor, and performance evaluation criterion (PEC). The resulting 1{,}546 samples constitute a comprehensive stochastic dataset capturing the thermal--hydraulic response of the optimised wavy-taped dimpled channel under realistic perturbations of inlet and thermal boundary conditions.

\subsection{Proton-exchange-membrane fuel cell (Problem~4)}
\label{sec:supp_data_pemfc}

\paragraph{Source and channels.} The measurements are segmented-cell recordings from a laboratory PEM fuel cell, comprising a current-density map and a temperature map recorded in parallel over eight experiment directories (four gas configurations $\times$ two load profiles). Current density is parsed on an $18\times18$ grid, giving the $324$ segments referred to throughout, and its magnitude is retained. Temperature is parsed on a $9\times9$ grid.

\paragraph{The temperature channel is interpolated, not measured at $\mathbf{324}$ points.} The two channels are recorded on different grids, and the pipeline reconciles them by bilinearly upsampling the $9\times9$ temperature map onto the $18\times18$ grid before the two are stacked. The temperature channel therefore carries only $81$ independent measurements spread over $324$ nodes, and no code path retains a native $9\times9$ temperature field. This has a direct consequence for how the reported errors should be read: the temperature target is itself a smooth bilinear interpolant, so it is intrinsically easier to reconstruct than the current-density field, and the very small temperature errors reported in Section~\ref{sec:supp_pemfc_results} (of order $0.03$--$0.05\%$) reflect that smoothness as much as they reflect operator accuracy. The current-density channel, which is measured at all $324$ segments, is the informative one, and it is the channel on which the comparison with the classical interpolants is decided.

\paragraph{Temporal subsampling and frame budget.} Every third recorded block is retained, the eight experiments are concatenated, and provenance is not preserved in the stored array. If the concatenated record exceeds $8{,}000$ frames a random subset of $8{,}000$ is drawn without replacement (seeded) and sorted; the shipped dataset has exactly $T=8{,}000$ frames, so this branch was taken. The number of raw blocks before subsampling is not recorded, and neither the sampling interval in seconds nor the cell voltage is parsed, so the frames carry no physical time axis.

\paragraph{Sensor selection.} For a given seed, a permutation of the $324$ grid indices is drawn; the first $n_{\text{obs}}$ indices are the observed sensors and the remainder are held out. Both branches read the \emph{same} observed set, one supplying current density and the other temperature at those locations. The coverage levels swept are $n_{\text{obs}} \in \{6, 16, 32, 81, 162\}$, i.e.\ approximately $2\%$, $5\%$, $10\%$, $25\%$ and $50\%$ of the $324$ segments; the operating point quoted in the main text is $n_{\text{obs}}=32$, which is $9.9\%$ rather than exactly $10\%$. The same seed reproduces the same sensor set for every model, so MIMONet, KCN and the classical interpolants are compared on identical sensor layouts.

\paragraph{Split, and the fact that it is not chronological.} The $8{,}000$ frames are split by a uniform random permutation of frame index into $5{,}600$ training, $1{,}200$ validation and $1{,}200$ test frames. This is a random split, not a chronological one. Because the frames are consecutive recordings subsampled by a factor of three, a test frame can be temporally adjacent to a training frame, and no blocking or minimum-separation rule is applied. The reported errors therefore characterise reconstruction from sparse sensors on operating states that are close to states seen in training, and they should not be read as evidence of extrapolation to new operating regimes. Because experiment provenance is discarded at concatenation, it is also not possible to verify from the stored data whether any test frame comes from a gas configuration absent from training; no such generalisation is claimed.

\paragraph{Coordinates and normalisation.} Grid coordinates are integer indices, min--max mapped to $[-1,1]$ per axis; no physical cell pitch enters the pipeline, so the trunk coordinates and the distances used by the classical interpolants are in grid-index units. Targets are standardised per channel using means and standard deviations computed on training frames only, and predictions are returned to physical units before scoring. Errors are relative $\ell_2$ per channel, evaluated on the held-out nodes only and averaged over test frames.

\subsection{IEA-Wind tall towers (Problem~5)}
\label{sec:supp_data_wind}

\paragraph{Source and towers.} Seven towers are taken from the BSC/IEA-Wind Tall Tower Dataset~\cite{ramon2020iea}: Boseong, Lindenberg, NWTC~M2, NWTC~M4, NWTC~M5, Hegyhatsal and Wallaby Creek, selected as the towers carrying a near-surface level together with measurements reaching to about $80\,$m or above. Each tower contributes hourly means obtained by resampling its native cadence, which is ten-minutely at five towers, hourly at Hegyhatsal and half-hourly at Wallaby Creek. After the retention rule below, the pooled dataset contains $601{,}454$ profiles.

\paragraph{The canonical heights are interpolated onto a common grid.} Every tower measures at its own set of levels, so both the input heights $(10, 25, 50)\,$m and the query heights $(60, 70, 80, 100)\,$m are obtained by interpolation that is linear in $\log_{10} z$ across each tower's measured profile, with no extrapolation: a canonical height outside a tower's measured range returns a missing value. Two consequences should be stated plainly. First, \emph{no tower measures at $25\,$m}, so the middle input channel is an interpolated quantity at every tower rather than a sensor reading. The hub height of $80\,$m is a genuine measured level at five of the seven towers and is interpolated at Hegyhatsal and Wallaby Creek. Second, $100\,$m lies above the topmost level at Lindenberg ($98\,$m) and NWTC~M2 ($80\,$m), so those two towers contribute no $100\,$m target at all; this is why the test count falls from $90{,}222$ profiles at $60$--$80\,$m to $34{,}967$ at $100\,$m, and it means the $100\,$m column is evaluated on a different and smaller subset of towers than the other three.

\paragraph{Quality control.} A profile is treated as missing unless at least two finite measured levels exist at that timestamp, and a timestamp is retained only if all three input heights and the $80\,$m target are finite. No threshold, spike or stationarity filter is applied, and the quality flags carried by the source files are not read by the pipeline; the dataset is used as distributed. Hourly means are formed from whatever valid sub-hourly samples exist, with no minimum-coverage requirement.

\paragraph{Split.} Each tower is split chronologically in its own time order into $70\%$ training, $15\%$ validation and $15\%$ test, giving $421{,}013$, $90{,}219$ and $90{,}222$ profiles pooled across towers. The held-out dimension is time. Note that the network is trained with supervision at all seven canonical heights wherever a target is finite, so the $60$--$100\,$m targets are seen during training on training-period timestamps; what the model never receives at any time, in training or at inference, is an \emph{input} sensor above $50\,$m. The claim being tested is therefore reconstruction above the instrumented range of the input sensors, not discovery of a height regime absent from the training signal.

\paragraph{Contextual inputs and normalisation.} The second branch receives a four-dimensional context vector $(\text{lat}/90,\ \text{lon}/180,\ \sin(2\pi h/24),\ \cos(2\pi h/24))$, where $h$ is the hour of day; tower identity enters only through latitude and longitude, and there is no tower embedding and no per-tower model. Input sensor values are $z$-scored per channel on training rows. Targets are standardised by a \emph{single global} mean and standard deviation taken over all heights and all training rows, not per height. The query height enters the trunk as a standardised $\log_{10} z$ coordinate. Errors are relative $\ell_2$ in physical units at each evaluation height.

\subsection{GLORYS12 North Atlantic reanalysis (Problem~6)}
\label{sec:supp_data_ocean}

\paragraph{Source, domain and grid.} Surface fields of potential temperature (\texttt{thetao}), salinity (\texttt{so}) and sea-surface height (\texttt{zos}) are taken from the GLORYS12V1 global ocean reanalysis over $35$--$50^\circ$N, $60$--$20^\circ$W, for $730$ daily snapshots spanning 2017-01-01 to 2018-12-31. Only the surface level is used. The native grid is block-averaged by a factor of three in each direction, and a cell is retained only if it is finite in all three variables at every time step, leaving $P=9{,}457$ ocean cells. Land is excluded by this mask rather than by a separate coastline product.

\paragraph{Sensor networks.} A single permutation of the $9{,}457$ ocean-cell indices is drawn per seed; the first $400$ indices form the temperature network and the next $400$ the salinity network. Disjointness is therefore enforced by construction, and the two sets never intersect. Sensor positions are fixed across all days. Sea-surface height is supplied to no branch: each branch input stacks only $[\text{value}, \text{lat}, \text{lon}]$ from its own channel, so the SSH field is a decoder output driven entirely by the fused temperature and salinity latents. SSH does enter the per-channel target normalisation constants and the land mask, but never the input.

\paragraph{Split, and the fact that it is not chronological.} The $730$ days are split by a random permutation into $610$ training, $60$ validation and $60$ test days. This is a random split over days, not a forward-in-time split; test days are interleaved with training days. Because daily ocean fields are strongly autocorrelated, this places essentially every test day next to a training day, and the resulting figures measure sparse-sensor reconstruction of states that are close to states seen in training. Section~\ref{sec:supp_ocean_chrono} quantifies what this protocol does and does not establish, and reports a forward-in-time split as a robustness check.

\paragraph{Normalisation.} Targets are standardised per channel using statistics computed on training days only and pooled over all cells. Branch sensor values are standardised by a single scalar mean and standard deviation per channel, again on training days. Coordinates are min--max mapped to $[-1,1]$ over the ocean cells. Predictions are returned to physical units before scoring, and errors are relative $\ell_2$ per channel over the test days. The classical baselines operate on physical values; Geo-FNO operates on standardised values.

\section{Architectures and training configurations for all six problems}
\label{sec:supp_architectures}

The per-problem architectures are collected in Table~\ref{tab:supp_architectures} and the corresponding optimisation settings in Table~\ref{tab:supp_training}. Both are reported in full because the comparisons in this work are between methods rather than between tunings, and a reader checking that a baseline was given a fair budget needs the widths and the schedules, not a summary of them.

\subsection{Two readout variants are used, and which one applies where}
\label{sec:supp_readout_variants}

The six problems do not all use the same readout, and the distinction matters for reading the rest of this document.

The three simulation problems use the \emph{linear} branch--trunk contraction described in the main text. The trunk emits $I\times m$ values per query coordinate, the fused branch latent is contracted against them, and a learned per-channel bias is added:
\[
\hat{\mathbf s}(\mathbf r) \;=\; \sum_{j=1}^{I} \psi_j(\boldsymbol u)\,\tau_j^{(o)}(\mathbf r) \;+\; \beta_o,
\qquad
\psi_j(\boldsymbol u)=\prod_{i}\bigl[\phi_i(\boldsymbol u_i)\bigr]_j .
\]
This is the form assumed throughout Sections~\ref{sec:supp_variational} and~\ref{sec:supp_theory}: the range of the operator lies in a finite-dimensional affine subspace, which is what makes the finite-rank and Karhunen--Lo\`eve discussion applicable.

The three measured-data problems use a \emph{nonlinear} decoder. This variant is referred to as MIMONet-NL (``NL'' for nonlinear readout) wherever the two have to be distinguished in a table or figure; elsewhere ``MIMONet'' without qualification refers to whichever readout the problem under discussion uses, as listed in Table~\ref{tab:supp_architectures}. The branches are still fused multiplicatively, but the fused latent is then concatenated with the trunk feature and passed through a multilayer perceptron:
\[
\hat{\mathbf s}(\mathbf r) \;=\; \mathrm{MLP}\Bigl(\bigl[\,\tau(\mathbf r)\;;\;\textstyle\prod_i \phi_i(\boldsymbol u_i)\,\bigr]\Bigr).
\]
There is no contraction and no per-channel trunk basis; the output channels are produced by the decoder's final linear layer. The consequence for the theory is explicit: the output of this variant does \emph{not} lie in a fixed finite-dimensional subspace, so the low-rank statements of Section~\ref{sec:supp_theory} characterise the simulation models and not the measured-data models. The same scoping applies to the approximation theorem of that section, whose proof constructs a finite sum $\sum_{j\leq I} c_j(\boldsymbol u)\,\tau_j$ and therefore concerns the contraction readout: it is not a guarantee for the MLP-decoder variant used in Problems~4--6. Universality for that variant follows instead from the standard argument for branch--trunk architectures with a nonlinear decoder~\cite{lu2021learning,seidman2022nomad}, which we cite rather than reprove. Assumption~3 should likewise be read as an assumption about the simulation problems: the ocean comparison in Section~\ref{sec:supp_ocean_results} is direct evidence that a fixed low-rank basis is \emph{not} adequate for Problem~6, which is a result about that problem rather than a defect of the theory. The permutation invariance of the multiplicative fusion (Supplementary Note~1) is unaffected and holds for both variants, since it is a property of the fusion rather than of the readout.

This design was adopted for the measured-data problems because their targets are not smooth solution fields of a single governing system evaluated on a fixed mesh, and a fixed low-rank basis proved too restrictive; the ocean comparison in Section~\ref{sec:supp_ocean_results}, where the nonlinear decoder improves on the linear readout by $12$--$18\%$ relative at comparable parameter count, is the direct measurement of that.

\subsection{Architectures}

\begin{table}[h!]
\centering
\caption{\textbf{MIMONet configurations for all six problems.} Layer widths are listed as input followed by hidden and output widths. ``Contraction'' denotes the linear branch--trunk readout, ``MLP decoder'' the concatenate-and-decode variant of Section~\ref{sec:supp_readout_variants}. Fourier entries give the number of random frequencies and the sampling scale of the trunk feature map; ``---'' means the raw coordinate is used.}
\label{tab:supp_architectures}
\adjustbox{max width=\textwidth}{%
\begin{tabular}{@{}llllcccr@{}}
\toprule
Problem & Branch 1 & Branch 2 & Trunk & Fourier & $I$ & Readout & Params \\
\midrule
1. Lid-driven cavity & $[90,512,512,512,256]$ & --- & $[2,256,256,256,256{\times}3]$ & --- & $256$ & Contraction & $1{,}032{,}963$ \\
2. PWR subchannel & $[100,512,512,512,256]$ & $[2,512,512,512,256]$ & $[2,256,256,256,256{\times}3]$ & --- & $256$ & Contraction & $1{,}696{,}259$ \\
3. Heat exchanger & $[100,512,512,512,256]$ & $[2,512,512,512,256]$ & $[2,256,256,256,256{\times}4]$ & --- & $256$ & Contraction & $1{,}762{,}052$ \\
\midrule
4. PEM fuel cell & $[3n_{\text{obs}},256^{\times4}]$ & $[3n_{\text{obs}},256^{\times4}]$ & $[128,256,256,256]$ & $64$, $\sigma{=}3$ & $256$ & MLP decoder & $872{,}450$ \\
5. IEA-Wind towers & $[3,128^{\times4}]$ & $[4,128^{\times4}]$ & $[64,128,128,128]$ & $32$, $\sigma{=}3$ & $128$ & MLP decoder & $207{,}617$ \\
6. GLORYS12 ocean & $[1200,384^{\times4}]$ & $[1200,384^{\times4}]$ & $[256,384,384,384]$ & $128$, $\sigma{=}3$ & $384$ & MLP decoder & $2{,}795{,}907$ \\
\bottomrule
\end{tabular}}
\end{table}

For the measured-data problems the decoder widths are $[2I, I, I, I, m]$, i.e.\ four linear layers taking the concatenated trunk-and-latent vector of width $2I$ to the $m$ output channels, with GELU between layers. Problem~4 has $m=2$ ($J$, $T$), Problem~5 has $m=1$ (wind speed, evaluated at each query height), Problem~6 has $m=3$ ($T$, $S$, SSH). The branch input dimension for Problem~4 is $3n_{\text{obs}}$ because each observed segment contributes its value and its two normalised coordinates; at the $n_{\text{obs}}=32$ operating point this is $96$, and the parameter count grows by $1{,}536$ for each additional sensor. For Problem~6 it is $1{,}200 = 400\times3$ per branch. The trunk of Problem~5 takes the query \emph{height} alone, so its Fourier map acts on a one-dimensional coordinate; Problem~4 draws its basis over a three-column coordinate whose third entry is constant, so the encoding is effectively two-dimensional, and Problem~6 uses a two-dimensional latitude--longitude basis at the surface level. In all three the Fourier matrix is drawn from $\mathcal N(0,\sigma^2)$ with a generator seeded at $\text{seed}+7$, so the trunk basis itself changes with the run seed.

\subsection{Training configurations}

\begin{table}[h!]
\centering
\caption{\textbf{Training hyperparameters for all six problems.} ``Best-val'' denotes that the full epoch budget is always run and the parameters restored from the epoch with the lowest validation loss; where a validation interval is given, the validation loss is evaluated only every that many epochs plus the final epoch. Batch sizes are in samples for Problems~1--4 and~6 and in profiles for Problem~5; every batch evaluates all query coordinates unless a subsample is stated.}
\label{tab:supp_training}
\adjustbox{max width=\textwidth}{%
\begin{tabular}{@{}llllllll@{}}
\toprule
Problem & Optimiser & LR & Weight decay & Schedule & Epochs & Batch & Stopping \\
\midrule
1. Lid-driven cavity & Adam & $10^{-3}$ & $10^{-7}$ & $\times0.5$ on plateau & $\le 500$ & $8$ & Best-val checkpoint \\
2. PWR subchannel & Adam & $10^{-3}$ & $10^{-6}$ & $\times0.5$ on plateau & $\le 500$ & $4$ & Best-val checkpoint \\
3. Heat exchanger & Adam & $10^{-3}$ & $10^{-6}$ & $\times0.5$ on plateau & $\le 500$ & $4$ & Best-val checkpoint \\
\midrule
4. PEM fuel cell & Adam & $10^{-3}$ & $10^{-5}$ & Cosine to $10^{-6}$ & $200$ & $64$ & Best-val, interval $3$ \\
5. IEA-Wind towers & Adam & $10^{-3}$ & $10^{-5}$ & Cosine to $10^{-6}$ & $60$ & $4{,}096$ & Best-val, interval $3$ \\
6. GLORYS12 ocean & Adam & $10^{-3}$ & $10^{-5}$ & Cosine to $10^{-6}$ & $300$ & $16$ & Best-val, interval $5$ \\
\bottomrule
\end{tabular}}
\end{table}

No problem uses early stopping in the usual sense. On every case the full epoch budget is consumed and the best-validation checkpoint is restored at the end; the patience of $10$ reported in the Schedule column belongs to \texttt{ReduceLROnPlateau} and not to a stopping rule. Problem~6 subsamples $3{,}000$ of the $9{,}457$ query cells per optimisation step, drawn uniformly with replacement and shared across the batch; Problems~4 and~5 use all query coordinates in every step. The loss is a mean-squared error on standardised targets in every case, masked to the finite entries where a target can be missing (Problems~4 and~5).

\paragraph{Seeds and repeated runs.} The measured-data problems are each run at five seeds, $0$ through $4$, and the tables in Section~\ref{sec:supp_realworld_results} report the mean and standard deviation across those runs. A seed sets NumPy and PyTorch generators and, through them, the weight initialisation, the batch shuffling, the Fourier basis (via $\text{seed}+7$) and the sensor draw; for Problems~4 and~6 it also sets the data split, so the five runs differ in split as well as in initialisation. Problem~5 has a fixed chronological split, so its five runs differ only in initialisation, shuffling and Fourier basis, which is why its seed-to-seed spread is the smallest of the three. Bit-exact reproducibility on GPU is not enforced: none of the scripts calls \texttt{torch.use\_deterministic\_algorithms} or fixes the cuDNN backend, and Problem~4 does not seed \texttt{torch.cuda} explicitly, so re-running may shift the last reported digit.

\subsection{Normalisation, coordinate scaling and Fourier features}
\label{sec:supp_normalisation}

Every statistic used for normalisation is computed on training data alone and then applied unchanged to the validation and test partitions; no test sample contributes to any mean, standard deviation or extremum used by the pipeline. The schemes differ across problems because the quantities differ, and the differences are consequential enough to state in one place.

\paragraph{Inputs.} The simulation problems (1--3) standardise the functional and scalar branch inputs with training-set means and standard deviations. Problem~4 standardises each measured channel with means and standard deviations taken over training frames only. Problem~5 standardises wind speed with a \emph{single global} mean and standard deviation pooled over all heights and all training rows rather than per height, so that the model is not handed the height-conditional statistics it is asked to infer. Problem~6 standardises each of $T$, $S$ and SSH per channel with statistics computed on training days and pooled over all cells.

\paragraph{Outputs.} The simulation problems map each output channel independently to $[-1,1]$ using training-set extrema. The measured-data problems predict in the standardised space of the corresponding input channel and invert the transform before any error is computed, so all reported errors are in physical units.

\paragraph{Query coordinates.} Trunk coordinates are scaled to $[-1,1]$ per axis. No physical cell pitch enters the pipeline for Problems~4 and~6, so trunk coordinates and the distances used by the classical interpolants are in scaled units, and for Problem~6 the interpolant distances are scaled latitude--longitude rather than great-circle distance. Problem~5 uses a standardised $\log_{10} z$ height coordinate, which makes the trunk input linear in the variable that classical shear profiles are linear in.

\paragraph{Fourier features.} The measured-data problems encode the trunk coordinate with random Fourier features before the trunk MLP. A basis $B\in\mathbb R^{M\times d}$ is drawn once with $B_{ij}\sim\mathcal N(0,\sigma^2)$ and held fixed for the run, and the coordinate $\mathbf r$ is mapped to
\[
\gamma(\mathbf r) \;=\; \bigl[\sin(2\pi B\mathbf r);\, \cos(2\pi B\mathbf r)\bigr] \in \mathbb R^{2M},
\]
so the trunk input width is $2M$. The pairs $(M,\sigma)$ are the ``Fourier'' column of Table~\ref{tab:supp_architectures}: $(64,3)$ for Problem~4, $(32,3)$ for Problem~5 and $(128,3)$ for Problem~6. The basis is drawn from the run's seed, so it varies across the five seeds along with the sensor draw and the initialisation, and its contribution is therefore included in the reported seed spread rather than held fixed. The encoding is used because a plain coordinate MLP is slow to represent the high-frequency spatial structure these targets carry; the simulation problems, whose targets are smooth fields on a fixed mesh, do not use it.

\subsection{Baseline definitions}
\label{sec:supp_baseline_defs}

The classical baselines are stated here exactly as implemented, because several of them carry names that are used loosely in the literature.

\paragraph{Inverse-distance weighting (IDW).} Weights $w_k \propto \|\mathbf r - \mathbf r_k\|^{-2}$ over \emph{all} observed sensors, normalised to sum to one; there is no neighbour cut-off, so it is global IDW with power $p=2$. Distances are Euclidean in the coordinate system the problem uses: grid-index units for Problem~4, and normalised $[-1,1]$ latitude--longitude for Problem~6, not great-circle distance.

\paragraph{$k$-nearest neighbours.} For Problem~4, $k=8$ with weights $\propto \|\cdot\|^{-1}$, i.e.\ inverse distance to the \emph{first} power. This differs from the IDW baseline in both the neighbour set and the exponent, so the two are distinct interpolants rather than the same interpolant with and without a cut-off. For Problem~6, $k=10$ with scikit-learn's distance weighting, refitted per test day.

\paragraph{Power law (Problem~5).} A two-point exponent computed per profile from the $10\,$m and $50\,$m sensors, $\alpha=\log(U_{50}/U_{10})/\log(50/10)$, extrapolated as $U(z)=U_{50}(z/50)^{\alpha}$. Nothing is fitted on training data.

\paragraph{Two-point log-height extrapolation (Problem~5).} A straight line in $U$ against $\log_{10} z$ through the $25\,$m and $50\,$m sensors, extrapolated to the query height. We do not call this the logarithmic wind profile: it contains no roughness length, no von K\'arm\'an constant, no friction velocity and no stability correction, and it is not derived from surface-layer similarity theory. It is included as the natural two-point extrapolant in log-height coordinates, which is what a practitioner without a stability estimate would use.

\paragraph{Nearest sensor (Problem~5).} Returns the $50\,$m reading unchanged at every query height.

\paragraph{Climatology (Problem~6).} The per-cell mean over the training days, broadcast to every test day. It is a training-period mean, not a seasonal or day-of-year climatology.

\paragraph{Ridge regression (Problem~6).} A global two-feature linear map from the IDW-interpolated temperature and salinity at a cell to the sea-surface height at that cell, $\mathrm{SSH} \approx a\hat T + b\hat S + c$, fitted on all training (day, cell) pairs with $\alpha=1$. The regularisation strength was not selected by cross-validation; it is the library default and was left at that value.

\paragraph{KCN and the neural operators.} KCN embeds each of a query point's $k$ nearest sensors as a feature vector, scores them, applies a softmax over the neighbours and decodes the attention-weighted sum; $k=12$ for Problems~4 and~6 and $k=3$ for Problem~5, where the three input heights are the only sensors available. FNO is applied where the input has a natural grid: Problem~5, where the profile is an ordered one-dimensional sequence, and Problem~6, where the scattered sensors are first IDW-rasterised onto the $60\times160$ grid, which makes it a two-stage pipeline whose front end is itself one of the classical baselines. FNO is not applied to Problem~4, whose $32$ scattered segments would require the same rasterisation step on an $18\times18$ grid.

\section{Baseline models}

Two families of comparator appear in this work and they answer different questions. The classical
estimators defined in Section~\ref{sec:supp_baseline_defs}, inverse-distance weighting, $k$-nearest
neighbours, the analytic wind profiles and climatology, are what is actually deployed when an
interior field is wanted and no governing model is available at the point of use. They are the
comparison the paper's claim rests on, and they apply only to Problems~4--6, where the inputs are
samples of the field being reconstructed.

The operator baselines defined in this section, NOMAD, KCN and Geo-FNO in their various
combinations, answer a narrower question: whether the reported accuracy is a property of the
operator formulation or an artefact of one architecture. They are configured here in full so that
the reader can judge whether each was given a fair budget, and their per-problem results appear in
Sections~\ref{sec:supp_sim_results} and~\ref{sec:supp_realworld_results}.
\subsection{NOMAD (nonlinear manifold decoders)}
NOMAD \cite{seidman2022nomad} approximates the operator $\mathcal{G}:\mathcal{U}\to\mathcal{S}\subset L^{2}(\mathcal{Y};\mathbb{R}^{m})$ with an encoder--approximator--decoder architecture whose decoder is nonlinear in latent coordinates. Given a mixed-modal input $\boldsymbol{u}\in\mathcal{U}$, an encoder $E:\mathcal{U}\to\mathbb{R}^{p}$ produces a feature vector $z=E(\boldsymbol{u})$, an approximator $A:\mathbb{R}^{p}\to\mathbb{R}^{n_{\text{NOMAD}}}$ maps $z$ to latent coordinates $\zeta=A(z)$, and a decoder $D:\mathbb{R}^{n_{\text{NOMAD}}}\times\mathcal{Y}\to\mathbb{R}^{m}$ returns the $m$-channel field value at query locations. (We use $n_{\text{NOMAD}}$ for NOMAD's latent dimension to avoid collision with $n$, which denotes the number of input modalities in the MIMONet section, and with $n_x$, which denotes the state-space dimension of classical observers.) The pointwise prediction is therefore
\[
\hat{\mathbf{s}}(\mathbf{r}\,|\,\boldsymbol{u}) \;=\; D\!\left(\zeta,\mathbf{r}\right)\in\mathbb{R}^{m},
\qquad \zeta\in\mathbb{R}^{n_{\text{NOMAD}}},\ \mathbf{r}\in\mathcal{Y},
\]
where $m$ is the number of output channels, $n_{\text{NOMAD}}$ the latent dimension, and $p$ the encoder feature width; the per-case widths are listed in Table~\ref{tab:nomad_arch}. The latent coordinates $\zeta\in\mathbb{R}^{n_{\text{NOMAD}}}$ are shared across all $m$ output channels and the decoder $D(\zeta,\mathbf{r})$ returns the channels jointly at each query point; in this shared-latent sense NOMAD belongs to the same multi-field-decoding regime as MIMONet. For a batch of $P$ query points $\{\mathbf{r}_p\}_{p=1}^{P}$, independent evaluations of $D(\zeta,\mathbf{r}_p)$ yield $\hat{\mathbf{s}}\in\mathbb{R}^{P\times m}$. Unlike linear-basis decoders that confine $\hat{\mathbf{s}}(\cdot)$ to a finite-dimensional linear subspace, the nonlinear map $D(\zeta,\mathbf{r})$ permits curved (nonlinear) output manifolds, which is advantageous for solution families with slowly decaying spectra or transport-dominated structure. The distinguishing architectural choice between NOMAD and MIMONet is the branch-fusion operator that produces the shared latent: NOMAD's encoder $E$ aggregates multi-modal inputs additively, whereas MIMONet's branch fusion is multiplicative. Training uses standard pointwise losses over $\mathbf{r}\in\mathcal{Y}$, preserving query-based evaluation consistent with the notation above.

\begin{table}[htbp]
\centering
\caption{\textbf{NOMAD architectures used in this study.} Layers are listed as input and subsequent hidden/output widths per MLP block. Branch features are broadcast across query points before concatenation with trunk features.}
\label{tab:nomad_arch}
\begin{adjustbox}{width=\textwidth}
\begin{tabular}{@{}p{2.8cm}p{4.1cm}p{3.6cm}p{4.6cm}p{2.2cm}@{}}
\toprule
\textbf{Problem} & \textbf{Branch encoder(s)} & \textbf{Trunk} & \textbf{Combined head} & \textbf{Outputs} \\
\midrule
LDC & $[90,512,512,512,512,512]$ & $[2,512,512,512]$ & $[1024,512,512,512,512,3]$ & $\{p,\,\|\boldsymbol{v}\|,\,k\}$ \\
\addlinespace[3pt]
Subchannel & $[100,128,128]$, $[2,128,128]$ & $[2,128,128]$ & $[384,128,3]$ & $\{T,\,k,\,\|\boldsymbol{v}\|\}$ \\
\addlinespace[3pt]
Heat exchanger & $[100,128,128]$,$[2,128,128]$ & $[2,128,128]$ & $[384,128,4]$ & $\{p,\,u_x,\,u_y,\,u_z\}$ \\
\bottomrule
\end{tabular}
\end{adjustbox}
\end{table}

\subsection{KCN (Kriging Convolutional Networks)\,{+}\,NOMAD}
In our comparisons, we also evaluated a hybrid in which NOMAD's latent representation is decoded by a KCN-style local graph \cite{appleby2020kriging}, rather than an MLP.  In this hybrid, a NOMAD backbone first maps $\boldsymbol{u}\!\in\!\mathcal{U}$ to latent coordinates $\zeta\!\in\!\mathbb{R}^{n_{\text{NOMAD}}}$. For each query location $\mathbf{r}\!\in\!\mathcal{Y}$, a $K$-nearest-neighbour graph is formed over the query coordinates with edge weights given by an RBF kernel of bandwidth $\phi$,
\[
A_{jk}=\exp\!\Big(-\tfrac{\|\mathbf{r}_j-\mathbf{r}_k\|_2^2}{2\phi^2}\Big).
\]
Node features concatenate coordinates and the latent, e.g.\ centre $[\mathbf{r},\zeta]$ and neighbours $[\mathbf{x}_k,\zeta]$. The per-case configurations are given in Table~\ref{tab:kcn_nomad_configs}. A shallow message-passing GCN (three \texttt{GCNConv} layers, width $32$, ReLU, dropout as specified) aggregates these features and a final linear layer maps embeddings to $m$ output channels:
\[
\hat{\mathbf{s}}(\mathbf{r}\,|\,\boldsymbol{u}) \;=\; \Phi\!\big(\,[\mathbf{r},\zeta],\,\{[\mathbf{x}_k,\zeta]\}_{k=1}^{K},\,A\,\big)\in\mathbb{R}^{m}.
\]
The hybrid is trained end-to-end with the same pointwise loss on $\mathcal{Y}$ as other baselines. At inference, only coordinates and the NOMAD features are used; no labels enter the decoder. Across all problems (LDC, subchannel, heat exchanger), the KCN decoder used a fixed $K=5$ nearest neighbours. The RBF bandwidth $\phi$ was set automatically to the median of the $K$-NN distances computed on the training coordinates; the edge weights were $A_{jk}=\exp(-\|\mathbf{r}_j-\mathbf{r}_k\|_2^2/(2\phi^2))$ with $\gamma=1/(2\phi^2)$ in the implementation.

\begin{table}[htbp]
\centering
\caption{\textbf{KCN{+}NOMAD configurations.} ``Branch'' lists widths for each parallel branch MLP; ``Trunk'' lists query-MLP widths; ``Head'' lists the latent-fusion MLP prior to the KCN decoder. The KCN block (3\,\texttt{GCNConv} layers, width 32) is shared across problems.}
\label{tab:kcn_nomad_configs}
\begin{adjustbox}{width=\textwidth}
\begin{tabular}{@{}lccccc@{}}
\toprule
\textbf{Problem} 
& \textbf{NOMAD Branch} 
& \textbf{NOMAD Trunk} 
& \textbf{NOMAD Head} 
& \textbf{KCN GNN} 
& \textbf{Outputs} \\
\midrule
LDC
& $[90,32,32]$
& $[2,32]$
& $[64,32,32]$
& 3\,$\times$\,\texttt{GCNConv} (32)
& $\{p,\,\|\boldsymbol{v}\|,\,k\}$ \\
Subchannel
& $[100,32,32]$, $[2,32,32]$
& $[2,32]$
& $[96,32,32]$
& 3\,$\times$\,\texttt{GCNConv} (32)
& $\{T,\,k,\,\|\boldsymbol{v}\|\}$ \\
Heat exchanger
& $[100,32,32]$, $[2,32,32]$
& $[2,32]$
& $[96,32,32]$
& 3\,$\times$\,\texttt{GCNConv} (32)
& $\{p,\,u_x,\,u_y,\,u_z\}$ \\
\bottomrule
\end{tabular}
\end{adjustbox}
\end{table}

\subsection{GeoFNO\,{+}\,NOMAD}
A hybrid spectral--manifold decoder was evaluated in which a NOMAD front end produces a coordinate-conditioned latent map that is further refined by a Geo-FNO \cite{li2023fourier}; the per-case configurations are listed in Table~\ref{tab:geofno_nomad_arch}. Concretely, a NOMAD block maps inputs to an intermediate two-channel feature at query locations; these features are then lifted and passed through a Geo-FNO with four spectral convolution layers (2D FFT/linear/IFFT with learned low-rank weights) interleaved with $1{\times}1$ pointwise convolutions and bias terms. An inverse mapping network (\texttt{IPHI}) provides coordinate warping for nonuniform query sets. The final linear head projects to $m$ output channels. Models were trained end-to-end with the same pointwise loss on $\mathcal{Y}$ as other baselines.

\begin{table}[htbp]
\centering
\caption{\textbf{GeoFNO{+}NOMAD configurations per problem.} ``Branch/Trunk/Head'' list widths per MLP block in the NOMAD front end. The GeoFNO block uses 4 spectral layers and pointwise $1{\times}1$ convolutions with width 32; \texttt{IPHI} is used for coordinate warping.}
\label{tab:geofno_nomad_arch}
\begin{adjustbox}{width=\textwidth}
\begin{tabular}{@{}lccccc@{}}
\toprule
\textbf{Problem}
& \textbf{NOMAD Branch}
& \textbf{NOMAD Trunk}
& \textbf{NOMAD Head}
& \textbf{GeoFNO core}
& \textbf{Outputs} \\
\midrule
LDC
& $[90,32,32]$
& $[2,32]$
& $[64,32,2]$
& modes$(16,16)$, width $32$
& $\{p,\,\|\boldsymbol{v}\|,\,k\}$ \\
\addlinespace[3pt]
Subchannel
& $[100,32,32]$, $[2,32,32]$
& $[2,32]$
& $[96,32,2]$
& modes$(16,16)$, width $32$
& $\{T,\,k,\,\|\boldsymbol{v}\|\}$ \\
\addlinespace[3pt]
Heat exchanger
& $[100,32,32]$, $[2,32,32]$
& $[2,32]$
& $[96,32,2]$
& modes$(16,16)$, width $32$
& $\{p,\,u_x,\,u_y,\,u_z\}$ \\
\bottomrule
\end{tabular}
\end{adjustbox}
\end{table}

\noindent Notes: (i) The GeoFNO lift/projection uses \texttt{fc0:} $2{\to}32$, \texttt{fc1:} $32{\to}128$, \texttt{fc2:} $128{\to}m$ with GELU activations; $m$ equals the number of outputs per problem (3 for LDC/subchannel, 4 for heat exchanger). (ii) The NOMAD head outputs a two-channel feature per query, which serves as the GeoFNO input (\texttt{in\_channels}=2). (iii) All runs used identical spectral mode counts $(16,16)$ and width $32$; IPHI was enabled in all cases.

\subsection{FNO (Problems~5 and~6)}
\label{sec:supp_fno_arch}

FNO is reported as a baseline on the two problems whose inputs carry a natural grid, and its configuration is given here for completeness; no FNO is run on Problems~1--4.

Table~\ref{tab:supp_fno_arch} gives both configurations. For Problem~5 the vertical profile is an ordered one-dimensional sequence, so a one-dimensional FNO is used directly on the three input heights. For Problem~6 the scattered sensors carry no grid, so the sensor values are first rasterised onto the $60\times160$ ocean grid by inverse-distance weighting and a two-dimensional FNO is applied to the result; the rasterisation stage is part of the baseline and is the reason this model is not simply an operator on the same inputs MIMONet receives.

\begin{table}[h!]
\centering
\caption{\textbf{FNO baseline configurations.} Both use the standard lift--spectral--project structure with pointwise $1\times1$ convolutions in the residual path and GELU activations. ``Modes'' is the number of retained Fourier modes per dimension.}
\label{tab:supp_fno_arch}
\adjustbox{max width=\textwidth}{%
\begin{tabular}{@{}lcccccr@{}}
\toprule
Problem & Variant & Lift width & Modes & Depth & Projection & Params \\
\midrule
5. IEA-Wind towers & FNO1d & $48$ & $3$ & $4$ & $48\!\to\!64\!\to\!1$ & $68{,}289$ \\
6. GLORYS12 ocean & FNO2d & $26$ & $16$ & $4$ & $26\!\to\!128\!\to\!3$ & $2{,}775{,}703$ \\
\bottomrule
\end{tabular}}
\end{table}

On Problem~6 the parameter count is within $0.8\%$ of MIMONet-NL's $2{,}795{,}907$, so that comparison is capacity-matched. On Problem~5 it is not, as recorded in Section~\ref{sec:supp_wind_results}.

\section{Complete results for the simulation problems}
\label{sec:supp_sim_results}

The main text quotes the per-channel errors for Problems~1--3 in prose. They are collected in Table~\ref{tab:supp_sim_results} against the three operator baselines so that the comparison can be read in one place, and because the simulation problems are the ones for which no classical interpolant exists to compare against: their inputs are boundary data rather than samples of the output field, so the only available comparators are other operator formulations. These results establish that the framework is accurate on coupled multiphysics fields; the argument against what is deployed in practice is carried by Problems~4--6 in Section~\ref{sec:supp_realworld_results}.

\begin{table}[h!]
\centering
\caption{\textbf{Reconstruction error on the three simulation problems.} Mean relative $\ell_2$ error (\%) per output channel on the held-out test set of each case, with the channel mean in the final column. Parameter counts are given because the baselines span two orders of magnitude in capacity and the comparison is therefore not capacity-controlled; KCN in particular is far smaller than the others by construction, as its neighbour graph carries most of its structure. Every entry was regenerated from the released checkpoints and agrees with the value quoted in the main text to within $0.1$ percentage points; the harness is \texttt{Production/reproduce\_accuracy.py}.}
\label{tab:supp_sim_results}
\adjustbox{max width=\textwidth}{%
\begin{tabular}{@{}llccccc@{}}
\toprule
Model & Params & \multicolumn{4}{c}{Per-channel error (\%)} & Mean (\%) \\
\midrule
\multicolumn{7}{@{}l}{\emph{(a) Lid-driven cavity}} \\
 & & $p$ & $\|\boldsymbol v\|$ & $k$ & & \\
\cmidrule(lr){3-5}
\textbf{MIMONet} & $1.0$\,M & $\mathbf{2.1}$ & $\mathbf{5.0}$ & $\mathbf{2.8}$ & & $\mathbf{3.3}$ \\
NOMAD & $2.9$\,M & $23.7$ & $46.9$ & $55.1$ & & $41.9$ \\
GeoFNO & $2.7$\,M & $49.8$ & $82.4$ & $71.7$ & & $68.0$ \\
KCN & $10.4$\,K & $27.6$ & $6.0$ & $41.5$ & & $25.0$ \\
\addlinespace[0.5em]
\multicolumn{7}{@{}l}{\emph{(b) PWR subchannel}} \\
 & & $\|\boldsymbol v\|$ & $T$ & $k$ & & \\
\cmidrule(lr){3-5}
\textbf{MIMONet} & $1.7$\,M & $\mathbf{2.2}$ & $\mathbf{0.27}$ & $\mathbf{4.2}$ & & $\mathbf{2.2}$ \\
NOMAD & $113$\,K & $74.1$ & $98.5$ & $30.9$ & & $67.8$ \\
GeoFNO & $2.7$\,M & $70.2$ & $98.6$ & $47.7$ & & $72.2$ \\
KCN & $12.9$\,K & $91.5$ & $8.3$ & $59.2$ & & $53.0$ \\
\addlinespace[0.5em]
\multicolumn{7}{@{}l}{\emph{(c) Heat exchanger}} \\
 & & $p$ & $u_z$ & $u_y$ & $u_x$ & \\
\cmidrule(lr){3-6}
\textbf{MIMONet} & $1.7$\,M & $\mathbf{0.82}$ & $\mathbf{1.45}$ & $\mathbf{1.02}$ & $\mathbf{0.52}$ & $\mathbf{0.95}$ \\
NOMAD & $113$\,K & $11.5$ & $26.3$ & $17.5$ & $10.1$ & $16.4$ \\
GeoFNO & $2.7$\,M & $3.90$ & $8.70$ & $6.00$ & $3.40$ & $5.50$ \\
KCN & $12.9$\,K & $14.4$ & $22.2$ & $17.1$ & $11.4$ & $16.3$ \\
\bottomrule
\end{tabular}}
\end{table}

Two features are worth drawing out. The ordering among the baselines is not stable across the three problems: GeoFNO is the weakest on the lid-driven cavity and the subchannel but the strongest on the heat exchanger, and KCN is competitive on the single smooth channel of each case ($\|\boldsymbol v\|$ at $6.0\%$ on the cavity, $T$ at $8.3\%$ on the subchannel) while failing on the others. This is the behaviour a locality-based estimator should show, and it is why the paper does not present a ranking among operator families. Second, the capacity spread is large and uncontrolled: KCN carries between $10$ and $13$ thousand parameters, against $113$ thousand to $2.9$ million for the others, so its errors should be read as a statement about the locality of its construction rather than as a like-for-like comparison.

\section{Complete real-world results}
\label{sec:supp_realworld_results}

The main text quotes single operating points from each of the three measured-data problems. The complete five-seed results are given here. Throughout, the quoted spread is the standard deviation over the five seeds computed with the population convention (dividing by $n$), which is the convention used in the main text; the classical baselines are deterministic given the sensor draw and the split, but both of those are seed-dependent for Problems~4 and~6, so those baselines carry a seed spread and it is reported; where a baseline uses no sensors and the split is fixed it is reported without a spread.

\subsection{Fuel-cell reconstruction across sensor coverage}
\label{sec:supp_pemfc_results}

\begin{figure}[h!]
    \centering
    \includegraphics[width=\textwidth]{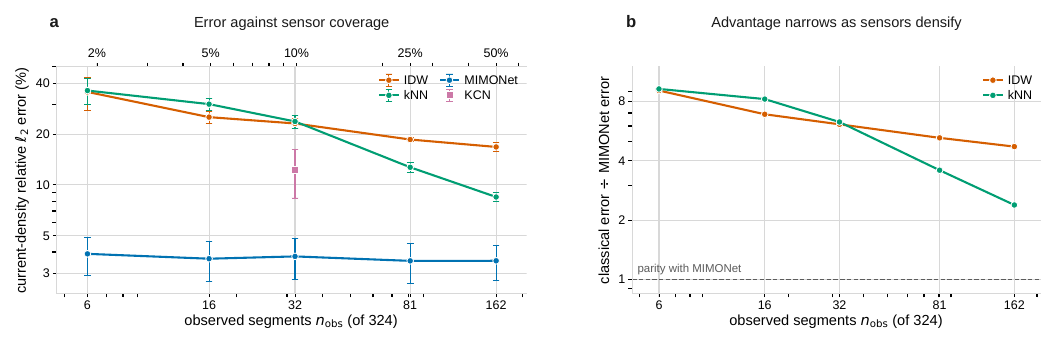}
    \caption{\textbf{Fuel-cell reconstruction against sensor coverage.} Current-density relative $\ell_2$ error on the held-out segments, mean over five seeds with error bars at $\pm1$ standard deviation. The upper axis of \textbf{a} gives coverage as a fraction of the $324$ segments of the $18\times18$ mapping grid. Only the current-density channel is shown; the temperature channel is a bilinear upsampling of an $81$-point measurement (Section~\ref{sec:supp_data_pemfc}) and does not discriminate between methods. \textbf{a}, Absolute error. MIMONet is nearly flat across a $27$-fold change in sensor count, whereas the classical interpolants improve steeply; KCN was run at the $32$-sensor operating point only. \textbf{b}, The same data as the ratio of classical error to MIMONet error. The operator's advantage is largest where sensors are sparsest, falling from roughly $9\times$ at six sensors to $2.4\times$ for $k$NN at $162$, so the sparse regime that motivates virtual sensing is also the regime in which the advantage is real.}
    \label{fig:supp_pemfc_coverage}
\end{figure}

Table~\ref{tab:supp_pemfc_results} reports every coverage level of the sweep, and Fig.~\ref{fig:supp_pemfc_coverage} plots it. The current-density channel $J$ is the informative one: it is measured at all $324$ segments, whereas the temperature channel is a bilinear upsampling of $81$ measurements (Section~\ref{sec:supp_data_pemfc}) and is correspondingly easy for every method, so the temperature column is reported for completeness rather than as a discriminating comparison.

\begin{table}[h!]
\centering
\caption{\textbf{PEM fuel-cell reconstruction across sensor coverage.} Relative $\ell_2$ error (\%) on the held-out segments, mean $\pm$ s.d.\ over five seeds. Coverage is $n_{\text{obs}}$ out of the $324$ segments of the $18\times18$ current-density grid. KCN was run at the $32$-sensor operating point only.}
\label{tab:supp_pemfc_results}
\begin{tabular}{@{}rlllr@{}}
\toprule
$n_{\text{obs}}$ & Coverage & Method & $J$ (\%) & $T$ (\%) \\
\midrule
\multirow{3}{*}{$6$} & \multirow{3}{*}{$1.9\%$} & MIMONet & $\mathbf{3.91 \pm 0.98}$ & $\mathbf{0.044 \pm 0.002}$ \\
 & & IDW & $35.51 \pm 7.97$ & $0.335 \pm 0.038$ \\
 & & $k$NN & $36.18 \pm 6.26$ & $0.356 \pm 0.027$ \\
\midrule
\multirow{3}{*}{$16$} & \multirow{3}{*}{$4.9\%$} & MIMONet & $\mathbf{3.66 \pm 0.99}$ & $\mathbf{0.039 \pm 0.004}$ \\
 & & IDW & $25.18 \pm 2.04$ & $0.269 \pm 0.012$ \\
 & & $k$NN & $30.07 \pm 2.35$ & $0.305 \pm 0.015$ \\
\midrule
\multirow{4}{*}{$32$} & \multirow{4}{*}{$9.9\%$} & MIMONet & $\mathbf{3.77 \pm 1.02}$ & $\mathbf{0.038 \pm 0.004}$ \\
 & & KCN & $12.29 \pm 3.93$ & $0.055 \pm 0.003$ \\
 & & IDW & $23.10 \pm 1.60$ & $0.241 \pm 0.018$ \\
 & & $k$NN & $23.74 \pm 2.04$ & $0.239 \pm 0.021$ \\
\midrule
\multirow{3}{*}{$81$} & \multirow{3}{*}{$25.0\%$} & MIMONet & $\mathbf{3.55 \pm 0.92}$ & $\mathbf{0.035 \pm 0.001}$ \\
 & & IDW & $18.56 \pm 0.57$ & $0.190 \pm 0.013$ \\
 & & $k$NN & $12.71 \pm 0.82$ & $0.131 \pm 0.011$ \\
\midrule
\multirow{3}{*}{$162$} & \multirow{3}{*}{$50.0\%$} & MIMONet & $\mathbf{3.55 \pm 0.85}$ & $\mathbf{0.034 \pm 0.002}$ \\
 & & IDW & $16.77 \pm 1.08$ & $0.168 \pm 0.009$ \\
 & & $k$NN & $8.48 \pm 0.49$ & $0.089 \pm 0.006$ \\
\bottomrule
\end{tabular}
\end{table}

The sweep shows two behaviours that the single operating point in the main text does not. MIMONet is nearly flat in coverage: the current-density error moves only from $3.91\%$ at six sensors to $3.55\%$ at $162$, a $9\%$ relative change across a $27$-fold increase in sensor count. The classical interpolants are not: $k$NN improves from $36.18\%$ to $8.48\%$ over the same range. The advantage of the operator is therefore largest where sensors are sparsest, which is the regime the application is about, and it narrows steadily as the field becomes densely observed. At $50\%$ coverage $k$NN is within a factor of $2.4$ of MIMONet on $J$, and simple extrapolation of the trend suggests the two would meet somewhere beyond the coverage this cell provides. Reporting only the $10\%$ point would leave that convergence invisible, which is why the full sweep is given here. The $\pm 1$ standard deviation on the MIMONet current-density errors is around a quarter of the mean at every coverage level, driven by which segments the seed happens to select as sensors.

\begin{figure}[h!]
    \centering
    \includegraphics[width=\textwidth]{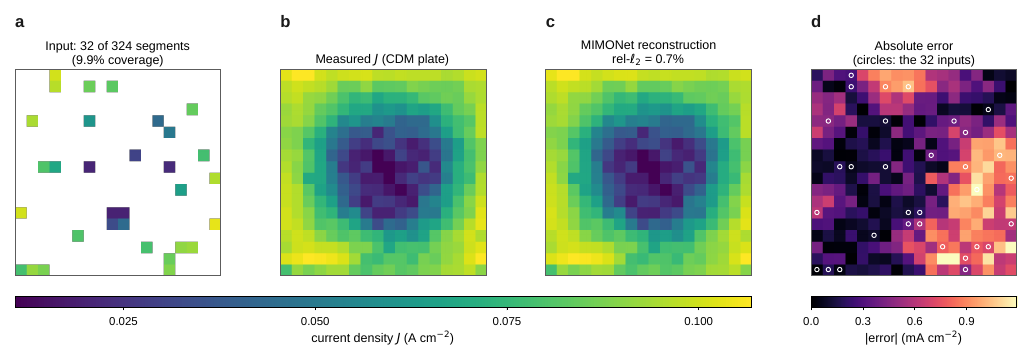}
    \caption{\textbf{Representative fuel-cell reconstruction and the spread behind it.} Seed-0 checkpoint, evaluated on the held-out frames at the $32$-sensor operating point. \textbf{a}, The input actually supplied: $32$ of the $324$ segments, $9.9\%$ coverage. \textbf{b}, The measured current density from the CDM plate on a held-out frame. \textbf{c}, The reconstruction at all $324$ segments. \textbf{d}, Absolute error, with the $32$ input locations circled; the error is not concentrated at the unobserved segments, which is the behaviour a spatial interpolant would show. Panels \textbf{b}--\textbf{d} show the median-error frame. That frame is not representative of the reported mean: the per-frame distribution is strongly right-skewed, with a median of $0.7\%$ against a mean of $3.4\%$ on this seed, so a frame carrying the mean error sits at roughly the $95$th percentile. The headline current-density figure is set by a minority of difficult frames rather than by typical performance, and both statistics are given here so that neither reading stands alone.}
    \label{fig:supp_pemfc_recon}
\end{figure}

A representative reconstruction is shown in Fig.~\ref{fig:supp_pemfc_recon}, together with the distribution of per-frame errors behind it. The distribution matters for reading the headline number. Errors are concentrated well below the mean, with a median of $0.7\%$ against a mean of $3.4\%$ on this seed, so the reported current-density error is set by a minority of frames rather than by the typical case. We report the mean throughout for consistency with the rest of the paper and with the main text, but a reader comparing a virtual sensor against an operational tolerance would want the distribution rather than its first moment, and the skew here is large enough that the two lead to different judgements. The spatial pattern of the error is also informative: it is not concentrated at the unobserved segments, which is what a spatial interpolant would produce, but follows the regions of steepest current-density gradient near the channel bends.

\subsection{Wind-profile reconstruction above the instrumented range}
\label{sec:supp_wind_results}

Table~\ref{tab:supp_wind_results} reports all four query heights. The main text quotes the $80\,$m row.

\begin{table}[h!]
\centering
\caption{\textbf{Vertical wind-speed reconstruction at all four query heights.} Relative $\ell_2$ error (\%), mean $\pm$ s.d.\ over five seeds for the learned models; the analytic baselines are deterministic and carry no seed spread. Input sensors are at $10/25/50\,$m in every case. The test set contains $90{,}222$ profiles at $60$, $70$ and $80\,$m but only $34{,}967$ at $100\,$m, because two of the seven towers do not reach that height (Section~\ref{sec:supp_data_wind}); the $100\,$m column is therefore evaluated on a smaller and differently composed subset and is not directly comparable with the other three.}
\label{tab:supp_wind_results}
\begin{tabular}{@{}lcccc@{}}
\toprule
Method & $60\,$m & $70\,$m & $80\,$m (hub) & $100\,$m \\
\midrule
MIMONet & $\mathbf{2.67 \pm 0.15}$ & $\mathbf{3.79 \pm 0.06}$ & $\mathbf{4.71 \pm 0.15}$ & $\mathbf{14.70 \pm 0.59}$ \\
FNO & $4.05 \pm 0.48$ & $5.82 \pm 1.03$ & $7.30 \pm 1.16$ & $18.88 \pm 1.00$ \\
KCN & $11.03 \pm 0.37$ & $19.21 \pm 0.40$ & $25.28 \pm 0.57$ & $43.06 \pm 1.14$ \\
Nearest sensor ($50\,$m) & $11.92$ & $20.95$ & $27.86$ & $47.73$ \\
Two-point log-height extrapolation & $12.43$ & $21.86$ & $28.95$ & $52.49$ \\
Power law & $12.89$ & $23.68$ & $34.30$ & $95.92$ \\
\bottomrule
\end{tabular}
\end{table}

The margin over the analytic profiles widens with distance above the topmost input sensor: at $60\,$m, ten metres above it, the operator is $4.5\times$ better than the nearest-sensor estimate, and at $80\,$m it is $5.9\times$ better. The reason is structural rather than a matter of tuning. The power law and the log-height extrapolation are two-parameter forms fitted to two sensors on each profile, so they can express only a single shape and cannot represent a profile that flattens, accelerates or reverses curvature above the instrumented range; when the atmosphere departs from that shape the error is bounded below by the model class, not by the data. An operator that has learned the distribution of profile shapes at these sites is not constrained in that way. This is the same argument as in the fuel-cell and ocean problems: the gain comes from replacing a fixed functional form with a learned one, not from a better fit within the same form.

FNO reaches comparable quality, which is expected and worth stating plainly, because the vertical profile is an ordered one-dimensional sequence and therefore a natural input for a spectral operator; the two learned models are not capacity-matched: MIMONet carries $207{,}617$ parameters against FNO's $68{,}289$ and KCN's $66{,}818$, so FNO and KCN are matched to each other while MIMONet is roughly three times larger. We state this rather than presenting the comparison as capacity-controlled, and note that it cuts against the operator-versus-operator reading: FNO reaches comparable accuracy at a third of the budget. The comparison the paper rests on is against the classical estimators, where the gap is not attributable to capacity. Their similarity is the point rather than a complication: where an operator formulation applies, it is the formulation and not the particular architecture that produces the advance over the analytic profiles.

\begin{figure}[h!]
    \centering
    \includegraphics[width=\textwidth]{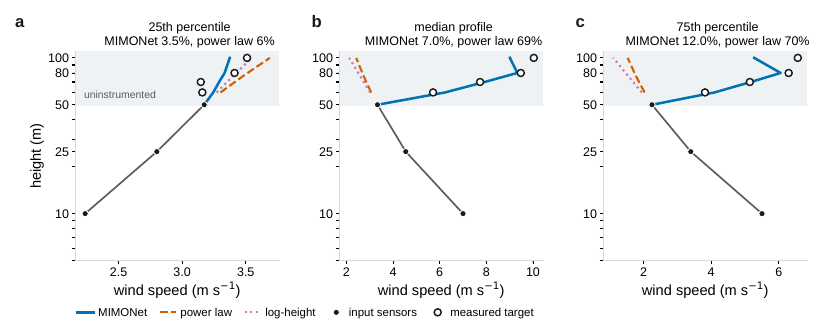}
    \caption{\textbf{Why a fixed-form profile cannot follow the measured one above the mast.} Seed-0 checkpoint on held-out profiles; the shaded band is the uninstrumented region above the topmost input sensor at $50\,$m. \textbf{a}--\textbf{c}, Three held-out profiles, with filled markers the three inputs at $10$, $25$ and $50\,$m, open markers the measured targets, and the grey segment below $50\,$m the measured input rather than a prediction. The three are taken at the $25$th, $50$th and $75$th percentile of \emph{MIMONet's own} error, so the selection does not depend on how the classical extrapolants happen to fare on them. Aggregate errors at every query height are in Table~\ref{tab:supp_wind_results}. In \textbf{b} and \textbf{c} the measured profile decreases with height through the instrumented range and then increases above it. The power law and the two-point log-height extrapolant are two-parameter shapes fitted to two of the three input sensors ($10$ and $50\,$m for the power law, $25$ and $50\,$m for the log-height form), so they can only continue the shear those sensors show and both run the wrong way; the operator turns with the measurement. This is the qualitative content of the error numbers: the failure of the analytic profiles is structural rather than a matter of tuning.}
    \label{fig:supp_wind_profile}
\end{figure}

Figure~\ref{fig:supp_wind_profile} shows what the error numbers mean profile by profile, and it makes the limitation of the analytic baselines concrete rather than statistical. Both the power law and the log-height extrapolant are two-parameter forms fitted to two of the three input sensors, so the shape they extend above the mast is fixed by the shear those sensors show. When the measured profile reverses above $50\,$m, which happens in two of the three profiles drawn here at ordinary quantiles of the operator's error, no choice of the two parameters can follow it and both curves diverge from the measurement in the wrong direction. The operator is not constrained to a shape: it maps the three measured speeds together with site and diurnal context to a query height, so a reversal is representable provided the training data contains such profiles. That is the sense in which this problem is not an incremental improvement in fitting accuracy but a different class of estimator, and it is the same distinction, in a one-dimensional setting, that the ocean problem makes across modalities.

\subsection{Ocean-state reconstruction from disjoint sparse networks}
\label{sec:supp_ocean_results}

Table~\ref{tab:supp_ocean_results} reports the three channels for the operators and classical estimators listed; the KCN and IDW-gridded FNO runs quoted in the main text are tabulated in Table~\ref{tab:supp_ocean_learned}. Sea-surface height carries no sensors of its own in any method's input, so the SSH column measures cross-modal inference throughout.

\begin{table}[h!]
\centering
\caption{\textbf{North Atlantic ocean-state reconstruction.} Relative $\ell_2$ error (\%) per channel on the $60$ held-out days, mean $\pm$ s.d.\ over five seeds, under the random day split used in the main text. Inputs are $400$ temperature and $400$ salinity sensors at disjoint locations; no method receives any SSH observation. Dashes mark combinations that are undefined: IDW and $k$NN interpolate a field from its own sensors and have none for SSH, so the SSH entry for each of those rows is a ridge regression of SSH on the temperature and salinity fields interpolated by that same method, reported on its own line. The two ridge variants differ only in which interpolant supplies their inputs.}
\label{tab:supp_ocean_results}
\begin{tabular}{@{}lrccc@{}}
\toprule
Method & Params & $T$ (\%) & $S$ (\%) & SSH (\%) \\
\midrule
MIMONet-NL & $2{,}795{,}907$ & $\mathbf{3.10 \pm 0.06}$ & $\mathbf{0.46 \pm 0.01}$ & $\mathbf{9.11 \pm 0.17}$ \\
NOMAD & $2{,}943{,}363$ & $3.49 \pm 0.10$ & $0.51 \pm 0.01$ & $10.72 \pm 0.39$ \\
MIMONet (linear decoder) & $2{,}412{,}547$ & $3.54 \pm 0.11$ & $0.53 \pm 0.00$ & $11.05 \pm 0.25$ \\
\midrule
$k$NN ($k=10$) & --- & $6.79 \pm 0.32$ & $0.99 \pm 0.04$ & see below \\
IDW ($p=2$) & --- & $9.26 \pm 0.49$ & $1.35 \pm 0.06$ & see below \\
Ridge on $k$NN-interpolated $T,S$ & --- & --- & --- & $58.18 \pm 0.62$ \\
Ridge on IDW-interpolated $T,S$ & --- & --- & --- & $59.44 \pm 1.02$ \\
Climatology & --- & $17.68 \pm 0.43$ & $1.09 \pm 0.01$ & $37.71 \pm 0.29$ \\
\bottomrule
\end{tabular}
\end{table}

The comparison that carries the argument is the one against the classical estimators, and it separates into two very different situations. On temperature and salinity, which are directly sampled, interpolation is a well-posed problem and the classical methods do it competently; the operator improves on them by roughly a factor of two, which is a useful but incremental gain. Sea-surface height is the case the framework exists for, and it is the concrete instance of the disjoint sensing and reconstruction domains identified in the Introduction: the quantity to be recovered is of a different modality from every quantity observed. A spatial interpolant is defined by samples of the field it estimates, so with no SSH sensor anywhere IDW and $k$NN have no argument to take; this is a property of that estimator class, not a claim that SSH is unrecoverable by any classical route, and a data-assimilating ocean model with an explicit governing system would be a different matter. Within the classes that are deployed for this task, what remains is to predict the training-period mean ($37.71\%$) or to regress SSH on the interpolated temperature and salinity, which at $58.18\%$ and $59.44\%$ depending on the interpolant is worse than the mean and shows that a global linear map does not capture the relationship. Read as complete pipelines, the three classical routes average $21.99\%$ ($k$NN for $T$ and $S$ with its ridge for SSH), $23.35\%$ (IDW throughout) and $18.83\%$ (climatology throughout) across the three channels, against $4.22\%$ for the operator. Recovering SSH at $9.11\%$ from temperature and salinity sensors alone is therefore not an incremental improvement within an existing estimator class; it is a cross-modal reconstruction obtained from a single amortised forward pass, with no governing model evaluated online. That is the claim the ocean problem is here to support, and Section~\ref{sec:supp_ocean_chrono} examines the conditions under which it holds.

The three operator variants are reported for completeness and to document the design choice, not as a ranking: MIMONet-NL, NOMAD and the linear-readout MIMONet span $9.11$--$11.05\%$ on SSH, a spread far smaller than the gap between any of them and the classical alternatives. The nonlinear decoder is retained because it is the best of the three at comparable parameter count (Section~\ref{sec:supp_readout_variants}), but nothing in the argument of this paper depends on which operator architecture wins that comparison.

\subsubsection{Other learned methods on the ocean problem}
\label{sec:supp_ocean_learned}

Two further learned baselines were run on this problem and are reported here because the main text quotes them and because one of them is informative about what the operator formulation is doing.

\begin{table}[h!]
\centering
\caption{\textbf{Learned baselines on the GLORYS12 problem, random day split.} Relative $\ell_2$ error (\%) per channel, mean $\pm$ s.d.\ over five seeds. The FNO is applied to sensor values first rasterised onto the $60\times160$ grid by inverse-distance weighting, and is capacity-matched to MIMONet-NL. KCN operates directly on the scattered sensors through a learned neighbour graph.}
\label{tab:supp_ocean_learned}
\adjustbox{max width=\textwidth}{%
\begin{tabular}{@{}lccc@{}}
\toprule
Method & $T$ (\%) & $S$ (\%) & SSH (\%) \\
\midrule
MIMONet-NL (reported) & $3.10 \pm 0.06$ & $0.46 \pm 0.01$ & $9.11 \pm 0.17$ \\
IDW-gridded FNO & $3.10 \pm 0.07$ & $0.47 \pm 0.01$ & $9.58 \pm 0.08$ \\
KCN & $5.77 \pm 0.23$ & $0.86 \pm 0.01$ & $37.84 \pm 0.28$ \\
\midrule
Climatology (reference) & $17.68 \pm 0.43$ & $1.09 \pm 0.01$ & $37.71 \pm 0.29$ \\
\bottomrule
\end{tabular}}
\end{table}

The FNO result is stated plainly: on this problem it is indistinguishable from MIMONet-NL on the two sampled channels and within half a percentage point on sea-surface height. We do not present this as a defeat or a victory, because the paper's claim is not that one operator architecture outperforms another. It is the opposite of a complication for the argument: two independently constructed operator formulations both recover a field for which no sensor exists, which is stronger evidence that the capability follows from the operator formulation than a result from a single architecture would be.

KCN is the informative case, and it is why this table is included rather than omitted. It reconstructs the two sampled channels competently, as a learned interpolator should, but its sea-surface height error of $37.84\%$ is statistically indistinguishable from the climatology value of $37.71\%$: it has learned to predict the training-period mean and nothing more. KCN builds its estimate from a neighbourhood graph over the sensor locations, so it is a learned interpolant rather than an operator between function spaces, and on a channel with no sensors in the neighbourhood there is nothing for that construction to interpolate. Cross-modal recovery is therefore not a generic consequence of using a neural network on this data, and it is not obtained by any method that remains tied to the geometry of the sensor network. That is the distinction the paper draws, and this row is the control that establishes it.

\begin{figure}[h!]
    \centering
    \includegraphics[width=\textwidth]{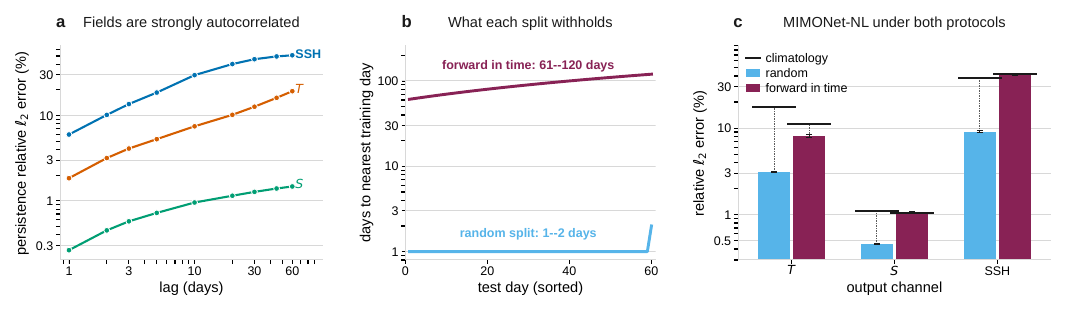}
    \caption{\textbf{Why the day split matters for the ocean problem.} \textbf{a}, Persistence error against lag: the relative $\ell_2$ error incurred by carrying a field forward by the stated number of days, averaged over the record. Daily fields are strongly autocorrelated, so a nearby day is already an informative predictor. \textbf{b}, Distance from each test day to the nearest training day, sorted, under the two protocols. The random split leaves every test day within one or two days of a training day; the forward-in-time split places them $61$ to $120$ days away. \textbf{c}, MIMONet-NL under both protocols, per channel, with the horizontal rules marking the climatology of the same protocol. Persistence is not a competing method and is not plotted as one: it requires the complete field on an adjacent day, whereas the operator observes $800$ point sensors on the target day and never sees another day.}
    \label{fig:supp_ocean_split}
\end{figure}

\subsection{Forward-in-time robustness check for the ocean problem}
\label{sec:supp_ocean_chrono}

The ocean results above use the random day split described in Section~\ref{sec:supp_data_ocean}. Because daily ocean fields are strongly autocorrelated, that protocol places test days next to training days, and it is worth establishing what the reported numbers do and do not demonstrate. This section reports a re-run under a forward-in-time split, and states plainly which parts of the ocean claim survive it.

\paragraph{Why the split is consequential here.} Figure~\ref{fig:supp_ocean_split}a measures the autocorrelation directly: carrying a field forward by one day gives a relative $\ell_2$ error of $1.8\%$ for temperature, $0.3\%$ for salinity and $5.9\%$ for sea-surface height, rising to $19\%$, $1.5\%$ and $51\%$ at a lag of sixty days. Figure~\ref{fig:supp_ocean_split}b shows what each protocol withholds: under the random split every test day has a training day within one or two days, whereas under the forward-in-time split the nearest training day is $61$ to $120$ days away. The random split therefore evaluates reconstruction of states that lie close to states already seen, and the forward split removes that proximity.

We note for completeness that carrying a neighbouring field forward is not a competing method and is not offered as a baseline. It requires the complete field on the adjacent day, all $9{,}457$ cells in all three channels, whereas the operator observes $800$ point sensors on the target day and never sees another day at training or inference. Panel~a bounds how much novelty a test day carries; it does not describe an estimator the operator is being compared against.

\paragraph{Protocol.} The re-run is identical to the reported experiment except in the assignment of days: training uses days $1$--$610$, validation days $611$--$670$ and testing days $671$--$730$, i.e.\ the last two months of the record. Sensor networks, architecture, optimisation schedule, normalisation and seeds are unchanged, and five seeds are run as before. The classical baselines are recomputed on the same split. Table~\ref{tab:supp_ocean_chrono} reports the two protocols side by side.

\begin{table}[h!]
\centering
\caption{\textbf{Ocean reconstruction under both day splits.} Relative $\ell_2$ error (\%) per channel, mean $\pm$ s.d.\ over five seeds; the classical estimators are deterministic given the split and the sensor draw. Under the forward-in-time protocol the split is fixed but the sensor draw is still seed-dependent, so $k$NN and IDW retain a spread; climatology uses no sensors at all and is therefore a single value under that protocol. IDW and $k$NN have no sea-surface-height sensors to interpolate, so no entry exists for them in that column.}
\label{tab:supp_ocean_chrono}
\adjustbox{max width=\textwidth}{%
\begin{tabular}{@{}lcccccc@{}}
\toprule
& \multicolumn{3}{c}{Random day split (reported)} & \multicolumn{3}{c}{Forward in time} \\
\cmidrule(lr){2-4}\cmidrule(lr){5-7}
Method & $T$ (\%) & $S$ (\%) & SSH (\%) & $T$ (\%) & $S$ (\%) & SSH (\%) \\
\midrule
MIMONet-NL & $3.10 \pm 0.06$ & $0.46 \pm 0.01$ & $9.11 \pm 0.17$ & $8.08 \pm 0.27$ & $1.08 \pm 0.01$ & $41.07 \pm 1.02$ \\
NOMAD & $3.49 \pm 0.10$ & $0.51 \pm 0.01$ & $10.72 \pm 0.39$ & $8.10 \pm 0.16$ & $1.08 \pm 0.01$ & $40.37 \pm 0.64$ \\
\midrule
$k$NN & $6.79 \pm 0.32$ & $0.99 \pm 0.04$ & --- & $7.40 \pm 0.28$ & $1.02 \pm 0.04$ & --- \\
IDW & $9.26 \pm 0.49$ & $1.35 \pm 0.06$ & --- & $9.81 \pm 0.48$ & $1.37 \pm 0.06$ & --- \\
Ridge on interpolated $T,S$ & --- & --- & $59.44 \pm 1.02$ & --- & --- & $59.82 \pm 0.24$ \\
Climatology & $17.68 \pm 0.43$ & $1.09 \pm 0.01$ & $37.71 \pm 0.29$ & $11.26$ & $1.04$ & $41.74$ \\
\bottomrule
\end{tabular}}
\end{table}

\paragraph{What the re-run shows.} Three things, and they point in different directions.

First, the degradation is a property of the protocol and of the problem, not of the architecture. MIMONet-NL and NOMAD are statistically indistinguishable under the forward split on all three channels ($8.08$ against $8.10$, $1.08$ against $1.08$, $41.07$ against $40.37$), although they differed under the random split, most visibly on sea-surface height ($9.11$ against $10.72$). Whatever the forward split removes, it removes from both operators equally, so the effect cannot be attributed to a particular fusion rule or decoder.

Second, on the directly observed channels the ordering against the classical interpolants reverses. Under the random split the operators improve on $k$NN by roughly a factor of two on temperature; under the forward split $k$NN is the more accurate of the two ($7.40\%$ against $8.08\%$), and on salinity the two are level. Where a field is sampled at $400$ points and the task is spatial interpolation, a local estimator that fits each test day independently is not disadvantaged by temporal distance, whereas a trained operator is. This is a real limitation and we state it as one: the operator's advantage on directly observed channels, as measured here, depends on the test states resembling the training states.

\begin{figure}[h!]
    \centering
    \includegraphics[width=\textwidth]{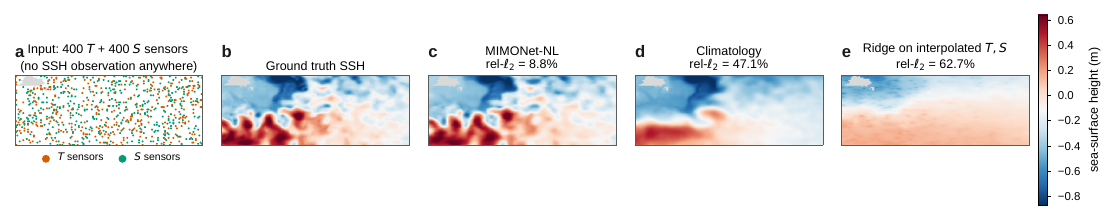}
    \caption{\textbf{Cross-modal reconstruction of an unobserved field.} Sea-surface height on a held-out day of the reported run, the day at the median SSH error of the sixty test days. \textbf{a}, The entire input: $400$ temperature and $400$ salinity point sensors at disjoint locations. No SSH value is supplied anywhere, which is why inverse-distance weighting and $k$-nearest neighbours have no entry for this field: an interpolant is defined by samples of the quantity it interpolates, and here there are none. This is the disjoint sensing and reconstruction regime of the Introduction, in its sharpest form. \textbf{b}, Ground truth. \textbf{c}, MIMONet-NL, reconstructing SSH from the temperature and salinity sensors alone through the learned coupling. \textbf{d}, Climatology, the training-period mean, which is the same field on every day and therefore cannot express the day's mesoscale structure at all. \textbf{e}, Ridge regression from the interpolated temperature and salinity, the only route among the classical estimators compared here that uses the day's observations, which is worse than the climatology. All panels share the colour scale. The point of the comparison is qualitative: the front position and the eddy field in \textbf{c} are recovered from measurements of different physical quantities, whereas \textbf{d} and \textbf{e} show what remains available without an operator.}
    \label{fig:supp_ocean_ssh}
\end{figure}

Third, sea-surface height behaves differently from the other two, and the distinction is the one that matters for the claim this problem supports. No classical \emph{interpolant} of the classes compared here can attempt it: IDW and $k$NN are defined by samples of the field they estimate and there are none, which is why their SSH entries are empty under both protocols. This is a statement about that estimator class and not a claim that SSH is unrecoverable by any classical route. The only classical options are climatology and a regression from interpolated temperature and salinity, and the regression is worse than climatology under both splits ($59.44\%$ and $59.82\%$ against $37.71\%$ and $41.74\%$). The operator remains the only method that produces a cross-modal estimate at all, and Fig.~\ref{fig:supp_ocean_ssh} shows what that difference looks like on a single day. What does not survive the forward split is the \emph{margin}: at $41.07\%$ against a climatology of $41.74\%$ the operator is level with the trivial predictor, where under the random split it was better by a factor of $4.1$.

\paragraph{What we conclude.} The qualitative capability the ocean problem is included to demonstrate, that an operator can infer a field for which no sensor exists by using the learned coupling to fields that are sensed, is demonstrated under the random day split. Under the forward split it is not: at $41.07\pm1.02\%$ against a climatology of $41.74\%$ the SSH estimate is not statistically distinguishable from the trivial predictor. We state this in the same terms used for KCN in Section~\ref{sec:supp_ocean_learned}, because the standard has to be the same: producing a cross-modal estimate at all is a property of the estimator's type signature, not evidence of skill, and we have not measured whether the forward-split output carries day-specific information. An anomaly correlation against the daily SSH anomaly would settle it in one number; we have not run it. What the two protocols jointly establish is that the operator formulation admits cross-modal reconstruction where the compared interpolants do not, and that the accuracy of that reconstruction depends on the test states resembling the training states. A data-assimilating ocean model with an explicit governing system would be a different matter; the contrast drawn here is with the interpolants that are actually deployed when no such model is available at the point of use. The quantitative figure quoted in the main text, a $4.1\times$ reduction against climatology, is specific to the random day split and should be read as such. We attribute the loss under the forward split to the temperature--salinity-to-SSH relationship being calibrated to the mean state of the training period: sea-surface height carries a slowly varying component that the operator cannot observe through its inputs, so a systematic offset accumulates as the test period moves away from the training period. Recalibration on recent data, which is available in any operational setting, is the natural remedy, and quantifying how often it would be required is a well-defined question that we leave open.

\paragraph{Limitation of this check.} The forward test window is the last two months of the record, November and December 2018. Temporal distance from the training period and the particular season are therefore confounded, and this experiment cannot separate them. Splitting the record into contiguous blocks and rotating the test window would do so; we have not run it, and the numbers above should be read as characterising a two-month-ahead gap at this season rather than temporal extrapolation in general.

\section{Uncertainty, robustness and data efficiency}

A virtual sensor reports values at locations where, by construction, no measurement exists to check
them against. An interval is therefore not a refinement but a precondition for using the output at
all, and the same argument applies to knowing how the estimate degrades when the instrumentation
does. This section reports the calibration, noise-robustness, data-efficiency and extrapolation
studies together, because they answer one question: under what conditions the reconstruction should
be trusted.

One restriction applies throughout and is stated once here rather than repeated. Every study in
this section was run on the heat-exchanger case. No uncertainty estimate, noise sweep or
out-of-distribution result is reported for the other five problems, and the properties established
here should not be assumed to transfer to them.

\subsection{Conformal prediction for operator outputs}

The Introduction lists calibrated uncertainty among the properties an operational virtual sensor must supply, and it matters most exactly where the framework is aimed: at interior locations where no measurement exists against which a prediction could be checked. An interval is the only quantitative statement of trustworthiness available there. Conformal prediction provides distribution-free, finite-sample valid uncertainty quantification. We extend the standard framework to spatially distributed operator outputs.

\begin{theorem}[Valid Coverage for Conformal Intervals]
Let \(\{(\boldsymbol u^{(k)}, \mathbf s^{(k)})\}_{k=1}^{N_{\text{cal}}}\) be a held-out calibration set exchangeable with the test distribution. Define normalised residuals
\[
z^{(k)}(\mathbf{r}) = \frac{|s^{(k)}(\mathbf{r}) - \hat{s}^{(k)}(\mathbf{r})|}{\sigma'^{(k)}_G(\mathbf{r}) + \epsilon},
\]
where \(\sigma'_G = s_c\cdot\sigma_G\) is the variance-rescaled predictive standard deviation from MC dropout, \(\sigma_G\) being the raw MC-dropout standard deviation and \(s_c\) a per-channel scaling factor fixed on the same calibration set (using \(s_c\) rather than \(\alpha\) here to free the symbol \(\alpha\) for the conformal miscoverage level). Normalising the score by the same quantity that scales the interval is what makes the two steps consistent. Let \(q_{1-\alpha}\) be the \(\lceil (1-\alpha)(N_{\text{cal}}+1)\rceil\)-th smallest order statistic of \(\{z^{(k)}(\mathbf{r})\}\) over the calibration set (the standard split-conformal quantile). Then the conformal prediction interval
\[
\mathcal{C}(\mathbf{r}) = [\hat{s}(\mathbf{r}) - q_{1-\alpha} \sigma'_G(\mathbf{r}), \, \hat{s}(\mathbf{r}) + q_{1-\alpha} \sigma'_G(\mathbf{r})],
\]
satisfies the split-conformal coverage guarantee
\[
1 - \alpha \;\leq\; \mathbb{P}\!\left[s_{\text{test}}(\mathbf{r}) \in \mathcal{C}(\mathbf{r})\right] \;\leq\; 1 - \alpha + \frac{1}{N_{\text{cal}} + 1}
\]
for any test point exchangeable with the calibration set~\cite{vovk2005algorithmic, lei2018distribution}.
\end{theorem}

\textbf{Implementation.} On the heat-exchanger case the $1{,}546$ samples are split $70/10/20$ into $1{,}082$ training, $154$ calibration and $310$ test cases, so $N_{\text{cal}}=154$. The scaling factor $s_c$ is the per-channel standard deviation of the raw z-scores $(\mathbf s-\hat{\mathbf s})/\sigma_G$ on the calibration set, and $q_{1-\alpha}$ is then taken per channel from the residuals of all $154\times3{,}977 = 612{,}458$ calibration (sample, node) pairs pooled together, at $\alpha=0.05$. Two consequences of pooling should be stated. First, the coverage delivered is \emph{marginal} over the field rather than pointwise at each $\mathbf r$, which is the quantity reported in the main text. Second, nodes within one reconstructed field are spatially correlated, so the pooled residuals are not exchangeable in the strict sense that the finite-sample guarantee assumes; the exchangeable units are the $154$ calibration samples. The guarantee should therefore be read as holding to the accuracy with which pooled nodes approximate exchangeable draws, and it is the measured coverage, not the bound, that establishes calibration here. Table~\ref{tab:supp_conformal} gives the per-channel calibration constants, the coverage before and after rescaling, and the resulting interval half-width.

\begin{table}[h!]
\centering
\caption{\textbf{Conformal calibration on the heat-exchanger case.} All quantities are per output channel, computed on the $N_{\text{cal}}=154$ calibration samples and evaluated on the $310$-case test set at $\alpha=0.05$. $s_c$ rescales the MC-Dropout standard deviation so that the normalised residuals have unit spread; $q_{1-\alpha}$ is the conformal quantile of those rescaled residuals. The calibrated $95\%$ interval is $\hat{\mathbf s}\pm q_{1-\alpha}s_c\sigma_G$, so the last column is the interval half-width expressed as a multiple of the raw MC-Dropout standard deviation. Values are from the reproduction script \texttt{Production/analyses/uq\_repro.py}; the MC-Dropout ensemble is redrawn at each run, so the final digit of the coverages moves by a few hundredths between runs.}
\label{tab:supp_conformal}
\begin{tabular}{@{}lccccc@{}}
\toprule
Channel & $s_c$ & $q_{1-\alpha}$ & Coverage, uncalibrated (\%) & Coverage, conformal (\%) & Half-width ($\times\sigma_G$) \\
\midrule
$p$   & $0.767$ & $2.132$ & $97.79$ & $95.21$ & $1.63$ \\
$u_z$ & $0.901$ & $2.094$ & $95.39$ & $94.97$ & $1.89$ \\
$u_y$ & $1.003$ & $2.083$ & $94.16$ & $94.96$ & $2.09$ \\
$u_x$ & $0.928$ & $2.100$ & $95.08$ & $94.95$ & $1.95$ \\
\midrule
Pooled & --- & --- & --- & $95.02$ & --- \\
\bottomrule
\end{tabular}
\end{table}

Two features of the table are worth reading together. The uncalibrated intervals are not uniformly wrong in the same direction: MC-Dropout is over-dispersed on pressure ($97.79\%$ coverage, $s_c=0.77$, so the raw spread is shrunk) and slightly under-covering on $u_y$ ($94.16\%$), where $s_c=1.00$ says the raw dropout spread is already correctly scaled and the correction comes from the residual tail through the quantile instead. A single global inflation factor would therefore not calibrate all four channels, which is why the scaling is fitted per channel. After calibration every channel sits within $0.25$ percentage points of nominal, and the half-widths differ by roughly $28\%$ across channels, so the calibrated intervals also carry information about which channels the model is genuinely less certain about.

\paragraph{Spatial considerations.} What is delivered and measured above is \emph{marginal} coverage over the field. Simultaneous coverage at every \(\mathbf{r} \in \mathcal{Y}\) is a strictly stronger requirement and would need more conservative quantiles, obtained for instance by a Bonferroni correction or by calibrating on supremum rather than pointwise residuals. Neither is applied here, and no simultaneous guarantee is claimed.

\subsection{Stability under sensor noise: Lipschitz analysis}

Virtual sensing under degraded instrumentation requires that the learned operator not amplify input perturbations. We treat this as a property that the operator is required to have and is verified to have \emph{by measurement}, not as one guaranteed by the training procedure: empirical risk minimisation places no constraint on the Lipschitz constant of the learned map, and the cross-architecture comparison in Section~\ref{sec:supp_gaussian_noise} measures the related but distinct quantity of RMSE inflation under input noise, finding it to vary by more than two orders of magnitude across operators trained on the same data. No Lipschitz constant was measured for those other architectures, so the two quantities should not be equated.

Let \(\tilde{\boldsymbol u} = \boldsymbol u + \boldsymbol \eta\), with \(\boldsymbol \eta \sim \mathcal{N}(\mathbf 0, \sigma_{\text{noise}}^2 \mathbf I)\), denote a noisy input, let \(\hat{\mathcal{G}}_{\boldsymbol \omega}\) be the sub-network realised by dropout mask \(\boldsymbol \omega\), and let \(\bar{\mathcal{G}}(\boldsymbol u) = \mathbb{E}_{\boldsymbol \omega}[\hat{\mathcal{G}}_{\boldsymbol \omega}(\boldsymbol u)]\) be the MC-Dropout estimator. Since the norm is convex, Jensen's inequality gives
\[
\bigl\|\bar{\mathcal{G}}(\tilde{\boldsymbol u}) - \bar{\mathcal{G}}(\boldsymbol u)\bigr\|_{\mathcal{S}}
\;\le\;
\mathbb{E}_{\boldsymbol \omega}\bigl\|\hat{\mathcal{G}}_{\boldsymbol \omega}(\tilde{\boldsymbol u}) - \hat{\mathcal{G}}_{\boldsymbol \omega}(\boldsymbol u)\bigr\|_{\mathcal{S}}
\;\le\;
\mathbb{E}_{\boldsymbol \omega}\!\left[L_{\boldsymbol \omega}\right] \|\boldsymbol \eta\|_{\mathcal{U}},
\]
where \(L_{\boldsymbol \omega}\) is the Lipschitz constant of the masked sub-network. The content of this inequality is exact but limited: averaging over dropout masks cannot make the estimator more sensitive to a perturbation than the \emph{average} of its members. It does not establish that the ensemble mean is less sensitive than the deterministic full network, which is a different quantity, and it yields no numerical bound, since \(\mathbb{E}_{\boldsymbol \omega}[L_{\boldsymbol \omega}]\) is not accessible in closed form. Both the magnitude of the constant and any ordering between inference modes are therefore established empirically, in Section~\ref{sec:supp_lipschitz} and in the main text.

\subsection{Sample complexity and convergence}
\label{sec:supp_sample_complexity}

Amortisation moves cost from inference to training, so the size of that up-front cost is part of the argument: a virtual sensor that needs an unattainable number of high-fidelity solutions is not deployable however fast it runs afterwards. This subsection measures how the error falls with the number of training solutions.

The sample-complexity sweep reports the held-out test-set error as a function of the training-pool size $N_{\text{train}}$. MIMONet was trained from a fresh, identically-seeded initialisation on four nested random subsets of each case's training pool; the fifth and largest point is the published checkpoint rather than a sweep run, which has a consequence for the LDC case that is stated below. The three cases reported here are the PWR subchannel (pool size $3{,}200$, $N_{\text{train}}\in\{320,\,800,\,1{,}600,\,2{,}400,\,3{,}200\}$), the heat exchanger (pool size $1{,}082$ from the 70/10/20 split, $N_{\text{train}}\in\{108,\,270,\,541,\,812,\,1{,}082\}$), and the LDC ($N_{\text{train}}\in\{320,\,800,\,1{,}600,\,2{,}400,\,3{,}200\}$ subsampled from the combined LDC training and validation pool of $3{,}949 = 3{,}159 + 790$ samples). The swept points share the architecture, optimiser (Adam, $\mathrm{lr}=10^{-3}$, weight decay $10^{-6}$), and batch size ($4$). The LR schedule is \texttt{ReduceLROnPlateau} (patience $50$, factor $0.5$, $\mathrm{lr}_{\min}=10^{-6}$) for subchannel and heat exchanger, and \texttt{CosineAnnealingLR} ($\eta_{\min}=10^{-6}$, $T_{\max}=500$) for LDC (the plateau detector was found unreliable across $N$ on LDC, so a deterministic cosine schedule is used to remove the dependency on val-loss noise). A fixed held-out validation split (800 / 154 / 749 samples for subchannel / heat exchanger / LDC) is held constant across the swept conditions in each case; the LDC sweep therefore partitions the $3{,}949$-sample pool as $3{,}200/749$ rather than as the $3{,}159/790$ split of record given in Table~\ref{tab:supp_overview}. Test errors are evaluated on the fixed, disjoint test set of each case ($1{,}000$ / $310$ / $988$ samples).

\noindent For $N_{\text{train}}=N_{\max}$ the headline checkpoint reported throughout the main text is reused to anchor the curve at the published per-channel relative $\ell_2$ values, and the consequence differs by case. On the subchannel and heat exchanger the headline configuration coincides with the sweep configuration, so the endpoint is continuous with the other four points. On the LDC it is not: the headline model was trained with batch size $8$, weight decay $10^{-7}$ and a plateau schedule on the $3{,}159$-sample split (Table~\ref{tab:supp_training}), whereas the swept points use batch size $4$, weight decay $10^{-6}$ and cosine annealing. The LDC endpoint is thus plotted at a nominal $N_{\text{train}}=3{,}200$ although the checkpoint was trained on $3{,}159$ samples, and it differs from the other four points in optimiser configuration as well. The fitted exponent $\gamma_{\text{LDC}} = 0.58 \pm 0.23$ is computed across that seam and should be read as indicative rather than as a controlled measurement. The subchannel and heat-exchanger exponents are unaffected.

\begin{figure}[h!]
    \centering
    \includegraphics[width=1.00\textwidth]{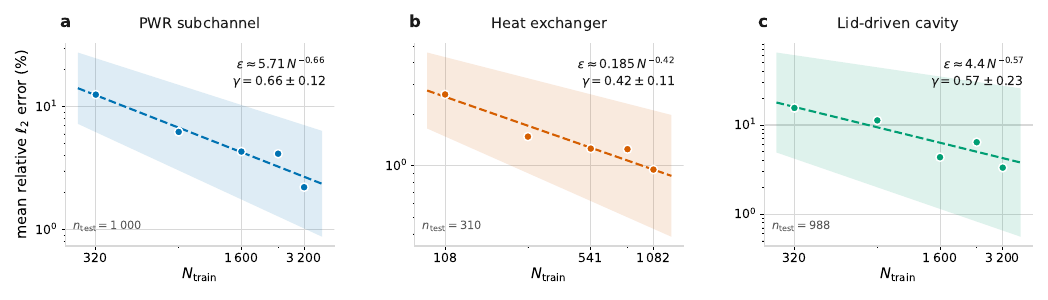}
    \caption{\textbf{Sample-complexity convergence.} Mean relative $\ell_2$ error on the held-out test set as a function of $N_{\text{train}}$, with a power-law fit $\epsilon(N) = a\cdot N^{-\gamma}$. \textbf{(a)} PWR subchannel ($1{,}000$-sample test set). Channel-mean errors $12.5\%$, $6.2\%$, $4.3\%$, $4.1\%$, $2.2\%$ at $N_{\text{train}}=320$, $800$, $1{,}600$, $2{,}400$, $3{,}200$; fit $\epsilon \approx 571\,N^{-0.66}$, $\gamma = 0.66 \pm 0.12$. \textbf{(b)} Heat exchanger ($310$-sample test set). Channel-mean errors $2.61\%$, $1.48\%$, $1.26\%$, $1.25\%$, $0.95\%$ at $N_{\text{train}}=108$, $270$, $541$, $812$, $1{,}082$; fit $\epsilon \approx 18.5\,N^{-0.43}$, $\gamma = 0.43 \pm 0.11$. \textbf{(c)} LDC ($988$-sample test set). Channel-mean errors $15.6\%$, $11.3\%$, $4.4\%$, $6.4\%$, $3.3\%$ at $N_{\text{train}}=320$, $800$, $1{,}600$, $2{,}400$, $3{,}200$; fit $\epsilon \approx 440\,N^{-0.58}$, $\gamma = 0.58 \pm 0.23$. Filled circles are per-condition errors; dashed curves are the least-squares fits with the 95\% confidence interval on the exponent.}
    \label{fig:sample_complexity}
\end{figure}

\noindent The convergence curves for all three cases are shown in Fig.~\ref{fig:sample_complexity}. All three fitted exponents ($\gamma_{\text{Sub}} = 0.66\pm0.12$, $\gamma_{\text{HX}} = 0.43\pm0.11$, $\gamma_{\text{LDC}} = 0.58\pm0.23$) are consistent with the operator-approximation rate bounds for smooth neural-operator classes~\cite{kovachki2023neural,de_ryck_mishra_2022}. The subchannel curve flattens between $N_{\text{train}}=1{,}600$ and $N_{\text{train}}=2{,}400$ (log--log slope $-0.10$ over a $1.5\times$ pool-size increase), then resumes its decline to $2.2\%$ at the full pool. We do not read the full-pool point as convergence: the error falls by $47\%$ over the final $1.33\times$ of data, so the curve is still descending steeply at its endpoint and the pool size cannot be described as saturating. The heat-exchanger curve descends monotonically from $2.6\%$ at $N_{\text{train}}=108$ to the headline $0.95\%$ at $N_{\text{train}}=1{,}082$ with a smaller exponent, reflecting that the HX virtual-sensing problem is data-efficient (low-error regime is reached with $\sim 5\times$ fewer training samples than for subchannel) while still benefiting from additional data up to the full pool. The LDC curve descends from $15.6\%$ at $N_{\text{train}}=320$ to the headline $3.3\%$ at $N_{\text{train}}=3{,}200$. One data point, $N_{\text{train}}=2{,}400$, sits above the power-law prediction (measured $6.4\%$ vs.\ fit $\approx 4.8\%$, a $+1.6$ percentage-point residual); the confidence interval on the fitted exponent ($\pm 0.23$) reflects the additional fitting uncertainty introduced by this point but is a statement about the exponent, not a prediction interval that contains the residual itself. We attribute the elevation at $N_{\text{train}}=2{,}400$ to a known sensitivity of the LDC optimisation to which $2{,}400$-sample subset is drawn from the $3{,}949$-sample pool (the LDC training-pool re-uses a deterministic random permutation seed of $42$; different seeds shift this point but not the overall trend), rather than to a violation of the power-law regime. The endpoint at $N_{\text{train}}=3{,}200$ matches the headline value reported in the main text.

\subsection{Empirical Lipschitz characterisation}
\label{sec:supp_lipschitz}

To quantify this directly, the empirical Lipschitz constant of the deterministic-mode MIMONet operator was measured on the held-out heat-exchanger test set. For $1{,}000$ random perturbation directions $\boldsymbol{\delta}$ drawn at each of $20$ magnitudes $\|\boldsymbol{\delta}\|_2 \in [0.01,\,0.50]$ in the combined branch-input space $\mathbb{R}^{102}$, the ratio $\|\mathcal{G}_\theta(\boldsymbol{u}+\boldsymbol{\delta}) - \mathcal{G}_\theta(\boldsymbol{u})\|_2 / \|\boldsymbol{\delta}\|_2$ was computed per output channel and aggregated to its 99th percentile. The combined boundary vector was split back into the two encoders so that perturbations propagate through both functional ($q_{\text{wall}}$) and scalar ($T_{\text{in}}, u_{\text{in}}$) input modalities. Inputs $\boldsymbol{u}$ were sampled uniformly from the $310$ test cases.

\begin{figure}[h!]
    \centering
    \includegraphics[width=1.00\textwidth]{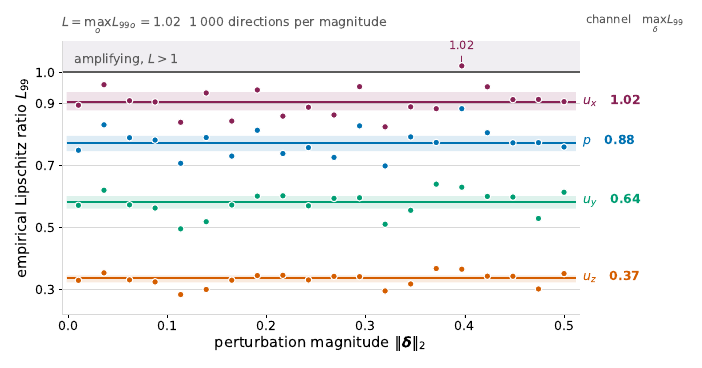}
    \caption{\textbf{Empirical Lipschitz characterisation on the heat-exchanger test set.} Empirical 99th-percentile Lipschitz ratio $L_{99}=\|\mathcal{G}(\boldsymbol{u}+\boldsymbol{\delta})-\mathcal{G}(\boldsymbol{u})\|_2/ \|\boldsymbol{\delta}\|_2$ of MIMONet on the heat-exchanger test set, against perturbation magnitude $\|\boldsymbol{\delta}\|_2$, for the four output channels $\{p, u_z, u_y, u_x\}$. Each marker aggregates $1{,}000$ random perturbation directions at that magnitude; the $20$ magnitudes are independent measurements and are therefore not joined. For each channel the horizontal line is the median over the $20$ magnitudes and the band the interquartile range, so a narrow band indicates a ratio that does not drift with $\|\boldsymbol{\delta}\|_2$. The right-hand labels give $\max_\delta L_{99}$ per channel: $0.88$ ($p$), $0.37$ ($u_z$), $0.64$ ($u_y$) and $1.02$ ($u_x$), the last of which sets the global proxy $L=\max_o L_{99,o}=1.02$. The shaded region above the solid line is where the operator would amplify a perturbation; exactly one of the $80$ measurements enters it, the $u_x$ value at $\|\boldsymbol{\delta}\|_2 = 0.40$, and it does so by two per cent. The flatness across a fiftyfold range of $\|\boldsymbol{\delta}\|_2$ is the behaviour expected of a smooth operator that is locally linear in this neighbourhood of the test manifold.}
    \label{fig:sample_lipchitz}
\end{figure}

The empirical Lipschitz characterisation is shown in Figure~\ref{fig:sample_lipchitz}. The 99th-percentile bounds are $L_p \approx 0.88$ (pressure), $L_{u_z}\approx 0.37$, $L_{u_y}\approx 0.64$, and $L_{u_x}\approx 1.02$, yielding the global $99$th-percentile proxy $L \approx 1.02$ reported. This is a measured percentile and not a bound: by construction one per cent of the sampled ratios exceed it. The ratios are essentially flat across the entire $\|\boldsymbol{\delta}\|_2$ range, indicating that the operator response is locally linear in this neighbourhood of the test manifold; this is the expected behaviour of a smooth approximated operator. The connection to the relative noise-robustness results in the main text is as follows: a bounded $L$ enforces a worst-case \emph{absolute} scaled-output deviation of $\leq L\,\|\boldsymbol\eta\|_2$ for an input perturbation $\boldsymbol\eta$. The two studies parametrise the perturbation differently and the conversion must be stated: the sweep above is indexed by the vector norm $\|\boldsymbol\delta\|_2 \in [0.01,\,0.50]$, whereas the Gaussian study of Section~\ref{sec:supp_gaussian_noise} is indexed by a \emph{per-component} standard deviation $\sigma_{\text{noise}}$, so that $\|\boldsymbol\eta\|_2 \approx \sigma_{\text{noise}}\sqrt{102} \approx 5.05$ at $\sigma_{\text{noise}}=0.5$. The measured envelope therefore covers the Gaussian sweep only up to $\sigma_{\text{noise}}\approx 0.05$, that is, its first non-zero level, and the linearity observed here cannot be extrapolated to the larger noise levels; that regime is characterised by direct measurement in Section~\ref{sec:supp_gaussian_noise} instead. Within the range actually probed, the deterministic-pass \emph{relative} RMSE inflation is much smaller than the worst-case envelope; MC-Dropout averaging further reduces it to the sub-$3\%$ level reported in the main text by marginalising over weight-space realisations.

\subsection{Structured (non-Gaussian) sensor-noise robustness}
\label{sec:supp_structured_noise}

The Gaussian noise model used in the main text isolates the operator's response to isotropic, zero-mean perturbations and serves as the simplest analytical proxy for sensor degradation. Real sensor failures in nuclear environments rarely follow this idealisation: instead, sensors exhibit systematic drift (radiation-induced bias), temporally correlated noise (thermal-electronic coupling between channels), and intermittent dropout (radiation-induced single-event upsets or temporary obstruction). To examine MIMONet's robustness beyond simulated smooth-noise data, three structured-noise models were applied to the held-out heat-exchanger test set, each evaluated against the MC-Dropout ensemble ($N=20$) baseline.

\paragraph{Sensor drift.} A spatially uniform additive bias was applied to all branch inputs, $\tilde{u} = u + d \cdot \boldsymbol{1}$, with $d \in \{0,\,0.02,\,0.04,\,0.06,\,0.08,\,0.10\}$ corresponding to up to 10\% of the normalised input range. This emulates a slow drift in calibration shared across the wall heat-flux profile $q(z)$ and the inlet scalars $(T_{\text{in}}, u_{\text{in}})$.

\paragraph{Correlated (AR(1)) noise.} Temporally correlated perturbations were generated via a first-order autoregressive process $\varepsilon_t = \rho\,\varepsilon_{t-1} + \sigma_{\text{AR}}\,\xi_t$ with $\rho = 0.8$, $\sigma_{\text{AR}}=0.1$, and $\xi_t\sim\mathcal{N}(0,1)$. Ten independent realisations were averaged. The strong correlation makes this model substantially harder to suppress than i.i.d.\ Gaussian noise because successive sensor readings carry similar systematic error.

\paragraph{Sensor dropout.} A fraction $p_{\text{drop}}=10\%$ of input channels (selected uniformly per sample) was zeroed, modelling intermittent failure of individual sensor lines.

\begin{figure}[h!]
    \centering
    \includegraphics[width=1.00\textwidth]{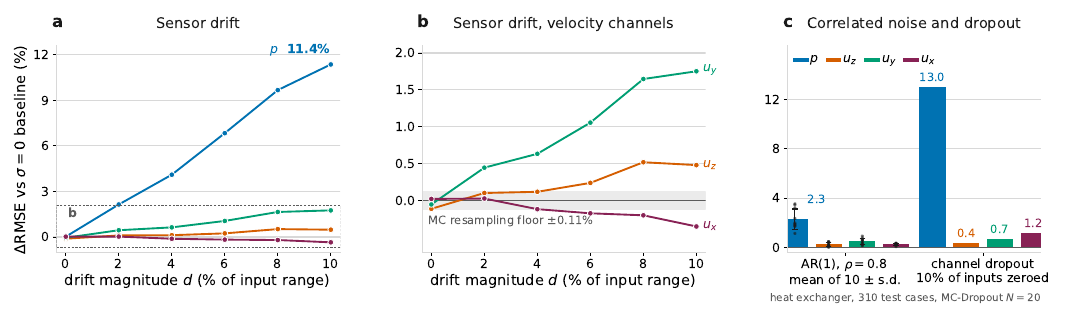}
    \caption{\textbf{Structured (non-Gaussian) sensor-noise robustness.} Relative RMSE inflation $\Delta\mathrm{RMSE}$ (\%) of the MC-Dropout MIMONet ($N=20$) relative to the $\sigma=0$ baseline, on the $310$-case heat-exchanger test set, for the four output channels $\{p, u_z, u_y, u_x\}$. \textbf{a}, Uniform additive sensor drift, swept over $d \in \{0, 0.02, \ldots, 0.10\}$ of the normalised input range. The pressure response is close to linear in $d$, reaching $11.4\%$ at $d=10\%$, which is the behaviour expected of a globally coupled elliptic field whose interior reconstruction depends on the full boundary-input vector: a coordinated bias does not average out. \textbf{b}, The same sweep on the velocity window marked by the dashed rectangle in \textbf{a}. All three velocity channels stay below $2\%$ over the entire sweep. The shaded band is the $\pm0.11\%$ Monte-Carlo resampling floor measured at $d=0$ from the independent dropout draws of the baseline and the $d=0$ evaluation, and values inside it are not distinguishable from no effect. $u_z$ and $u_y$ rise slightly but monotonically out of it; $u_x$ moves in the opposite direction, falling to $-0.35\%$ at $d=0.10$, about three times the floor. A small reduction in error under a coordinated input bias is not a result we can explain from the architecture, and we report it as measured rather than interpret it. \textbf{c}, AR(1) correlated noise ($\rho=0.8$, $\sigma_{\text{AR}}=0.1$) and random channel dropout ($10\%$ of input channels zeroed per sample). Bars are means, dots the ten individual AR(1) realisations and whiskers their standard deviation; the AR(1) spread is appreciable relative to the mean, $2.28 \pm 0.79\%$ for pressure, so single realisations range from $1.12\%$ to $3.54\%$. Every channel \emph{mean} stays below $3\%$ under AR(1) despite the strong temporal correlation, though the largest single pressure realisation reaches $3.54\%$, while dropout reproduces the pressure sensitivity seen under drift ($13.0\%$). Taking the worst case over all levels of all three models, the velocity channels remain below $2\%$ and pressure between $11\%$ and $13\%$.}
    \label{fig:structured_noise}
\end{figure}

\noindent The $\Delta\mathrm{RMSE}$ results are shown in the main text; the measured values are tabulated in Table~\ref{tab:supp_structured_noise} and the full sweeps are plotted in Figure~\ref{fig:structured_noise}.

\begin{table}[h!]
\centering
\caption{\textbf{Structured-noise inflation, tabulated.} $\Delta\mathrm{RMSE}$ (\%) of MC-Dropout MIMONet under structured noise on the heat-exchanger test set, relative to the $\sigma=0$ baseline. The drift row is the worst value over the six sweep levels rather than the value at $d=0.10$: the two coincide for $p$ and $u_y$, whereas the $u_z$ maximum occurs at $d=0.08$ (the $d=0.10$ value is $0.48$) and the $u_x$ maximum at $d=0.02$, the error at $d=0.10$ being $0.35$ below baseline. Figure~\ref{fig:structured_noise} shows every level. Computed channelwise on 310 test cases with 20 stochastic forward passes per evaluation.}
\label{tab:supp_structured_noise}
\begin{tabular}{@{}lcccc@{}}
\toprule
\textbf{Noise model} & $p(z,y)$ & $u_z$ & $u_y$ & $u_x$ \\
\midrule
Sensor drift, worst over $d\in[0,0.10]$ & 11.36 & 0.52 & 1.75 & 0.03 \\
AR(1), $\rho=0.8$, mean of 10 realisations & 2.28 & 0.24 & 0.48 & 0.25 \\
Random sensor dropout, $p_{\text{drop}}=10\%$ & 13.02 & 0.35 & 0.67 & 1.18 \\
\bottomrule
\end{tabular}
\end{table}

\textbf{Interpretation.} All three velocity channels remain below 2\% inflation under every structured-noise model, so that under every structured-noise model tested the velocity error stays within a small fraction of its clean value. We avoid the term ``operationally robust'' here because no application-specific tolerance has been defined against which robustness could be judged; what is established is the measured inflation, not its acceptability for a given deployment. Pressure exhibits a larger 11--13\% sensitivity to systematic drift and dropout. This channel-dependent behaviour is physically expected: pressure is a globally coupled elliptic field whose interior reconstruction depends on the full boundary-input vector, so systematic bias and missing channels propagate through the entire field rather than averaging out. By contrast, velocity components are governed by local advective--diffusive transport and are correspondingly less sensitive to coordinated bias. The AR(1) result is the most encouraging: even with strong $\rho=0.8$ temporal correlation that mimics realistic sensor coupling, $\Delta\mathrm{RMSE}$ stays below 3\% on every channel. The MC-Dropout averaging provides the noise-suppression mechanism by marginalising over weight-space realisations of the operator manifold, an effect that becomes especially valuable for the smoother (correlated) noise spectra encountered in deployed reactor instrumentation. Taken together, these results probe non-idealised sensor noise, which the isotropic Gaussian model does not given that ground-truth interior fields cannot be measured in operating reactors.

\subsection{Gaussian sensor-noise inflation, tabulated}
\label{sec:supp_gaussian_noise}

The main text quotes channel-mean RMSE inflation at $\sigma_{\text{noise}}=0.5$ for four operator architectures. The complete per-channel numbers are given in Table~\ref{tab:supp_gaussian_noise}, together with the protocol.

\paragraph{Protocol.} Isotropic Gaussian noise $\eta\sim\mathcal N(\mathbf 0,\sigma_{\text{noise}}^2\mathbf I)$ is added to the scaled branch inputs of the heat-exchanger case, at eleven levels $\sigma_{\text{noise}}\in\{0, 0.05, \ldots, 0.50\}$ in normalised input units, and the test-set RMSE is recomputed per output channel on all $310$ held-out cases. Every architecture receives the identical perturbed inputs. The reported quantity is the relative inflation $\Delta\mathrm{RMSE} = [\mathrm{RMSE}(\sigma) - \mathrm{RMSE}(0)]/\mathrm{RMSE}(0)$, so each model is measured against its own clean baseline and the comparison is of noise sensitivity rather than of accuracy. The MC-Dropout entry averages $N=20$ stochastic passes at every noise level; the deterministic entry is a single pass through the same trained weights with dropout disabled, so the two rows differ in inference mode and not in training.

\begin{table}[h!]
\centering
\caption{\textbf{Gaussian sensor-noise sensitivity across operator architectures.} Clean-input RMSE and relative RMSE inflation at $\sigma_{\text{noise}}=0.5$, per output channel, on the $310$-case heat-exchanger test set. Inflation is measured against each model's own $\sigma=0$ baseline, which is given in the upper block so that the two can be read together: a small inflation on a large baseline is not evidence of accuracy.}
\label{tab:supp_gaussian_noise}
\adjustbox{max width=\textwidth}{%
\begin{tabular}{@{}lccccc@{}}
\toprule
Model & $p$ & $u_z$ & $u_y$ & $u_x$ & Mean \\
\midrule
\multicolumn{6}{@{}l}{\emph{RMSE at $\sigma_{\text{noise}}=0$ (scaled units)}} \\
MIMONet, deterministic & $0.0245$ & $0.0358$ & $0.0604$ & $0.1302$ & --- \\
MIMONet, MC-Dropout ($N{=}20$) & $0.0245$ & $0.0354$ & $0.0557$ & $0.1117$ & --- \\
GeoFNO & $0.0152$ & $0.0193$ & $0.0231$ & $0.0313$ & --- \\
NOMAD & $0.0445$ & $0.0585$ & $0.0672$ & $0.0935$ & --- \\
KCN & $0.0585$ & $0.0499$ & $0.0670$ & $0.1054$ & --- \\
\midrule
\multicolumn{6}{@{}l}{\emph{$\Delta\mathrm{RMSE}$ (\%) at $\sigma_{\text{noise}}=0.5$}} \\
MIMONet, deterministic & $42.86$ & $5.31$ & $6.13$ & $2.53$ & $14.21$ \\
MIMONet, MC-Dropout ($N{=}20$) & $5.71$ & $1.13$ & $2.33$ & $0.63$ & $2.45$ \\
GeoFNO & $555.92$ & $227.46$ & $234.20$ & $258.15$ & $318.93$ \\
NOMAD & $67.19$ & $37.61$ & $38.10$ & $111.34$ & $63.56$ \\
KCN & $0.85$ & $0.40$ & $0.60$ & $3.98$ & $1.46$ \\
\bottomrule
\end{tabular}}
\end{table}

Two entries need reading with care, and the clean-input block is included so that they can be. KCN shows almost no inflation, but it starts from the worst clean RMSE of the five on pressure, and is mid-pack on the three velocity channels; its neighbour graph and edge weights are constructed once from the training-mesh coordinates and applied as a fixed operator at inference, so most of the noise pathway is short-circuited by a structure that is not responding to the input in the first place. Low sensitivity of that kind is not a virtue. The MC-Dropout and deterministic rows share their weights and differ only in inference mode, so their comparison isolates the effect of averaging over dropout masks; the MC-Dropout variant is reported in the main text as an uncertainty-aware mode of the same model rather than as a capacity-matched single-pass competitor.

\subsection{Local parameter extrapolation, tabulated}
\label{sec:supp_ood}

The main text reports out-of-distribution behaviour on the heat-exchanger case. The per-channel values are given in Table~\ref{tab:supp_ood}.

\paragraph{Protocol.} Fifty additional ANSYS Fluent simulations were generated with the inlet velocity, inlet temperature and wall heat-flux amplitude placed approximately $15\%$ outside their training ranges, twenty-five below the lower boundary and twenty-five above the upper. Geometry, solver configuration, convergence criteria, data processing and the target plane at $x=0.789$ are unchanged from the in-distribution dataset, and the trained models are evaluated on these cases without retraining or fine-tuning.

\begin{table}[h!]
\centering
\caption{\textbf{MIMONet under local parameter extrapolation.} Relative $\ell_2$ error (\%) per channel on the in-distribution test set and on the out-of-distribution cases, split by whether the operating parameters lie below or above the training range. The final column is the ratio of the pooled out-of-distribution error to the in-distribution error.}
\label{tab:supp_ood}
\begin{tabular}{@{}lccccc@{}}
\toprule
Condition & $p$ & $u_z$ & $u_y$ & $u_x$ & Channel mean \\
\midrule
In-distribution ($n=310$) & $0.82$ & $1.45$ & $1.02$ & $0.52$ & $0.95$ \\
OOD, below range ($n=25$) & $2.04$ & $2.81$ & $2.60$ & $1.23$ & $2.17$ \\
OOD, above range ($n=25$) & $0.92$ & $1.92$ & $1.63$ & $0.88$ & $1.34$ \\
OOD, pooled ($n=50$) & $1.48$ & $2.37$ & $2.11$ & $1.05$ & $1.75$ \\
\midrule
OOD $\div$ in-distribution & $1.81$ & $1.63$ & $2.07$ & $2.02$ & $1.84$ \\
\bottomrule
\end{tabular}
\end{table}

The degradation is graded rather than abrupt: a $15\%$ excursion in the operating parameters costs a factor of $1.84$ in channel-mean error, and the worst per-channel factor is $2.07$. The asymmetry between the two directions is consistent across channels, extrapolation below the training range costing roughly $1.6\times$ more than extrapolation above it, which is expected because the lower boundary approaches the laminar-transitional regime where the field structure changes more rapidly with the parameters. No claim of extrapolation beyond this margin is made; the experiment establishes that the operator degrades continuously over a $15\%$ excursion, not that it remains valid at larger ones.

\section{Computational complexity}

The property at issue is amortisation. A virtual sensor is useful inside a monitoring loop only if
the cost of producing a field does not scale with the effort that went into learning it, and the
operator formulation moves that cost to an offline training stage. This section states the forward
cost, compares it with the alternatives, and is explicit about which rows of that comparison were
measured here and which are reference figures taken from the literature.
\label{sec:supp_complexity}

\subsection{Forward-pass complexity}

The requirement at issue here is amortisation: a virtual sensor must answer a new operating condition within a monitoring loop, without solving a governing system online and without refitting. The costs below are what that amortisation buys, and they are the analytic counterpart of the measured latencies in Section~\ref{sec:supp_realtime}.

For batch size \(B\), query points \(P\), latent dimension \(I\), and \(m\) output channels:

\begin{itemize}
\item \textbf{Branch networks:} Each branch with \(L\) layers of width \(W\): \(\mathcal{O}(B \cdot L \cdot W^2)\) per branch.
\item \textbf{Multiplicative fusion:} \(\mathcal{O}(B \cdot I \cdot n)\) where \(n\) is number of input modalities.
\item \textbf{Trunk network:} \(\mathcal{O}(P \cdot L' \cdot W'^2)\) for trunk with \(L'\) layers of width \(W'\).
\item \textbf{Einstein summation:} \(\mathcal{O}(B \cdot P \cdot I \cdot m)\).
\end{itemize}

\textbf{Total:} \(\mathcal{O}(B \cdot P \cdot I \cdot m)\) dominates for large \(P\), yielding linear scaling in the number of query points.

\subsection{Comparison with competing methods}

\noindent The asymptotic costs of the methods compared in this work are collected in
Table~\ref{tab:supp_complexity}.

\begin{table}[h!]
\centering
\caption{\textbf{Computational complexity comparison for operator learning methods.} $P$ is the number of query coordinates, $I$ the branch--trunk latent width and $n$ the number of conditioning observations. The finite-element row is an order-of-magnitude figure taken from the literature and not a measurement reported here.}
\label{tab:supp_complexity}
\begin{tabular}{lcc}
\hline
\textbf{Method} & \textbf{Training} & \textbf{Inference} \\
\hline
Finite Element Method & \(\mathcal{O}(P^{1.5-2})\) & \(\mathcal{O}(P^{1.5-2})\) per solve \\
Gaussian Process & \(\mathcal{O}(n^3)\) & \(\mathcal{O}(P n)\) \\
Fourier Neural Operator & \(\mathcal{O}(P \log P)\) & \(\mathcal{O}(P \log P)\) per mode \\
\textbf{MIMONet (ours)} & \(\mathcal{O}(P \cdot I)\) & \(\mathcal{O}(P \cdot I)\) \\
\hline
\end{tabular}
\end{table}

Here \(P\) is the number of query coordinates and \(n\) the number of conditioning observations; the two are distinct, and conflating them is what produces the common claim that operators are asymptotically cheaper than Gaussian processes at query time. They are not: both are linear in \(P\). The difference is where the cubic term sits, as set out in Section~\ref{sec:supp_theory}. No solver timing is measured in this work, so the comparison with the finite-element row is an order-of-magnitude statement taken from the literature rather than a measurement reported here.

\subsection{Memory footprint}

\textbf{Training:} \(\mathcal{O}(N_{\text{params}} + B \cdot P \cdot m)\) for model parameters and batch activations.
\noindent
\textbf{Inference:} \(\mathcal{O}(N_{\text{params}} + P \cdot I)\), independent of training set size. Measured peak inference VRAM is $2.4$--$3.5\,$GB on the data-centre GPUs benchmarked here (Table~\ref{tab:supp_realtime}); no embedded or edge platform was tested, and we make no claim about one.

\section{Real-time inference benchmarking across GPU architectures}
\label{sec:supp_realtime}

The efficiency comparison in the main text is made on a single GPU. This section reports the same workload across four hardware generations, so that the architectural conclusions can be separated from the machine they were measured on.

Each metric is the mean over $N = 1{,}000$ independent inferences; per-run spreads were not retained by the benchmarking harness and are therefore not tabulated. Latency ($\bar{t}$) is the average forward-pass duration per full-field reconstruction, throughput ($\mathrm{TPS}$) is the reciprocal of latency, power ($\bar{P}$) is the mean instantaneous GPU power, and energy ($\bar{E}$) is the average per-inference energy cost computed as $\bar{E} = \bar{P}\,\bar{t}$. VRAM denotes the peak GPU memory usage during inference. All tests were performed using identical trained MIMONet weights with batch size $B = 1$ and trunk query size $P = 3{,}977$.

\begin{table}[!ht]
\caption{\textbf{MIMONet single-inference hardware-portability measurements.} HeatExchanger case ( $B{=}1$, $P{=}3{,}977$, $N{=}1{,}000$ timed forward passes). Latencies are wall-clock means over $1{,}000$ passes after $10$ warmups; power is sampled via NVML and averaged across the timed window; energy = power $\times$ latency. The PWR-subchannel and LDC cases give the same qualitative ordering and are reported in Supplementary Table~\ref{tab:supp_efficiency_full}.}
\centering
\adjustbox{max width=\textwidth}{%
\begin{tabular}{lcccccc}
\hline
\textbf{GPU} & \textbf{Architecture} & $\bar{t}$ (ms) &
$\mathrm{TPS}$ (samples~s$^{-1}$) & $\bar{P}$ (W) &
$\bar{E}$ (mJ) & VRAM (GB) \\
\hline
NVIDIA A40 & Ampere & $0.84$ & $1{,}190$ & $72.6$ & $61.2$ & 2.38 \\
NVIDIA A100 (40~GB) & Ampere & $0.81$ & $1{,}230$ & $124.9$ & $100.4$ & 2.54 \\
NVIDIA H200 & Hopper & $\mathbf{0.35}$ & $\mathbf{2{,}860}$ & $129.2$ & $\mathbf{45.6}$ & 3.00 \\
NVIDIA GH200 (120~GB) & Grace Hopper & $1.17$ & $855$ & $156.7$ & $182.6$ & 3.45 \\
\hline
\end{tabular}%
}
\label{tab:supp_realtime}
\end{table}

\begin{figure}[h!]
    \centering
    \includegraphics[width=0.85\textwidth]{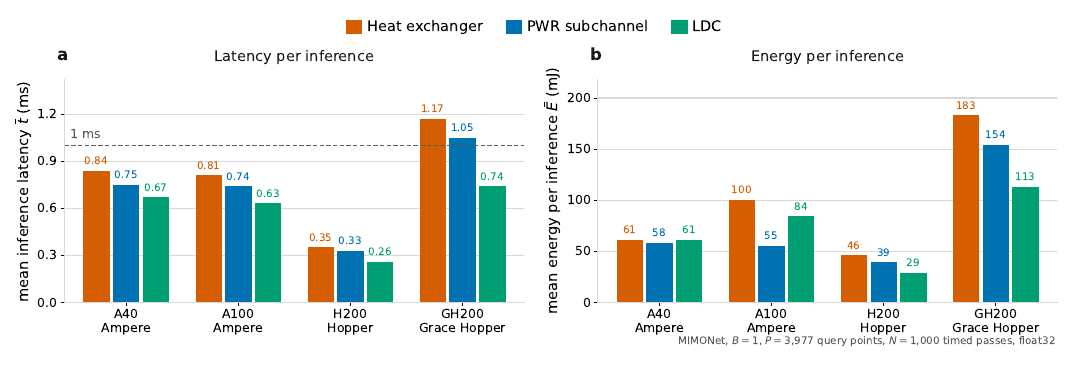}
    \caption{\textbf{Hardware portability of MIMONet across NVIDIA data-centre GPUs.} Measurements on the A40, A100, H200 and GH200 under identical workload conditions ($B{=}1$, $P{=}3{,}977$ query points, $N{=}1{,}000$ timed forward passes, float32), reported separately for each of the three case studies. \textbf{a}, Mean inference latency $\bar{t}$; the dashed line marks $1\,\mathrm{ms}$. \textbf{b}, Mean energy per inference $\bar{E} = \bar{P}\,\bar{t}$, with $\bar{P}$ the mean NVML board power over the timed window. Three of the four GPU generations achieve sub-millisecond single-inference latency on every case, and the slowest measurement anywhere is $1.17\,\mathrm{ms}$ on the GH200. The Hopper-generation H200 is both fastest and most energy-efficient ($\bar{t} \le 0.35\,\mathrm{ms}$, $\bar{E} \le 46\,\mathrm{mJ}$ across the three cases). The GH200 yields larger latency and energy than the H200 at this workload because Grace Hopper's coherent CPU--GPU interconnect is optimised for larger working sets and does not improve the launch envelope of a single $B{=}1$ inference. The architectural-efficiency comparison across operator families on a single GPU is given in the main text; this figure isolates the hardware-generation contribution under a fixed model. All values are the MIMONet column of Supplementary Table~\ref{tab:supp_efficiency_full}.}
    \label{fig:supp_realtime_inference}
\end{figure}

\noindent The measurements are given in Table~\ref{tab:supp_realtime} and plotted in Figure~\ref{fig:supp_realtime_inference}. Single-inference MIMONet latency is sub-millisecond on three of the four GPUs tested on every case study, and the slowest measurement anywhere is $1.17\,\mathrm{ms}$ on the GH200. The lowest latency and energy are obtained on the Hopper-generation H200 ($0.26$--$0.35\,\mathrm{ms}$, $29$--$46\,\mathrm{mJ}$ across the three cases), followed by A40 ($0.67$--$0.84\,\mathrm{ms}$, $58$--$61\,\mathrm{mJ}$) and A100 ($0.63$--$0.81\,\mathrm{ms}$, $55$--$100\,\mathrm{mJ}$). The Grace Hopper GH200 is the outlier at this workload ($0.74$--$1.17\,\mathrm{ms}$, $113$--$183\,\mathrm{mJ}$): the coherent CPU--GPU interconnect adds a fixed launch overhead that dominates at the sub-millisecond scale of a single $B{=}1$ inference, so although peak throughput of GH200 is high, its single-shot latency is larger than H200's. Normalised per-point energy is therefore lowest on H200 ($\bar{E}/P\!\approx\!11.5\,\mu\mathrm{J}$/point on HX) and highest on GH200 ($\!\approx\!45.9\,\mu\mathrm{J}$/point on HX). The architectural ranking established in the main text is preserved across all four GPUs (Supplementary Table~\ref{tab:supp_efficiency_full}), confirming that the cross-model conclusions are properties of the architecture, not of the specific GPU on which the comparison is reported. The hardware-portability panel above is therefore retained as a separate characterisation rather than a model-efficiency claim.

\begin{table}[!ht]
\centering
\caption{\textbf{Full architecture$\times$GPU$\times$case efficiency matrix.} Each cell shows latency~/~energy per inference (ms~/~mJ) under the identical workload ($B{=}1$, $P\!=\!3{,}977$, $N\!=\!1{,}000$ timed passes, float32). Bold marks the lowest entry within each (case, metric, GPU) group of four architectures. The cross-architecture ordering established in the main text is preserved on every GPU tested, confirming the conclusions are architectural rather than hardware-specific.}
\label{tab:supp_efficiency_full}
\adjustbox{max width=\textwidth}{%
\begin{tabular}{llcccc}
\toprule
\textbf{Case} & \textbf{GPU} & \textbf{MIMONet} & \textbf{GeoFNO} & \textbf{NOMAD} & \textbf{KCN} \\
\midrule
Heat exchanger & A40 & 0.84 / 61 & 5.36 / 611 & \textbf{0.50} / \textbf{60} & 3.11 / 270 \\
 & A100 & 0.81 / 100 & 5.46 / 414 & \textbf{0.50} / \textbf{37} & 3.15 / 207 \\
 & H200 & 0.35 / 46 & 3.08 / 460 & \textbf{0.23} / \textbf{36} & 1.75 / 216 \\
 & GH200 & 1.17 / 183 & 11.52 / 1727 & \textbf{0.80} / \textbf{121} & 4.28 / 611 \\
\midrule
PWR subchannel & A40 & 0.75 / 58 & 5.42 / 522 & \textbf{0.52} / \textbf{51} & 3.12 / 259 \\
 & A100 & 0.74 / 55 & 5.46 / 367 & \textbf{0.50} / \textbf{33} & 3.14 / 195 \\
 & H200 & 0.33 / 39 & 3.03 / 402 & \textbf{0.22} / \textbf{31} & 1.73 / 206 \\
 & GH200 & 1.05 / 154 & 7.13 / 1046 & \textbf{0.80} / \textbf{117} & 4.03 / 569 \\
\midrule
LDC & A40 & \textbf{0.67} / \textbf{61} & 5.22 / 634 & 1.47 / 340 & 2.97 / 284 \\
 & A100 & \textbf{0.63} / \textbf{84} & 5.27 / 437 & 1.42 / 336 & 3.01 / 201 \\
 & H200 & \textbf{0.26} / \textbf{29} & 2.91 / 453 & 0.67 / 159 & 1.62 / 208 \\
 & GH200 & \textbf{0.74} / \textbf{113} & 6.82 / 1066 & 1.12 / 251 & 3.90 / 561 \\
\bottomrule
\end{tabular}}
\end{table}

\section*{Code Availability}
Reproducible code and data are available at \url{https://doi.org/10.5281/zenodo.21986357}; a permanent DOI will be assigned upon publication.


\end{document}